\documentclass[11pt]{article}
\usepackage[margin=1in]{geometry}
\usepackage{breakurl}           
\usepackage{url}                
\usepackage[table]{xcolor}             
\usepackage[]{hyperref}         
\hypersetup{                    
  colorlinks,
  linkcolor={green!80!black},
  citecolor={red!70!black},
  urlcolor={blue!70!black}
}

\usepackage{tikz}
\usepackage{amsmath}
\usepackage{url}

\usepackage[noadjust]{cite}
\usepackage[table]{xcolor}

\usepackage{bm}
\usepackage{mathtools}
\usepackage{xfrac}
\usepackage{bbm}
\usepackage{tikz}
\usepackage{amsmath,amsfonts,bm,amssymb}

\newcommand{\transpose}{\mathsf{T}}

\newcommand{\makeTransparent}[1]{#1!20}

\renewcommand{\vec}{\bm}

\newcommand{\placeholderCount}{m}

\DeclarePairedDelimiterX{\infdivx}[2]{(}{)}{%
  #1\;\delimsize\|\;#2
}

\newcommand{\sampleElement}{x}
\newcommand{\labelElement}{y}

\newcommand{\sampleSpace}{\mathcal{X}}
\newcommand{\labelSpace}{\mathcal{Y}}
\newcommand{\sampleDist}{\mathcal{D}}
\newcommand{\drawnFrom}{\sim}

\DeclareMathOperator*{\argmax}{arg\,max}
\DeclareMathOperator*{\argmin}{arg\,min}
\newcommand{\centroid}{\vec{c}}

\newcommand{\PDF}{p}

\newcommand{\normalDist}{\mathcal{N}}

\newcommand{\rp}{\mathrm{rp}}

\newcommand{\metric}{\rho}

\renewcommand{\Pr}{\mathop{\mathbb{P}}}

\newcommand{\given}{\mid}

\newcommand{\indicatorFunction}{\mathbbm{1}}

\newcommand{\threshold}{\tau}

\DeclarePairedDelimiter{\abs}{\lvert}{\rvert}

\newcommand{\sample}{\vec{\sampleElement}}

\newcommand{\realNumbers}{\mathbb{R}}

\newcommand{\dimension}{d}
\newcommand{\trainingData}{S}

\newcommand{\placeholderDist}{\mathcal{Q}}

\newcommand{\placeholder}{z}

\newcommand{\definedAs}{\triangleq}

\newcommand{\norm}[1]{\left\lVert#1\right\rVert}

\newcommand{\groupA}{\mathsf{a}}
\newcommand{\groupB}{\mathsf{b}}

\newcommand{\group}{\mathsf{g}}
\newcommand{\placeholderGroup}{\mathsf{g}}

\newcommand\MTkillspecial[1]{
\bgroup
\catcode`\&=9
\let\\\relax%
\scantokens{#1}%
\egroup
}
\DeclarePairedDelimiter\brparen
\lparen\rparen
\reDeclarePairedDelimiterInnerWrapper\brparen{star}{
\mathopen{#1\vphantom{\MTkillspecial{#2}}\kern-\nulldelimiterspace\right.}
#2
\mathclose{\left.\kern-\nulldelimiterspace\vphantom{\MTkillspecial{#2}}#3}}

\newcommand{\getPerturbation}{\Delta}
\newcommand{\embeddingFunc}{\mathrm{f}} 
\newcommand{\perturbation}{\vec{\delta}}

\newcommand{\hinge}{\mathrm{G}}

\usepackage{clipboard}

\usepackage{amsthm}
\usepackage[normalem]{ulem}
\usepackage[shortlabels]{enumitem}
\usepackage[super]{nth}
\usepackage{tikz}
\usepackage{pgfplots}
\theoremstyle{remark}
\newtheorem{definition}{Definition}[section]

\usepackage{cleveref}

\newtheorem{proposition}[definition]{Proposition}

\newtheorem{nullHyp}[definition]{Null Hypothesis}

\crefname{nullHyp}{null hypothesis}{null hypotheses}  
\Crefname{nullHyp}{Null hypothesis}{Null hypotheses}

\usepackage[english]{babel}
\usepackage{blindtext}

\usepackage{fancyhdr}

\newcommand{\newDeletion}[1]{}

\usepackage{rotating}

\makeatletter
\newcommand{\linebreakand}{%
  \end{@IEEEauthorhalign}
  \hfill\mbox{}\par
  \mbox{}\hfill\begin{@IEEEauthorhalign}
}
\makeatother

\usepackage{algorithm}
\usepackage[noend]{algpseudocode}
\usepackage{mwe}
\usepackage{tikz}
\usetikzlibrary{arrows}
\usepackage{verbatim}

\usepackage{booktabs,arydshln,multirow}

\usepackage{dirtree}
\usepackage{wrapfig}

\usepackage[font=footnotesize,labelfont=bf]{caption}
\usepackage{subcaption}

\makeatletter
\def\adl@drawiv#1#2#3{%
        \hskip.5\tabcolsep
        \xleaders#3{#2.5\@tempdimb #1{1}#2.5\@tempdimb}%
                #2\z@ plus1fil minus1fil\relax
        \hskip.5\tabcolsep}
\newcommand{\cdashlinelr}[1]{%
  \noalign{\vskip\aboverulesep
           \global\let\@dashdrawstore\adl@draw
           \global\let\adl@draw\adl@drawiv}
  \cdashline{#1}
  \noalign{\global\let\adl@draw\@dashdrawstore
           \vskip\belowrulesep}}
\makeatother

\begin{document}
 \newclipboard{clipboard}
\newpage
\date{}

\title{\Large \bf Fairness Properties of Face Recognition and Obfuscation Systems}

\author{
  Harrison Rosenberg\\
  University of Wisconsin--Madison\\
  \texttt{hrosenberg@ece.wisc.edu}
  \and
  Brian Tang\\
  \;\;\;\;\;\;\;\;\;University of Michigan\;\;\;\;\;\;\;\;\;\\
  \texttt{bjaytang@umich.edu} 
  
  \and
  
  Kassem Fawaz\\
  University of Wisconsin--Madison\\
  \texttt{kfawaz@wisc.edu}
  \and
  Somesh Jha\\ 
  University of Wisconsin--Madison\\
  \texttt{jha@cs.wisc.edu}
}

\maketitle

\begin{abstract}
The proliferation of automated face recognition in the commercial and government sectors has caused significant privacy concerns for individuals.  One approach to address these privacy concerns is to employ evasion attacks against the metric embedding networks powering face recognition systems: Face obfuscation systems generate imperceptibly perturbed images that cause face recognition systems to misidentify the user. Perturbed faces are generated on metric embedding networks, which are known to be unfair in the context of face recognition.  A question of demographic fairness naturally follows: 
\textit{are there demographic disparities in face obfuscation system performance?} We answer this question with an analytical and empirical exploration of recent face obfuscation systems. Metric embedding networks are found to be demographically aware: face embeddings are clustered by demographic. We show how this clustering behavior leads to reduced face obfuscation utility for faces in minority groups. An intuitive analytical model yields insight into these phenomena.
\end{abstract}

\section{Introduction}

Automated face recognition has proliferated in various commercial and government sectors. Face recognition systems can identify users on social media, search for missing persons, aid law enforcement and surveillance, and verify identities of individuals~\cite{feiner2021police,roussi2020facerecognition}. The widespread adoption of face recognition systems has been swift with the emergence of metric embedding networks such as FaceNet~\cite{schroff2015facenet} and ArcFace~\cite{deng2018arcface} as well as the abundance of labeled face data~\cite{LFWTech,caovggface2}.

Recent coverage of data breaches, privacy law violations, and the adoption of face recognition by law enforcement entities have shed light on the significant privacy issues with face recognition systems. To mitigate growing privacy concerns, face obfuscation systems have been proposed to hide user identities. Several of these systems, such as Face-Off~\cite{DBLP:FaceOff}, LowKey~\cite{DBLP:LowKey}, and Foggysight~\cite{evtimov21foggysight}, leverage properties of evasion attacks against machine learning models~\cite{goodfellow2014explaining, madry2017towards, carlini17wagner}. By introducing small, structured, and imperceptible perturbations to their face, a user can evade identification by a face recognition system. Such systems are attractive for an end-user: perturbations are often visually acceptable to the user; features of social media applications, such as face-augmenting filters, do not suffer; and the obfuscation mechanism may run locally, without access to target face recognition systems.

However, face obfuscation systems suffer major shortcomings. Researchers have identified the ability of face recognition systems to adaptively learn from perturbed faces~\cite{tramer2021poisoning} and re-identify them in the future. In this work, we uncover another shortcoming of such systems: \textit{the presence of performance disparities with respect to demographics.} This disparity leads to the following research questions:
\begin{itemize}[noitemsep,leftmargin=0.4cm,topsep=5pt]
    \item Are the metric embedding networks underlying face obfuscation systems aware of demographic attributes in faces?
    \item How does the behavior of face obfuscation systems depend on the demographic attributes of faces?
    \item Do bias mitigation strategies for face recognition systems also mitigate bias in face obfuscation systems?
\end{itemize}

This paper characterizes demographic disparities of face obfuscation systems and their underlying metric embedding networks\footnote{Code repository: \url{https://github.com/wi-pi/fairness_face_obfuscation}}. Previous research has studied the fairness and robustness properties of face recognition~\cite{DBLP:Gendershades, nanda20umdfairness}\footnote{See \cref{sec:related} for further discussion.} in the classification setting. However, we study the fairness properties of face recognition and obfuscation in the context of the metric embedding networks --- the real-world setting for such systems. Our work yields the following insights into fairness implications of face recognition and obfuscation. 

%
 
\noindent
\textbf{Are the metric embedding networks underlying face obfuscation systems aware of the demographic attributes of faces?}
We observe face recognition systems are better at discerning individuals in different demographic groups than discerning individuals within the same demographic.  Without explicit access to demographic information, face recognition systems still learn to differentiate demographic groups.


\noindent
\textbf{How does the behavior of face obfuscation systems depend on the demographic attributes of faces?}
We analyze two recent face obfuscation systems. Face-Off~\cite{DBLP:FaceOff}  and LowKey~\cite{DBLP:LowKey} serve as proxies for targeted face obfuscation and untargeted face obfuscation, respectively. Minority groups require stronger perturbations to successfully obfuscate a face. This is especially true in a black-box setting, such as the Face++, Azure, and AWS Rekognition face recognition APIs. We also show that faces perturbed by untargeted attacks often retain their original demographic attributes. We conclude that larger, more visible perturbations are necessary to successfully target identities in demographic groups different from the original image.

\noindent
\textbf{How does the training regime affect the behavior of face obfuscation?}
 We compare the effects of a standard face recognition network training method with two alternative training regimes designed to mitigate bias.  The first regime is the training procedure defined by Xu et al~\cite{xu2021robustfair}. The second regime is the training of models on demographically balanced datasets. We show these techniques do not entirely eliminate demographic-wise performance disparities.

To aid our response to these three questions, we devise an analytical model based on a mixture of Gaussian distributions and Principal Component Analysis (PCA). This analytical model allows us to formalize the apparent behavior of face recognition and obfuscation when conditioning on the demographic group. Our model reveals that obfuscated faces are more likely to belong to their original demographic group.

\section{Background}


The terminology with respect to population demographics used in this paper follows that of Buolamwini and Gebru~\cite{DBLP:Gendershades} and Nanda et al.~\cite{nanda20umdfairness}, two leading works on face recognition fairness.  In the dataset annotations, there are only two sexes, hence we use ``male'' and ``female.''  As for ethnicity, previous literature utilizes terms such as ``White'' ``Black,'' ``Asian,'' and ``Indian'' within their attribute annotations~\cite{kumar2009lfwattributes}.  We find it more accurate to refer to these demographic labels as ``race.''  For consistency, we use the same demographic attribute labels in the VGGFace2 dataset in our face recognition and face obfuscation performance evaluations.

\subsection{Notation}
\label{subsec:notation}

We consider the setting in which there exists an input space $\sampleSpace \subseteq \realNumbers^{\dimension}$ and a discrete label set $\labelSpace$.  A subset of examples, also referred to as a dataset, is denoted as $\trainingData{} \subseteq \sampleSpace \times \labelSpace{}$. Sometimes we abuse notation and let $\trainingData$ contain only unlabeled examples $\left\{\sample_1,\sample_2,\dotsc\right\}$.  A sample $\sample$ is a $\dimension$-dimensional real-valued vector. Often, $\sample$ refers to a cropped face, $\dimension$ is the number of pixels in the cropped face multiplied by three (the RGB channels), and $\labelSpace$ to the set of identities. 
  Given a vector $\vec{\placeholder}$, $\placeholder_j$ denotes its $j^{\mathrm{th}}$ entry. 
Calligraphic capital letters denote probability distributions. $\sampleDist$ represents the distribution from which training data $\trainingData$ are drawn. $\indicatorFunction$ denotes the indicator function.  Let $\placeholder$ be a Boolean expression.  $\indicatorFunction[\placeholder]$ evaluates to $1$ if $\placeholder$ is true, otherwise $\indicatorFunction[\placeholder]$ evaluates to $0$.  A metric is denoted by $\rho : \sampleSpace \times \sampleSpace \to \realNumbers_+$. 

\subsection{Face Recognition System}
\begin{figure}[t]
    \centering
    \includegraphics[width=0.6\columnwidth]{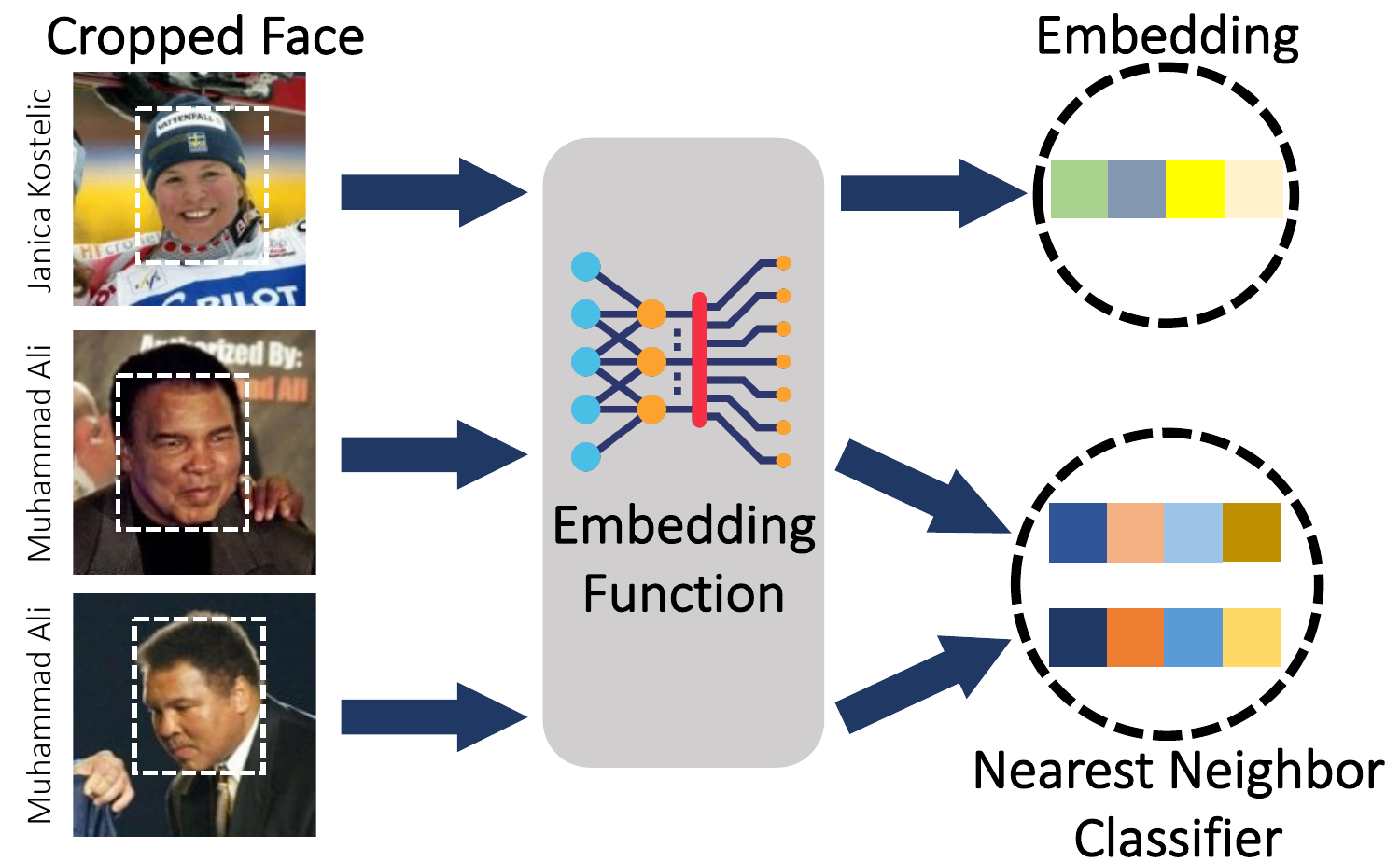}
    \caption{For well-trained embedding functions, embeddings of images belonging to the same identity will have smaller pairwise distances than the pairwise distances between embeddings of different identities. 
    }
    \label{fig:embeddingFunction}
\end{figure}

The core component of a face recognition system is a \emph{metric embedding network}. The metric embedding network, denoted by $\embeddingFunc_k : \realNumbers^{\dimension} \to \realNumbers^{k}$, is a neural network which takes an RGB face image $\sample$ as input, and returns a $k$ dimensional embedding. Thus, the term \emph{embedding} refers to the $k$-dimensional face representation output by the metric embedding network. We sometimes omit the subscript $k$ when referring to a generic embedding function.  The goal of a metric embedding network is to map high dimensional images into an embedding space such that any two images belonging to the same identity have lower pairwise distance than any two images with different identities. 
Metric embedding networks are typically trained with one of two classes of loss functions: contrastive loss~\cite{Hadsellcontrastive} and triplet loss~\cite{schroff2015facenet}. The functionality of a face recognition system is depicted in \cref{fig:embeddingFunction}. 

Upon obtaining an embedding, tasks such as clustering, matching, or classification may be performed. A more performant metric embedding network is one which only matches embeddings of faces belonging to the same identity.  A match occurs when two embeddings are sufficiently close: Given a non-negative, real-valued threshold $\threshold$ and two examples $\sample$, $\sample'$, a match occurs when $\metric(\sample,\sample') \leq \threshold$. Often, $\metric$ is the $\ell_2$ norm.

Matching performance is measured by $\mathrm{TPR}_\placeholder$ which is a parametrized notion of true positive rate.  For choice of metric $\metric$, the match threshold threshold $\threshold$ is chosen such that it satisfies a false acceptance rate upper-bounded by $\placeholder$.   
{\small
\begin{equation}
    \mathrm{TPR}_{\placeholder}(\trainingData,\trainingData') \definedAs \sum_{(\sample_i,\labelElement_i,\sample'_i,\labelElement'_i) \in \left\{\trainingData,\trainingData' \given \labelElement_i = \labelElement_i'\right\}}^{\placeholderCount} \frac{\indicatorFunction\left[\metric\left(\embeddingFunc(\sample),\embeddingFunc(\sample')\right) <\threshold\right]}{\abs{\left\{\trainingData,\trainingData' \given \labelElement_i = \labelElement_i'\right\}}}
\end{equation}
}
where threshold $\threshold$ is defined as
{\small
\begin{align}
\begin{split} \label{eq:FARDefine}
    \max \;  \threshold \quad \mathrm{s.t.} \quad \placeholder > \sum_{(\sample,\labelElement,\sample',\labelElement') \in \left\{\trainingData,\trainingData' \given \labelElement_i \neq \labelElement_i'\right\}} \frac{\indicatorFunction\left[\metric\left(\embeddingFunc(\sample),\embeddingFunc(\sample')\right) <\threshold\right]}{{\abs{\left\{\trainingData,\trainingData' \given \labelElement_i \neq \labelElement_i'\right\}}}}
    \end{split}
\end{align}
}
 and the right side of inequality \eqref{eq:FARDefine} is false acceptance rate.


\subsection{Face Obfuscation Systems}
\label{sec:faceObfuscation}

Face recognition systems are vulnerable to adversarial examples. Such vulnerabilities were famously discovered by Szegedy et al~\cite{DBLP:journals/corr/SzegedyZSBEGF13}. Small, structured perturbations, imperceptible to humans, may cause networks to misclassify given inputs. This branch of machine learning research has led to the design of many attack algorithms, of which the most prominent are so-called evasion attacks. Examples of evasion attacks include the Projected Gradient Descent (PGD)~\cite{madry2017towards} and Carlini-Wagner (CW)~\cite{carlini17wagner} attacks. The bulk of study focuses on algorithmic generation of $\ell_p$ norm-bounded perturbations via noisy gradient-based optimization.

Instead of seeing adversarial examples as a challenge, privacy researchers have demonstrated that systems which leverage such structured perturbations can provide users with privacy utility against face recognition.  These privacy benefits are encapsulated in systems known as face obfuscation systems. In this section, we present notation and discuss work relevant to such face obfuscation systems.

\subsubsection{Threat Model}
\label{subsec:threat}
 In the face obfuscation setting, an end user considers the machine learning provider to be the main threat. Such threats include data breaches~\cite{holmes2021facebook},  cyber-stalkers~\cite{harwell_2021}, web scrapers, big government entities~\cite{Goodwin_2021}, and more. Face obfuscation systems counter such threats: Prior to upload, the user applies (imperceptible) perturbations to their face images. Perturbations are designed so the face recognition system is aware of the presence of the face, but the predicted identity of the face is incorrect. In other words, perturbations are constructed so that, from the perspective of the face recognition system, the user impersonates another identity. This is the \textit{white-box} setting:  examples are generated on the target model directly.

We also consider the black-box setting. Perturbations applied in the black-box setting leverage an important property of adversarial examples: their ability to transfer across models~\cite{liu17transferable,papernot2016transferability}.  Transferability allows users to impersonate identities without accessing a target model. In the \textit{black-box} setting, face obfuscation systems leverage transferability by querying a surrogate model to generate perturbed faces. 

\subsubsection{Overview of Face Obfuscation Systems}

The earliest works in face obfuscation explore physically perturbed examples. For example, Sharif et al.~\cite{sharif2016glasses} present physically realizable glasses that allow a human user to impersonate a target individual. With the recent societal focus of online privacy, researchers have shifted their focus to digital face obfuscation systems. Digital face obfuscation systems are better suited for social media and internet applications.  Examples of such systems include FAWKES~\cite{DBLP:Fawkes}, Face-Off~\cite{DBLP:FaceOff}, Low-Key~\cite{DBLP:LowKey}, and FoggySight~\cite{evtimov21foggysight}. With the exception of FAWKES, which leverages data poisoning attacks, these face obfuscation systems utilize evasion attacks in an attempt to hide the user's identity. 

We now describe face obfuscation systems more formally. Let $\getPerturbation$ denote an evasion attack function (e.g. PGD, CW), and let $\perturbation$ denote a generic perturbation output by the evasion attack function, i.e. $\getPerturbation(\sample)$. A face obfuscation system feeds $\sample + \perturbation$ into the face recognition system.  Note that $\getPerturbation$ may include dependence on the metric embedding network, the underlying dataset, or some other surrogate model.  
\subsubsection{Evasion Attacks on Metric Embedding Networks}
\label{subsec:Adversarial}

 There are two types of evasion attacks: targeted and untargeted. We first describe the embedding centroid before defining the attacks: Given dataset $\trainingData$, denote by $\centroid_{\embeddingFunc,\labelElement}$ the embedding centroid of identity $\labelElement$ as computed on embedding function $\embeddingFunc$:
{\small
\begin{equation}
    \centroid_{\embeddingFunc,\labelElement} \definedAs \frac{1}{\abs{\left\{(\sample',\labelElement') \in \trainingData  \; \middle| \; \labelElement' = y \right\}}}\sum_{(\sample',\labelElement')\in \trainingData} \embeddingFunc(\sample')\cdot\indicatorFunction\left[\labelElement' = y\right]
\end{equation}
}




\noindent \textbf{Untargeted Attacks:}
\Copy{copy:Untargeted_attack}{
Given a labeled example $(\sample,\labelElement)$, untargeted attacks find a perturbation $\perturbation$ for which obfuscated face $\sample + \perturbation$ impersonates some identity $\labelElement_0$, where $\labelElement_0 \neq \labelElement$. Given a metric embedding network $\embeddingFunc: \sampleSpace \to \labelSpace$, and metrics $\metric_1,\metric_2$ an \emph{untargeted attack} is formulated as:
{\small
\begin{align*}
\begin{gathered}
  \min_{\perturbation \given \left\{\sample + \perturbation\in \sampleSpace\right\}}  \metric_1 (\perturbation, \vec{0})\\
  \textrm{s.t.}  \quad \argmin_{\labelElement' \in \labelSpace} \metric_2(\centroid_{\embeddingFunc,\labelElement'},\embeddingFunc(\sample))  
     \neq \argmin_{\labelElement' \in \labelSpace} \metric_2(\centroid_{\embeddingFunc,\labelElement'},\embeddingFunc(\sample + \perturbation))
\end{gathered} 
\end{align*}}
}

The problem with the formulation above is its intractable constraint.  To make the attack implementable in practice, the constraints must be relaxed.  For a labeled example $(\sample,\labelElement)$, the optimization objective which yields an untargeted perturbation can be written as:
{\small
\begin{align}
\begin{split}
\label{eq:untar_lowkey}
\argmax_{\perturbation \given  \left\{\sample' + \perturbation \in \sampleSpace\right\}}  \metric_2(\embeddingFunc(\sample'+\perturbation),\centroid_{\embeddingFunc,\labelElement}) \quad
\textrm{s.t.} \quad \metric_1(\perturbation,\vec{0}) \leq \epsilon
\end{split}
\end{align}
}
In this paper, we use the LowKey attack~\cite{DBLP:LowKey} to instantiate the attack described in \cref{eq:untar_lowkey}. Within LowKey, PGD is used to solve  \cref{eq:untar_lowkey}, $\metric_1$ is the Learned Perceptual Image Patch Similarity (LPIPS) metric~\cite{zhang2018lpips}, and  $\metric_2$ is the distance between the original face and perturbed face in the embedding space. Note that the latter distance is averaged through an ensemble of models and after applying Gaussian smoothing.  \smallskip

\noindent \textbf{Targeted Attacks:}  
\Copy{copy:Targeted_attack}{
Given a labeled example $(\sample,\labelElement)$, and target identity $\labelElement_0$, targeted attacks find perturbation $\perturbation$ so that obfuscated face $\sample + \perturbation$ impersonates $\labelElement_0$. Given a metric embedding network $\embeddingFunc: \sampleSpace \to \realNumbers^k$, and metrics $\metric_1,\metric_2$ a \emph{targeted attack} may be formulated as:}
{\small
\begin{align}
\Copy{copy:targeted_attack_formula}{
\begin{split}
  \min_{\perturbation \given \left\{\sample + \perturbation\in \sampleSpace\right\}}  \metric_1 (\perturbation, \vec{0}) \quad
  \textrm{s.t.} \quad \labelElement_0 = \argmin_{\labelElement \in \labelSpace} \metric_2(\centroid_{\embeddingFunc,\labelElement},\embeddingFunc(\sample + \perturbation))
\end{split}}\label{eq:targetedAttack}
\end{align}
}
Similar to the untargeted case, targeted attacks on face recognition systems relax the above constraints to arrive at a more tractable optimization formulation.  Multi-class hinge loss, as used by FaceOff~\cite{DBLP:FaceOff}, provides the desired relaxation of the constraints in the targeted attack \cref{eq:targetedAttack}.  Given a perturbed example $\sample + \perturbation$, target label $\labelElement_0$, and positive real number $\kappa$, the multi-class hinge loss is denoted by $\hinge_{\kappa}(\sample + \perturbation, \labelElement_0)$ where: 
{\small
\begin{align*}
\begin{gathered}
    \hinge_{\kappa}(\sample+\perturbation,\labelElement_0) \definedAs \max\Big\{0,\kappa+\metric_2(\sample+\perturbation, \centroid_{\embeddingFunc,\labelElement_0}) -\max_{\labelElement' \neq \labelElement_0}\metric_2(\sample+\perturbation, \centroid_{\embeddingFunc,\labelElement'})\Big\}
\end{gathered}
\end{align*}
}
For a labeled example $(\sample,\labelElement)$, and target label $\labelElement_0 \neq \labelElement$, the CW attack can minimize the following optimization objective to yield a targeted perturbation:

{\small
\begin{equation}
\label{eq:face-off}
\argmin_{\perturbation \given \left\{\sample + \perturbation \in \sampleSpace\right\}}  \metric_1(\perturbation, \vec{0}) \quad
\textrm{s.t.} \quad \metric_1(\perturbation,\vec{0}) \leq \epsilon \;
\textrm{and} \;  \hinge_{\kappa}(\sample + \perturbation,\labelElement_0) \leq 0
\end{equation}}

\subsection{Analytical Techniques}\label{sec:analyticalTechniques}
We utilize a combination of established data analysis techniques to assess the fairness of both face recognition and face obfuscation systems.

\noindent \textbf{PCA:}
Neural networks with nonlinearities are notoriously difficult to analyze. To gain intuition for face obfuscation, we analyze PCA. PCA is a theoretically tractable embedding function which can be represented as a neural network. Hence, we use PCA as a proxy for non-linear embedding functions.
  \smallskip

\noindent \textbf{t-SNE:}
The t-Distributed Stochastic Neighbor Embedding (t-SNE)~\cite{maaten2008tsne} is a dimensionality reduction technique useful for visualizing high dimensional embedding spaces and image datasets.  Visualizations rendered by t-SNE are two or three dimensional. t-SNE is a variant of Stochastic Neighbor Embeddings~\cite{hinton2003sne} which avoids crowding data points and can capture the implicit structure of data. t-SNE plots aid our discussion of the embedding space geometry. \smallskip

\noindent \textbf{TCAV:}
Testing with Concept Activation Vectors (TCAV)~\cite{kim18tcav} is a tool used to interpret deep neural networks.  Given user-specified high-level concepts, such as patterns or colors, linear classifiers are trained on the neural network's activations for those concepts.  Concept Activation Vectors (CAVs) are extracted from the vector orthogonal to the linear classifier's decision boundary, and a statistical significance test is performed on the TCAV score generated from the dot product of each CAV and the model's gradients.    \smallskip

\noindent \textbf{Fairness Definitions:}
There is no single definition of fairness, mathematical or otherwise~\cite{barocas-hardt-narayanan}.  Hence, we study several quantities which have an intuitive connection to both face recognition and face obfuscation systems. These quantities include a comparison, by demographic, of face obfuscation success.  When success rates are equal across demographics, the fairness constraint known as statistical parity~\cite{dwork_fairness} is satisfied.  We also compare, by demographic, the strength of perturbations necessary to yield a successful face obfuscation.  For targeted obfuscation, we study the strength of such perturbations necessary to impersonate identities within both the inter-demographic group and intra-demographic group settings. We also study how likely untargeted perturbations are to change the perceived demographic of an image. 

True positive rate balancing is another notion of fairness which appears in machine learning literature.  In \cref{subsubsec:Xu_et_al_model}, we see that a specialized training procedure by Xu et al.~\cite{xu2021robustfair} yields a metric embedding network for which demographic groups are approximately equal in their matching performance $\mathrm{TPR}_{\placeholder}$.  An optimization constraint explicitly enforcing equal true positive rates between groups is the fairness metric known as Equalized Odds~\cite{hardt2016equality}. \smallskip

\noindent \textbf{Statistical Significance:}
We use statistical significance testing to draw conclusions from experiments. The most general two-sample $t$-test, used to determine if two distributions have unequal means, is Welch's $t$-test \cite{welch-t-test}. Welch's $t$-test is applicable when population variances are nonequal and/or the number of elements in each sample differs. For statistical tests on more than two samples, we apply the Alexander-Govern test \cite{alexandergoverntest}, a multi-sample generalization of Welch's $t$-Test. Each statistical test is designed to determine if any two or more samples are drawn from the same distribution.  In all our statistical tests, we consider a $p$-value less than $0.05$ to be significant. For the remainder of the paper, we will omit the specific $t$-test used to obtain $p$-values. $p$-values are implicitly assumed to have been obtained by either Welch's $t$-test or the Alexander-Govern test. Because we have only applied one null hypothesis per statistical test, our statistical tests do not suffer from the multiple testing problem.


\section{Experiment Motivation and Overview}
\label{sec:methodology}

From our experiments and analysis, we wish to understand how biases inherent in both face recognition datasets and metric embedding networks impact the performance of face obfuscation systems. We answer three questions:

\begin{enumerate}[noitemsep,leftmargin=0.4cm,topsep=5pt]
    \item{\textbf{\textit{Bias in Face Recognition}: What is the baseline bias present in face recognition systems?}}
     Our experiments concur with existing literature: when conditioning by demographic, there is a disparity face recognition system performance.  Further, we identify that networks learn to identify skin-tone in early layers of the network.
    \item \textbf{\textit{Effectiveness of Obfuscation}:  How does the strength of perturbation necessary to obfuscate a face depend on a face's demographic?}
    Our findings indicate face recognition systems are less robust to perturbations applied to faces from minority demographic groups. For minority demographic groups, the perturbation strength necessary to impersonate an identity is smaller compared to majority demographics. Consequently, obfuscated faces tend to remain classified as a member of the same demographic group.  Performance disparities are evident between demographics in both targeted and untargeted obfuscation.
    \item \textbf{\textit{Bias Mitigation}: How do bias mitigation strategies applied during training affect the utility of face obfuscation?}
    We apply two bias mitigation strategies to the training of metric embedding networks. The first strategy is a training procedure defined by Xu et al.~\cite{xu2021robustfair}. The second strategy  is training on demographically balanced datasets. With such training, we see the resulting embeddings are less clustered. Performance disparities of the face obfuscation system with respect to demographics are also attenuated.  The benefits of bias mitigation are not free: overall model accuracy is reduced. 
\end{enumerate}
\subsection{Experimental Setup}\label{subsec:ExperimentalSetup}
\noindent 
\textbf{Datasets and Models:}
We utilize the LFW and VGGFace2 datasets in our evaluation. In the white-box setting, we generate perturbed faces on the FaceNet model. In the black-box setting, we test perturbed faces for transferability  by evaluating them on seven different models. Three of the seven models are commercial face recognition APIs: Face++~\cite{facepp}, Azure Face~\cite{azure}, and AWS Rekognition~\cite{awsrekognition}. The remaining four are pre-trained open source models: OpenFace~\cite{amos2016openface}, DeepFace~\cite{taigman2014deepface}, ArcFace~\cite{deng2018arcface}, and VGGFace~\cite{caovggface2}. Each model uses convolutional neural network architecture similar to FaceNet. 

\noindent
\textbf{Fundamental Comparison:}
To determine the relationship between demographics and face obfuscation performance, we consider the performance of obfuscation targeting identities in the same demographic group as the source identity, and obfuscation targeting identities in demographic groups different from that of the source identity. In particular, we consider six demographic attributes, four for race (Black, White, Indian, Asian) and two for sex (female, male). 50 identities per attribute from LFW are sampled\footnote{The LFW demographic attributes were annotated using attribute classifiers described by Kumar et al.~\cite{kumar2009lfwattributes}.}.  These images are referred to as source images. Source images are inputs to the untargeted attack. Two targeted attacks scenarios are considered:
\begin{itemize}[noitemsep,leftmargin=0.4cm,topsep=5pt]
    \item \textbf{Same demographic:} We choose 49 pairwise combinations of target identities from the same race/sex for each source identity. This sampling leads to 2450 source-target pairs of the same demographic.
    \item \textbf{Different demographic:} We subsample, uniformly at random, 15 target identities from each race group of which the source identity is not a member for a total of 45 target identities. For the sex demographic, we assemble 50 target identities from the opposite sex.  This sampling leads to 2250 source-target pairs of the different races and 2500 source-target pairs of the different sex.
\end{itemize}
We generate untargeted adversarial examples for each of the 5,749 identities in the LFW dataset and their associated images.  We generate targeted adversarial examples for earlier scenario's 300 identities and 80,000+ targeted examples corresponding to the 28,700 pairs of identities. \Copy{copy:impliedDataset}{If unstated within a caption, the chosen dataset is LFW and the chosen metric embedding network is FaceNet.} \smallskip

\noindent
\textbf{Obfuscation Techniques:}
We perform our evaluation using untargeted and targeted variants of face obfsucation systems, as described in \cref{sec:faceObfuscation}. Utilizing the Face-off face obfuscation system~\cite{DBLP:FaceOff} and the FaceNet model~\cite{schroff2015facenet}, we generate adversarial examples on a subset of LFW~\cite{LFWTech}. For \cref{subsec:error,subsec:obfuscationEffectiveness} we use 0, 5, and 10 as our margin values ($\kappa$ in \cref{eq:face-off}).  We also generate untargeted perturbations with the LowKey~\cite{DBLP:LowKey} procedure described in \cref{subsec:Adversarial}. The ensemble used by LowKey includes two pre-trained ArcFace models~\cite{deng2018arcface} and two pre-trained Cosface models~\cite{cosface_paper}. We report the obfuscation success rate as an indicator of perturbation effectiveness. \Copy{copy:Obfuscation_Success_Slope}{In the targeted case, obfuscation success rate measures the proportion of perturbed faces which match their intended targets, so we expect targeted obfuscation success rate to decrease as a function of the threshold $\threshold$. Furthermore, the distance between an embedding and its target directly controls obfuscation success. In the untargeted case, obfuscation success rate measures the proportion of perturbed faces that evade their source identity. We expect the untargeted obfuscation rate to increase as a function of the threshold $\threshold$.}

\subsection{An analytical model for face obfuscation}\label{subsec:analyticalModel}
 \Copy{copy:Analytical_model_introduce}{ Consistent with usage of toy models in previous machine learning literature~\cite{schmidt2018adversarially,hashimoto18a,pmlr-v97-cohen19c}, we devise a hierarchical Gaussian distribution with which we explore fair face obfuscation. While our model does entirely capture neural network behavior, the model conveys intuition on how how discrepancies in demographic sampling lead to disparities we observe in face obfuscation utility. To this end, the analytical
model consists of a $k$-component PCA and hierarchical Gaussian distribution. The $k$-component PCA, which utilizes projections onto the leading $k$ principal components to perform its dimensionality reduction, is the theoretically tractable proxy for a nonlinear metric embedding network. Use of the hierarchical Gaussian is inspired by both the hierarchical nature of popular face recognition datasets and the embedding space geometry. Though all samples drawn from our simplified model are vectors, we use terms \emph{identity} and \emph{image} to draw parallels between the hierarchical nature of our probabalistic model and the hierarchical structure of existing face recognition datasets.}


Within the hierarchical Gaussian are two mutually exclusive groups: group~$\groupA$ and group~$\groupB$. Sometimes a placeholder $\placeholderGroup$ is used to represent a group $\placeholderGroup \in \left\{\groupA,\groupB\right\}$. $\vec{\mu}_{\groupA} \in \realNumbers^{\dimension}$ and $\vec{\mu}_{\groupB} \in \realNumbers^{\dimension}$ denote the mean vectors for population groups $\groupA$ and $\groupB$, respectively. Moreover, $\vec{\mu}_{\groupA} = -\vec{\mu}_{\groupB}$, $\norm{\vec{\mu}_{\groupA}}_2 = 1$, and $\norm{\vec{\mu}_{\groupB}}_2 = 1$. \Cref{fig:analyticalModel} is a visual depiction of our analytical model.

The $i^{\mathrm{th}}$ identity in group $\placeholderGroup$ is denoted by $\vec{\nu}_{\placeholderGroup,i} \in \realNumbers^{\dimension}$.  The $j^{\mathrm{th}}$ image representing identity $\vec{\nu}_{\placeholderGroup,i}$ is denoted by $\sample_{\placeholderGroup,i,j} \in \realNumbers^\dimension$.  $\vec{\Sigma}_{\groupA} \in \realNumbers^{\dimension \times \dimension}$ and $\vec{\Sigma}_{\groupB}\in \realNumbers^{\dimension \times \dimension}$ are diagonal covariance matrices. Furthermore, $\vec{\Sigma}_{\groupA} = \gamma\vec{\Sigma}_{\groupB}$ where $\gamma \in \realNumbers^+$. 

\Copy{copy:gamma_parameter_description}{By construction, the $\gamma$ parameter captures demographic imbalance in sampling. Let us assume, that if both groups $\groupA$ and $\groupB$ are sampled equally in a manner matching the natural distribution, then $\vec{\Sigma}_{\groupA}=\vec{\Sigma}_{\groupB}$.  Without loss of generality, let us assume $\gamma \in (0,1]$. As $\gamma$ decreases, the sampling from the natural distribution deviates further from the underlying distribution. Disparities in embedding density become apparent: When group $\groupA$ has few samples, embeddings of identities within group $\groupA$ are close together, similar to how minority groups are depicted in \cref{fig:tsne}. }

An identity $\vec{\nu}_{\placeholderGroup,i}$ is drawn from the identity distribution $\normalDist(\vec{\mu}_{\placeholderGroup},\vec{\Sigma}_{\placeholderGroup})$.  The identity distribution may be thought of as a hyperprior on images. An image $\sample_{\placeholderGroup,i,j}$ is drawn from $\normalDist(\vec{\nu}_{\placeholderGroup,i}, \beta \vec{I})$ where $\beta$ is a positive, real-valued number. For each identity $\vec{\nu}_{\placeholderGroup,i}$, exactly $\placeholderCount$ images are drawn from $\normalDist(\vec{\nu}_{\placeholderGroup,i}, \beta \vec{I})$.  Lastly, we denote by $\sampleDist_{\placeholderGroup}$ the distribution of images in group $\placeholderGroup$. Thus, given $\alpha \in (0,1)$, we can represent, the hierarchical Gaussian distribution as $\sampleDist = \alpha\sampleDist_{\groupA} + (1-\alpha)\sampleDist_{\groupB}$.

 In the context of fair face obfuscation, we concern ourselves with how well the embedding represents a particular group from the lens of that group.  This is captured by relative projection distance. The relative projection distance measures how well the $k$-component PCA embedding function represents a particular group, from the lens of that group:

\begin{definition}
Let $\trainingData$ be a sample of images drawn from the overall synthetic distribution $\sampleDist$.  The relative projection distance of a point $\sample$, a member of group $\placeholderGroup$, with respect to the leading $k$ principal components of a dataset $\trainingData$ is denoted by $\metric_{\rp, \trainingData,\placeholderGroup,k}: \sampleSpace \to \realNumbers^+$.  More precisely:
{\small
\begin{equation}
\label{eq:relative_proj_dist}
    \metric_{\rp, \trainingData,\placeholderGroup,k}(\sample) \definedAs \frac{\norm{\left(\sample-\sum_{i=1}^{k}\left\{\frac{\vec{q}_i^\transpose \sample}{\norm{\vec{q}_i}_2 \norm{\sample}_2}\vec{q}_i\right\}\right)}_2}{\sum_{j=1}^{\dimension}\left(\vec{\Sigma}_{\placeholderGroup}\right)_{jj}},
\end{equation}
}
where the eigen-decomposition of the covariance of overall synthetic data distribution $\sampleDist$ is  $\vec{\Sigma} = \vec{Q}\vec{\Lambda}\vec{Q}^{\transpose}$.  Furthermore, $\vec{Q}$ may be decomposed as $\vec{Q} = \left[\vec{q}_1,\dotsc,\vec{q}_\dimension\right]^\transpose$.
\end{definition}




The numerator of relative projection distance is the norm of the portion of sample $\sample$ which is not captured by $\embeddingFunc_k$.   The denominator represents a group specific normalization factor representing the overall variance within all identity vectors $\vec{\nu}_{\group,i}$.  It is this normalization factor which allows us to show how error can be measured in the context of group $\placeholderGroup$.

\usetikzlibrary{pgfplots.groupplots}

\begin{figure}[t]
\centering
\begin{tikzpicture}
\begin{groupplot}[group style={group size= 2 by 1},width = 8cm]
\nextgroupplot[
	xmin=-3,   xmax=3,
	ymin=-3,   ymax=3,
	line width=1.0pt,
	extra x ticks={-1,1},
	extra y ticks={-2,2},
	extra tick style={grid=major},
 legend style={at={(1.8,-0.2),anchor=north west},/tikz/every even column/.append style={column sep=0.5cm},font=\footnotesize},
    legend cell align=left,
    legend columns=2, 
]
 \addlegendimage{only marks,color=black,mark=star, thin};
    \addlegendentry{{ Group Mean ($\vec{\mu}_{\placeholderGroup}$)}};
    
    \addlegendimage{only marks,color=black,mark=triangle*};
    \addlegendentry{{ Identity ($\nu$)}};
    
    \addlegendimage{only marks,color=black,mark=*};
    \addlegendentry{{ Image ($\sample$)}};
    
    \addlegendimage{no markers,color=black,style=dotted};
    \addlegendentry{{ Group Covariance ($\vec{\Sigma}_{\placeholderGroup}$)}};
    
    \addlegendimage{no markers,color=black,style=dashed};
    \addlegendentry{{ Individual Covariance ($\beta\vec{I}$)}};
	\draw[red,dotted,thin] \pgfextra{
	  \pgfpathellipse{\pgfplotspointaxisxy{1}{1}}
		{\pgfplotspointaxisdirectionxy{0}{0.5}}
		{\pgfplotspointaxisdirectionxy{0.5}{0}}
	};
		\draw[red,dotted,thin] \pgfextra{
	  \pgfpathellipse{\pgfplotspointaxisxy{1}{1}}
		{\pgfplotspointaxisdirectionxy{0}{1}}
		{\pgfplotspointaxisdirectionxy{1}{0}}
	};
		\draw[red,dotted,thin] \pgfextra{
	  \pgfpathellipse{\pgfplotspointaxisxy{1}{1}}
		{\pgfplotspointaxisdirectionxy{0}{1.5}}
		{\pgfplotspointaxisdirectionxy{1.5}{0}}
	};
	\draw[blue,dotted,thin] \pgfextra{
	  \pgfpathellipse{\pgfplotspointaxisxy{-1}{-1}}
		{\pgfplotspointaxisdirectionxy{0}{0.2}}
		{\pgfplotspointaxisdirectionxy{0.2}{0}}
	};
	\draw[blue,dotted,thin] \pgfextra{
	  \pgfpathellipse{\pgfplotspointaxisxy{-1}{-1}}
		{\pgfplotspointaxisdirectionxy{0}{0.4}}
		{\pgfplotspointaxisdirectionxy{0.4}{0}}
	};
		\draw[blue,dotted,thin] \pgfextra{
	  \pgfpathellipse{\pgfplotspointaxisxy{-1}{-1}}
		{\pgfplotspointaxisdirectionxy{0}{0.6}}
		{\pgfplotspointaxisdirectionxy{0.6}{0}}
	};
	\addplot [only marks,mark=star,color=blue,mark size=2pt,thin] coordinates { (-1,-1) };

	\addplot [only marks,mark=triangle*,color=blue,mark size=0.5pt] coordinates { (-1.2,-1.2) };
	\addplot [only marks,mark=triangle*,color=blue,mark size=0.5pt] coordinates { (-1.3,-1.6) };
	\addplot [only marks,mark=triangle*,color=blue,mark size=0.5pt] coordinates { (-0.8,-1.0) };
	\addplot [only marks,mark=triangle*,color=blue,mark size=0.5pt] coordinates { (-1.1,-0.5) };
	\addplot [only marks,mark=triangle*,color=blue,mark size=0.5pt] coordinates { (-1.4,-0.6) };
	\addplot [only marks,mark=triangle*,color=blue,mark size=0.5pt] coordinates { (-1,-0.85) };
	\addplot [only marks,mark=triangle*,color=blue,mark size=0.5pt] coordinates { (-0.7,-0.7) };
	\addplot [only marks,mark=triangle*,color=blue,mark size=0.5pt] coordinates { (-0.7,-1.2) };
	\addplot [only marks,mark=triangle*,color=blue,mark size=0.5pt] coordinates { (-0.5,-0.6) };
	\addplot [only marks,mark=triangle*,color=blue,mark size=0.5pt] coordinates { (-1.3,-0.95) };
	
	\addplot [only marks,mark=star,color=red, mark size=2pt,thin] coordinates { (1,1) };
	
	\addplot [only marks,mark=triangle*,color=red,mark size=0.5pt] coordinates { (1.4,1.4) };
	\addplot [only marks,mark=triangle*,color=red,mark size=0.5pt] coordinates { (2.1,1.2) };
	\addplot [only marks,mark=triangle*,color=red,mark size=0.5pt] coordinates { (0.75,2.3) };
	\addplot [only marks,mark=triangle*,color=red,mark size=0.5pt] coordinates { (1.6,2.2) };
	\addplot [only marks,mark=triangle*,color=red,mark size=0.5pt] coordinates { (0.6,1.0) };
	\addplot [only marks,mark=triangle*,color=red,mark size=0.5pt] coordinates { (1.2,0) };
	\addplot [only marks,mark=triangle*,color=red,mark size=0.5pt] coordinates { (1.8,0.2) };
	\addplot [only marks,mark=triangle*,color=red,mark size=0.5pt] coordinates { (1,0.7) };
	\addplot [only marks,mark=triangle*,color=red,mark size=0.5pt] coordinates { (0.4,0.4) };
	\addplot [only marks,mark=triangle*,color=red,mark size=0.5pt] coordinates { (0.4,1.4) };
	\addplot [only marks,mark=triangle*,color=red,mark size=0.5pt] coordinates { (0,0.2) };
	\addplot [only marks,mark=triangle*,color=red,mark size=0.5pt] coordinates { (1.6,0.9) };
	\addplot [only marks,mark=triangle*,color=red,mark size=0.5pt] coordinates { (-0.3,0.65) };

\nextgroupplot[xmin=-3,   xmax=3,
	ymin=-3,   ymax=3,
	line width=1.0pt,
	extra x ticks={-1,1},
	extra y ticks={-2,2},
	extra tick style={grid=major},
]

	\draw[\makeTransparent{red},dotted,thin] \pgfextra{
	  \pgfpathellipse{\pgfplotspointaxisxy{1}{1}}
		{\pgfplotspointaxisdirectionxy{0}{0.5}}
		{\pgfplotspointaxisdirectionxy{0.5}{0}}
	};
		\draw[\makeTransparent{red},dotted,thin] \pgfextra{
	  \pgfpathellipse{\pgfplotspointaxisxy{1}{1}}
		{\pgfplotspointaxisdirectionxy{0}{1}}
		{\pgfplotspointaxisdirectionxy{1}{0}}
	};
		\draw[\makeTransparent{red},dotted,thin] \pgfextra{
	  \pgfpathellipse{\pgfplotspointaxisxy{1}{1}}
		{\pgfplotspointaxisdirectionxy{0}{1.5}}
		{\pgfplotspointaxisdirectionxy{1.5}{0}}
	};
	\draw[\makeTransparent{blue},dotted,thin] \pgfextra{
	  \pgfpathellipse{\pgfplotspointaxisxy{-1}{-1}}
		{\pgfplotspointaxisdirectionxy{0}{0.2}}
		{\pgfplotspointaxisdirectionxy{0.2}{0}}
	};
	\draw[\makeTransparent{blue},dotted,thin] \pgfextra{
	  \pgfpathellipse{\pgfplotspointaxisxy{-1}{-1}}
		{\pgfplotspointaxisdirectionxy{0}{0.4}}
		{\pgfplotspointaxisdirectionxy{0.4}{0}}
	};
		\draw[\makeTransparent{blue},dotted,thin] \pgfextra{
	  \pgfpathellipse{\pgfplotspointaxisxy{-1}{-1}}
		{\pgfplotspointaxisdirectionxy{0}{0.6}}
		{\pgfplotspointaxisdirectionxy{0.6}{0}}
	};
	\addplot [only marks,mark=star,color=blue,thin, mark size=1pt] coordinates { (-1,-1) };

	\addplot [only marks,mark=triangle*,color=\makeTransparent{blue},mark size=0.5pt] coordinates { (-1.2,-1.2) };
	
	\addplot [only marks,mark=triangle*,color=\makeTransparent{blue},mark size=0.5pt] coordinates { (-1.3,-1.6) };
	\addplot [only marks,mark=triangle*,color=\makeTransparent{blue},mark size=0.5pt] coordinates { (-0.8,-1.0) };
	\addplot [only marks,mark=triangle*,color=\makeTransparent{blue},mark size=0.5pt] coordinates { (-1.1,-0.5) };
	\addplot [only marks,mark=triangle*,color=\makeTransparent{blue},mark size=0.5pt] coordinates { (-1.4,-0.6) };
	\addplot [only marks,mark=triangle*,color=\makeTransparent{blue},mark size=0.5pt] coordinates { (-1,-0.85) };
	\addplot [only marks,mark=triangle*,color=\makeTransparent{blue},mark size=0.5pt] coordinates { (-0.7,-0.7) };
	\addplot [only marks,mark=triangle*,color=\makeTransparent{blue},mark size=0.5pt] coordinates { (-0.7,-1.2) };
	\addplot [only marks,mark=triangle*,color=blue,mark size=0.5pt] coordinates { (-0.5,-0.6) };
	\addplot [only marks,mark=triangle*,color=\makeTransparent{blue},mark size=0.5pt] coordinates { (-1.3,-0.95) };
	
	\draw[blue,dashed,thin] \pgfextra{
	  \pgfpathellipse{\pgfplotspointaxisxy{-0.5}{-0.6}}
		{\pgfplotspointaxisdirectionxy{0}{0.2}}
		{\pgfplotspointaxisdirectionxy{0.2}{0}}
	};
	\draw[blue,dashed,thin] \pgfextra{
	  \pgfpathellipse{\pgfplotspointaxisxy{-0.5}{-0.6}}
		{\pgfplotspointaxisdirectionxy{0}{0.4}}
		{\pgfplotspointaxisdirectionxy{0.4}{0}}
	};
	
	\addplot [only marks,mark=*,color=blue,mark size=0.5pt] coordinates { (-0.7,-0.5) };
	\addplot [only marks,mark=*,color=blue,mark size=0.5pt] coordinates { (-0.4,-0.4) };
	\addplot [only marks,mark=*,color=blue,mark size=0.5pt] coordinates { (-0.4,-0.8) };
	\addplot [only marks,mark=*,color=blue,mark size=0.5pt] coordinates { (-0.75,-0.7) };
	
	\addplot [only marks,mark=star,color=red,mark size=0.5pt,thin] coordinates { (1,1) };
	
	\addplot [only marks,mark=triangle*,color=\makeTransparent{red},mark size=0.5pt] coordinates { (1.4,1.4) };
	\addplot [only marks,mark=triangle*,color=\makeTransparent{red},mark size=0.5pt] coordinates { (2.1,1.2) };
	\addplot [only marks,mark=triangle*,color=\makeTransparent{red},mark size=0.5pt] coordinates { (0.75,2.3) };
	\addplot [only marks,mark=triangle*,color=\makeTransparent{red},mark size=0.5pt] coordinates { (1.6,2.2) };
	\addplot [only marks,mark=triangle*,color=\makeTransparent{red},mark size=0.5pt] coordinates { (0.6,1.0) };
	\addplot [only marks,mark=triangle*,color=\makeTransparent{red},mark size=0.5pt] coordinates { (1.2,0) };
	\addplot [only marks,mark=triangle*,color=\makeTransparent{red},mark size=0.5pt] coordinates { (1.8,0.2) };
	\addplot [only marks,mark=triangle*,color=\makeTransparent{red},mark size=0.5pt] coordinates { (1,0.7) };
	\addplot [only marks,mark=triangle*,color=\makeTransparent{red},mark size=0.5pt] coordinates { (0.4,0.4) };
	\addplot [only marks,mark=triangle*,color=\makeTransparent{red},mark size=0.5pt] coordinates { (0.4,1.4) };
	\addplot [only marks,mark=triangle*,color=\makeTransparent{red},mark size=0.5pt] coordinates { (0,0.2) };
	\addplot [only marks,mark=triangle*,color=\makeTransparent{red},mark size=0.5pt] coordinates { (1.6,0.9) };
	\addplot [only marks,mark=triangle*,color=red,mark size=0.5pt] coordinates { (-0.3,0.65) };
	
	\draw[red,dashed,thin] \pgfextra{
	  \pgfpathellipse{\pgfplotspointaxisxy{-0.3}{0.65}}
		{\pgfplotspointaxisdirectionxy{0}{0.2}}
		{\pgfplotspointaxisdirectionxy{0.2}{0}}
	};
	\draw[red,dashed,thin] \pgfextra{
	  \pgfpathellipse{\pgfplotspointaxisxy{-0.3}{0.65}}
		{\pgfplotspointaxisdirectionxy{0}{0.4}}
		{\pgfplotspointaxisdirectionxy{0.4}{0}}
	};
	\addplot [only marks,mark=*,color=red,mark size=0.5pt] coordinates { (-0.5,0.55) };
	\addplot [only marks,mark=*,color=red,mark size=0.5pt] coordinates { (-0.2,0.45) };
	\addplot [only marks,mark=*,color=red,mark size=0.5pt] coordinates { (-0.2,0.85) };
	\addplot [only marks,mark=*,color=red,mark size=0.5pt] coordinates { (-0.55,0.75) };
	

\end{groupplot}
\end{tikzpicture}

    \caption{\Copy{copy:analytical_model_depiction}{The hierarchical Gaussian distribution models patterns we observe with respect to faces in the embedding space. In the left plot, identities are drawn. Identities, $\vec{\mu_{\groupA}}$ and $\vec{\mu_{\groupB}}$ are depicted with a red star and a blue star, respectively. Identities $\vec{\nu}_{\placeholderGroup,i}$ in group $\placeholderGroup$ are sampled from $\normalDist\left(\vec{\mu}_{\group},\vec{\sigma}_{\group}\right)$. Identities are depicted as triangles. In the right plot, images are drawn. These images $\sample_{\placeholderGroup,i,j}$, depicted as circles, are drawn from $\normalDist\left(\vec{\nu}_{\placeholderGroup,i},\beta\vec{I}\right)$. }}
    \label{fig:analyticalModel}
\end{figure}
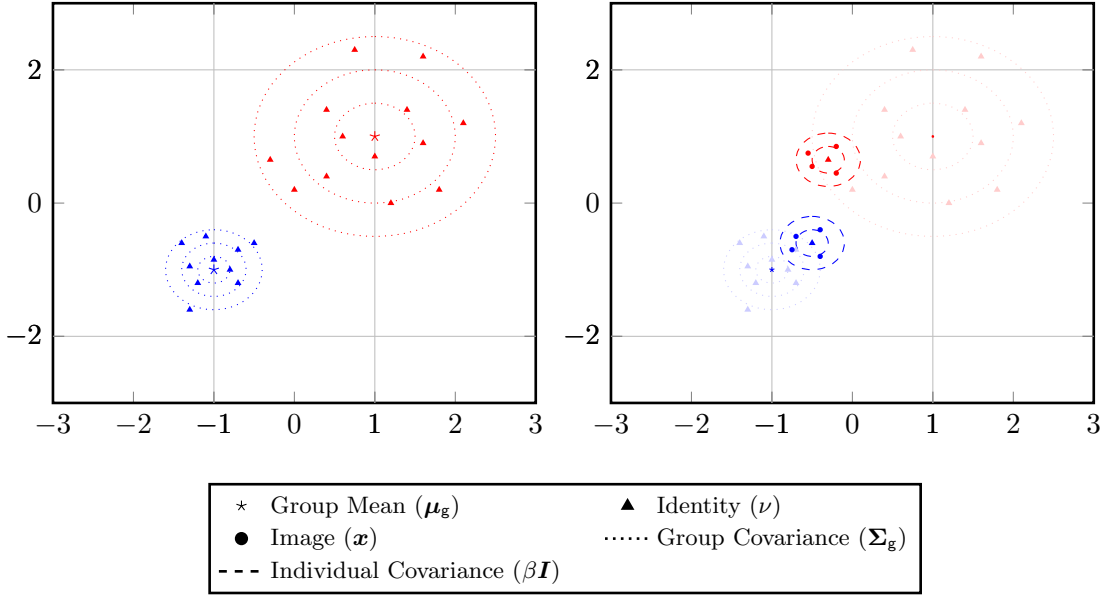

    


\section{Fairness on Face Obfuscation Systems}
\label{sec:experiments}
\definecolor{Gray}{gray}{0.9}
\newcolumntype{g}{>{\columncolor{Gray}}c}

Utilizing the experimental procedure from \cref{subsec:ExperimentalSetup}, we answer the three questions appearing at the beginning of \cref{sec:methodology}. Our findings highlight demographic disparities in face obfuscation systems. In particular, we show that it is easier to impersonate identities in the same demographic than it is to impersonate identities in a different demographic.

\subsection{Error Disparity Among Groups} 
\label{subsec:error}
Face recognition systems are known to be biased with respect to population demographics. We show such demographic disparities exist in the embedding space, and can be traced through the learning process of the embedding network. 

\begin{figure}[t]
     \centering
     \begin{subfigure}[b]{0.3\textwidth}
         \centering
         \includegraphics[width=\textwidth]{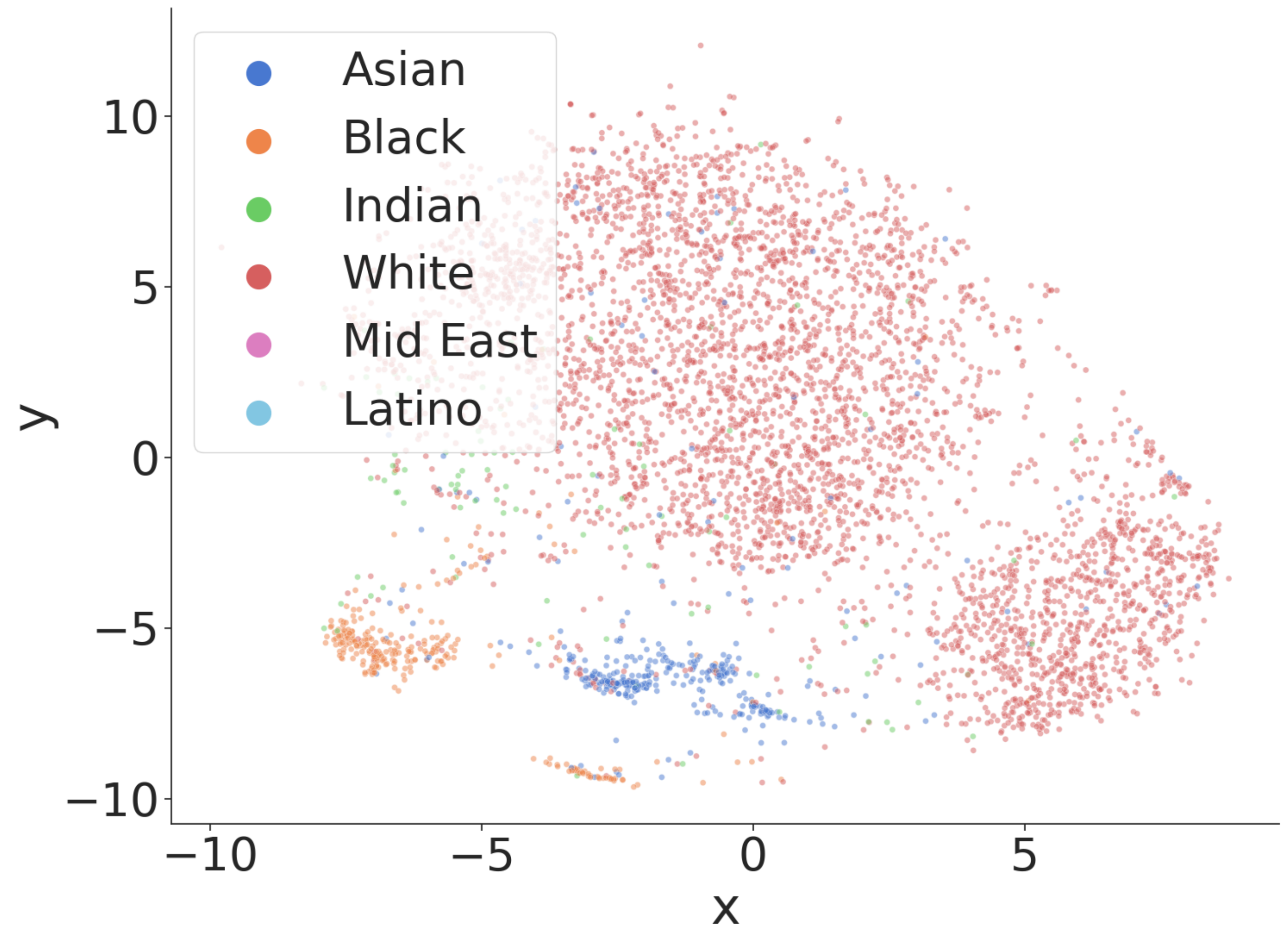}
         \caption{LFW - Race}
         \label{fig:tsne_lfw_race}
     \end{subfigure}
     \begin{subfigure}[b]{0.3\textwidth}
         \centering
         \includegraphics[width=\textwidth]{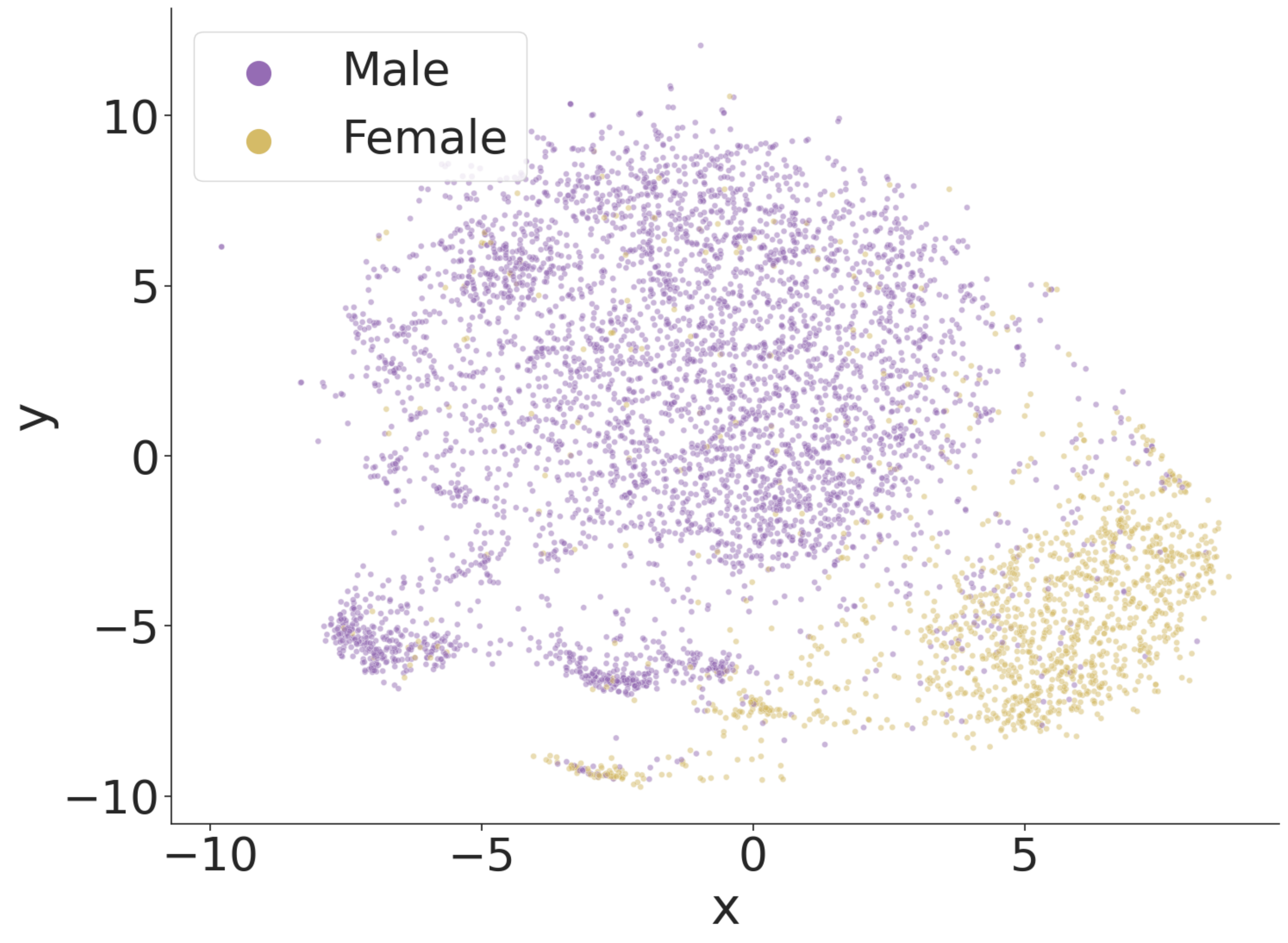}
         \caption{LFW - Sex}
         \label{fig:tsne_lfw_sex}
     \end{subfigure}
     \caption{t-SNE~\cite{maaten2008tsne} of LFW embeddings generated by FaceNet. 
     }
     \label{fig:tsne}
\end{figure}

\smallskip
 \noindent
\textbf{Face Recognition Empirical Performance:}
\Cref{table:k_fold_confusion} demonstrates the demographic disparities in face recognition. 
 When evaluating matching performance $\mathrm{TPR}_{0.001}$ on the pretrained FaceNet model, pairs of identities selected within the same demographic group perform worse when compared to pairs selected without any such demographic restriction (\cref{table:k_fold_confusion}). Because identities in the same demographic group are closer together, as seen in \cref{fig:tsne}, false accepts are more likely thereby lowering $\mathrm{TPR}_{0.001}$. This idea also explains demographic disparities in matching performance. With the exception of the Indian demographic group, which is too small to source any significant conclusions, minority groups have lower matching performance. This is attributed to the tightly clustered embeddings we observe for minority demographic groups.


\begin{table}[t]

\begin{center}
  \begin{tabular}{g gggggg}
    \toprule
    \multicolumn{7}{c}{\textbf{Pre-trained Facenet}}\\
    \midrule
    $\mathrm{TPR}_{0.001}$ & .9618 & .8516 & .9594 & .8536 & .9242 & 1.000 \\
    AUC & .9994 & .9977 & .9996 & .9981 & .9995 & 1.000 \\
    \midrule
    \rowcolor{white}
    $\mathrm{TPR}_{0.001}$ & .9678 & .9436 & .9732 & .9448 & .9742 & 1.000 \\
    \rowcolor{white}
    AUC & .9995 & .9988 & .9993 & .9998 & .9999 & 1.000 \\
    \midrule
    \rowcolor{white}
    $N$ & 10000 & 5000 & 10000 & 2500 & 1240 & 20 \\
    \midrule
    \rowcolor{white}
    {} & {\bf Male} & {\bf Female} & {\bf White} & {\bf Asian} & {\bf Black} & {\bf Indian} \\ 
    \bottomrule
    \end{tabular}
    \vspace{-0.15in}
    \phantom{\begin{tabular}{g gggggg}
    \hline
    \multicolumn{7}{c}{\textbf{Fairness Promoting Adversarial Training}}
    \end{tabular}}
\fbox{\begin{tabular}{llll}
\textcolor{Gray}{$\blacksquare$} & Same Demographic & $\square$ & Any Demographic
\end{tabular}
} 
\end{center}

\caption{Matching accuracy on LFW embeddings generated by FaceNet. Pairs within only the same demographic have lower accuracy to pairs matching any demographic. $N$: number of pairs.}
\label{table:k_fold_confusion}
\end{table}

\smallskip
 \noindent
 \textbf{Intuition from Analytical Model:}
 Understanding why embedding networks have disparate demographic-wise behavior is intractable given the state of current literature in neural network analysis, so we turn to our analytical model and PCA for intuition. Given our interest in studying the impact of the frequency of each group in a training set on the efficacy of the learned embedding network, we formulate a proposition which relates a relative projection distance, our experiments, and our analytical model:

\begin{proposition}\label{prop:relProjectionDistance}
    For fixed $\vec{\mu}_{\groupA}$ and fixed $\vec{\Sigma}_{\groupA}$, as $\gamma$ approaches $0$, the relative projection distance (defined in \cref{eq:relative_proj_dist}) of examples in group $\groupA$  increases.
\end{proposition}

This proposition suggests minority demographic groups are poorly represented by metric embedding networks. The magnitude of performance disparities can be explained by $\gamma$, the parameter capturing demographic imbalance in sampling. For smaller values of $\gamma$, the sampled distribution does not represent the natural distribution of the minority group. For the metric embedding network, such poor representation leads the embedding network to have trouble discerning identities in the minority demographic group. Hence, we see more errors and ease of impersonation for members of the minority group. \Cref{prop:relProjectionDistance} is discussed further in \cref{subsec:pcaAdversarialAttack}.\smallskip

\noindent
\textbf{What the Metric Embedding Network Learns:} \label{subsec:learning}
To understand \emph{what} the metric embedding network architecture learns, we plot the resulting embedding structure of FaceNet using t-SNE~\cite{maaten2008tsne} in \cref{fig:tsne}. Embedding networks trained without explicit demographic information can discern demographic groups in the embedding space. For each demographic group, separate clusters appear within the embedding space. The male and female clusters appear almost linearly separable.

\begin{figure}[t]
     \centering
     \begin{subfigure}[b]{0.3\textwidth}
         \centering
         \includegraphics[width=\textwidth]{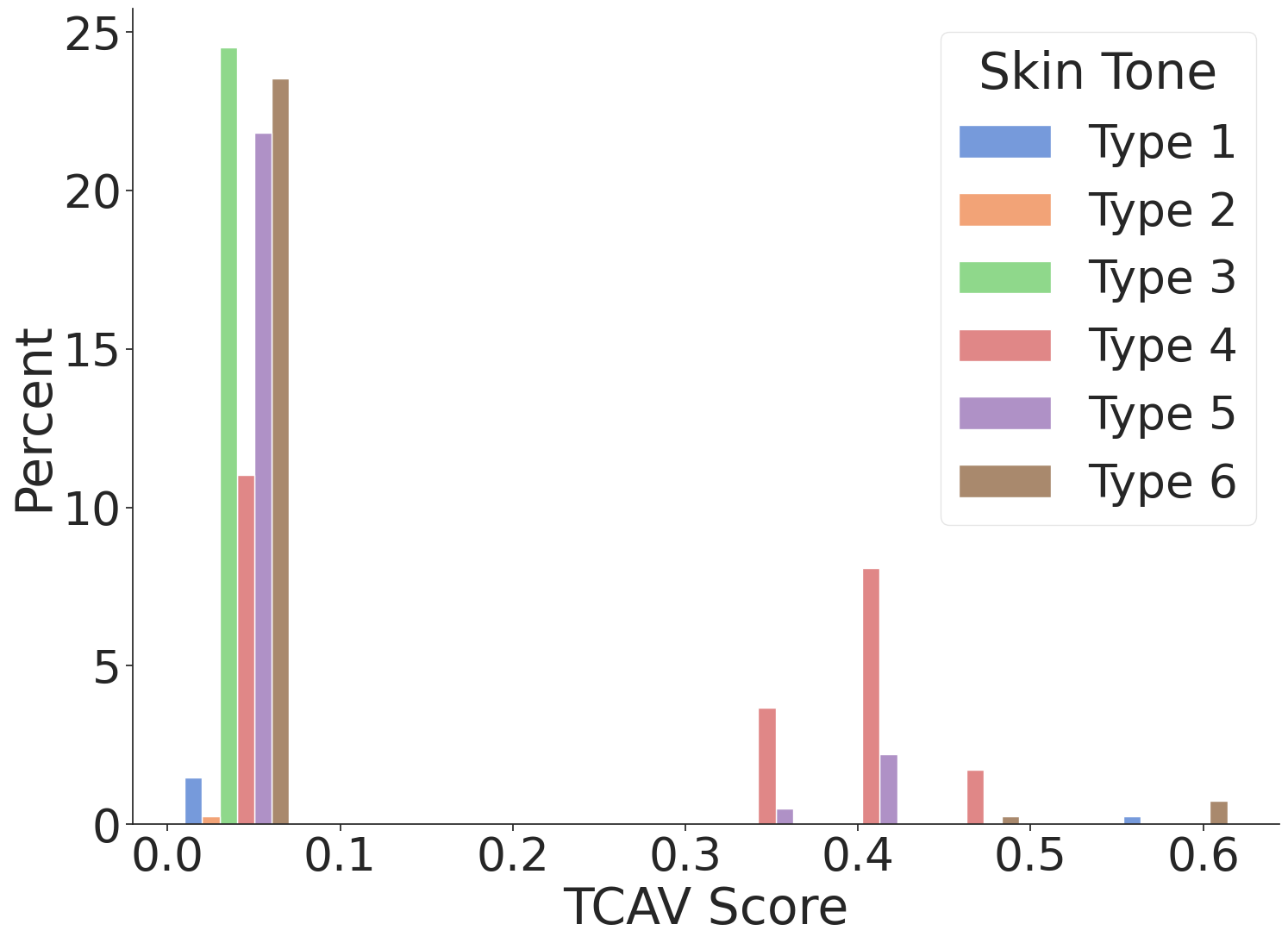}
         \caption{\Copy{copy:TCAV_2_subcaption}{Block 35 Activation 3}}
         \label{fig:tcav_2}
     \end{subfigure} 
     \begin{subfigure}[b]{0.3\textwidth}
         \centering
         \includegraphics[width=\textwidth]{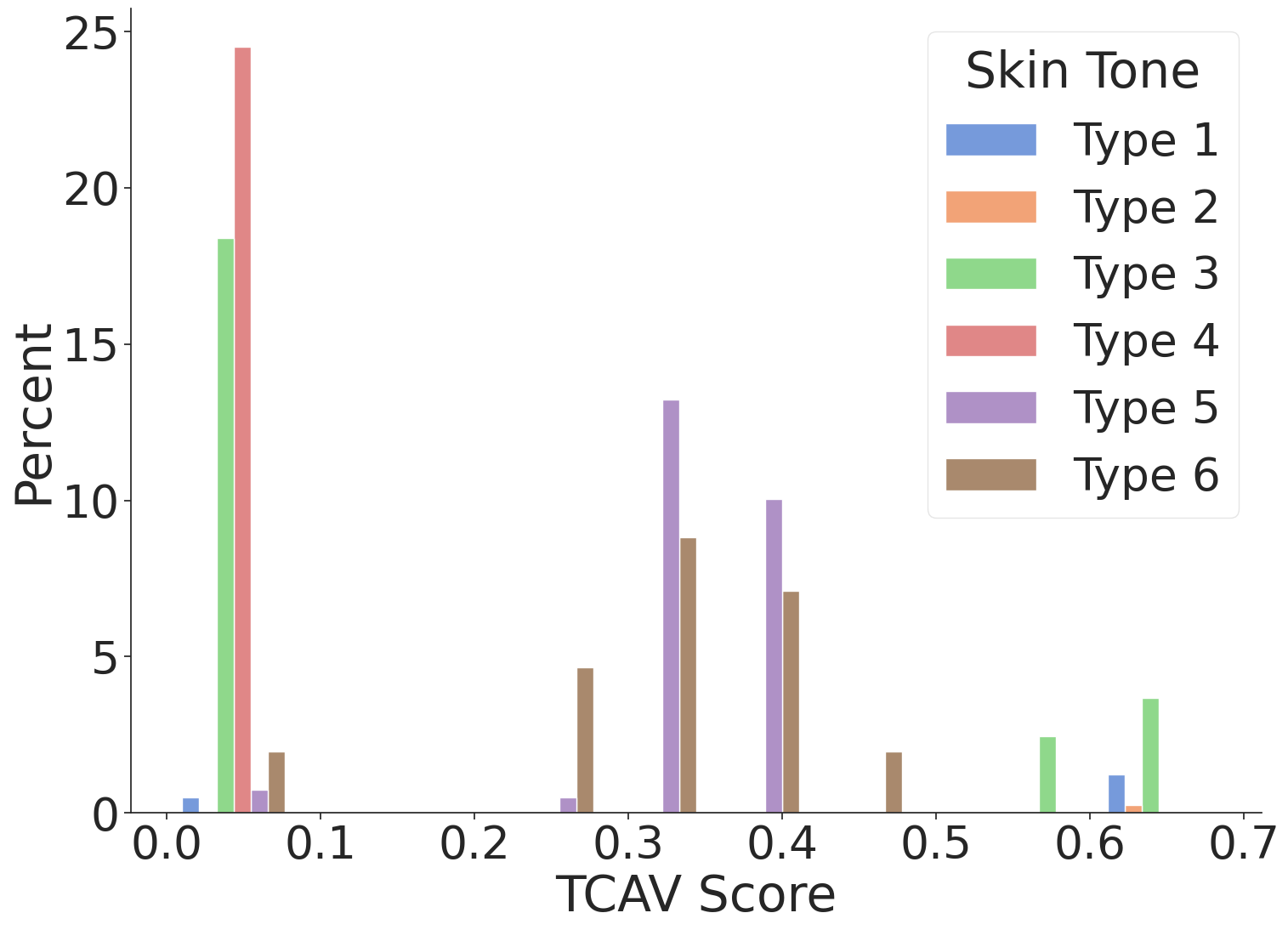}
         \caption{Conv2d 2a 3x3 Activation}
         \label{fig:tcav_4}
     \end{subfigure}
     \caption{\Copy{copy:tcav_figure_caption}{TCAV score distribution of 408 identities in VGGFace2. 
     \Cref{fig:tcav_2,fig:tcav_4} learn concepts for lighter and darker skin tones, respectively. }}
     \label{fig:tcav}
\end{figure}

\smallskip
\noindent
\textbf{Where the Metric Embedding Network Learns:}
Given the demographic-wise clustering behavior we see in \cref{fig:tsne}, we determine the degree to which the network itself contributes to observed clustering behavior. We utilize TCAV~\cite{kim18tcav} to investigate if intermediate layers of the network learn to distinguish demographic attributes. As it was originally defined for classification networks, we retrofit the TCAV framework to metric embedding networks; we associate each identity with an anchor embedding, corresponding to its embedding centroid. We then use the $\ell_2$ distance between the embedding of an input face and its anchor embedding to estimate the gradient from the output layer to the relevant activation layer.

We use skin tones as the demographic concepts. We annotate the skin tones with the Fitzpatrick scale~\cite{skin_tones}, depicted in \cref{table:Fitzpatrick}. At each activation layer, we train linear classifiers that distinguish the layer activation according each skin tone. Each classifier is trained on 6 skin tones, each containing 75 images. The TCAV score associated with each identity is the proportion of images for which the dot product of this linear classifier and gradient is positive. Nonzero TCAV scores indicate a concept heavily utilized in embedding construction.

Our evaluation involves annotated subsets of the VGGFace2 dataset, one containing 408 identities and another with 4102 identities, with each identity containing 100 images of a person's face. The subset with 408 identities contains a balanced subset of skin tones, whereas the subset with 4102 mainly consists of faces with type 4 and type 5 skin tones. Only the TCAV scores with exactly matching skin-tone annotations are reported in \cref{fig:tcav}. We see high utilization of the skin tone concepts by two early layers in the network. We see that darker skin tones are learned separately (Conv2d 2a) than lighter skin tones (Block 35). This suggests metric embedding networks differentiate between skin tones in early layers.


\subsection{Effectiveness of Obfuscation}\label{subsec:obfuscationEffectiveness}

After characterizing the baseline disparities in face recognition systems, we study the subsequent disparities in face obfuscation. We assess, for each demographic group,  the difficulty with which identities may be impersonated. We show embedding space geometry  induces demographic disparities in the perturbation norm, $\norm{\perturbation}_2$, necessary to successfully impersonate an identity. Further, we study the demographic disparities in the obfuscation success rates. These disparities are present in both the white-box and black-box settings.  
\smallskip

\begin{figure}[t]
     \centering
     \begin{subfigure}{0.3\columnwidth}
         \centering
         \includegraphics[width=\textwidth]{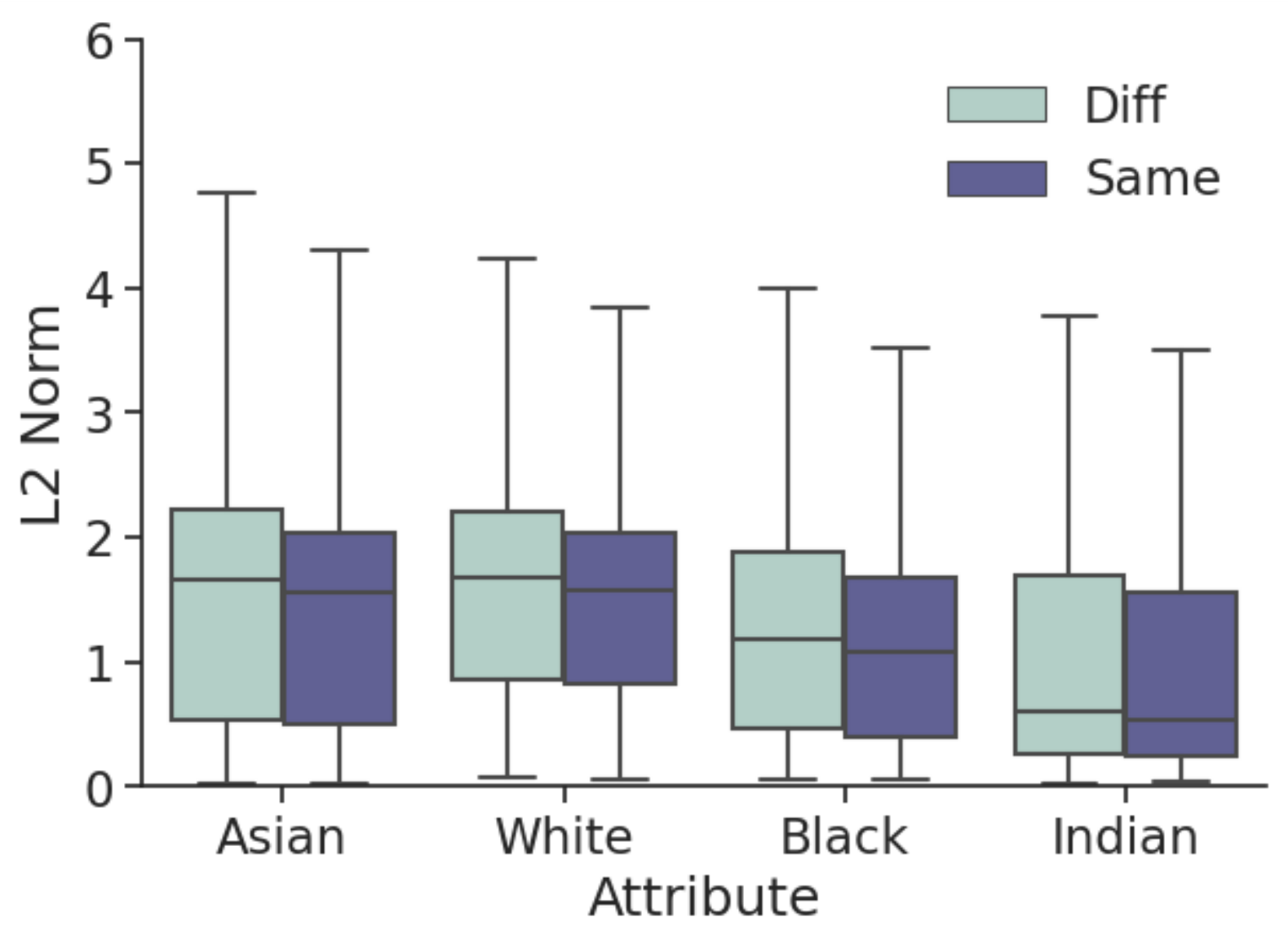}
         \caption{$\ell_2$ Norm: Race}
         \label{fig:cdf_race}
     \end{subfigure}
     \begin{subfigure}{0.3\columnwidth}
         \centering
         \includegraphics[width=\textwidth]{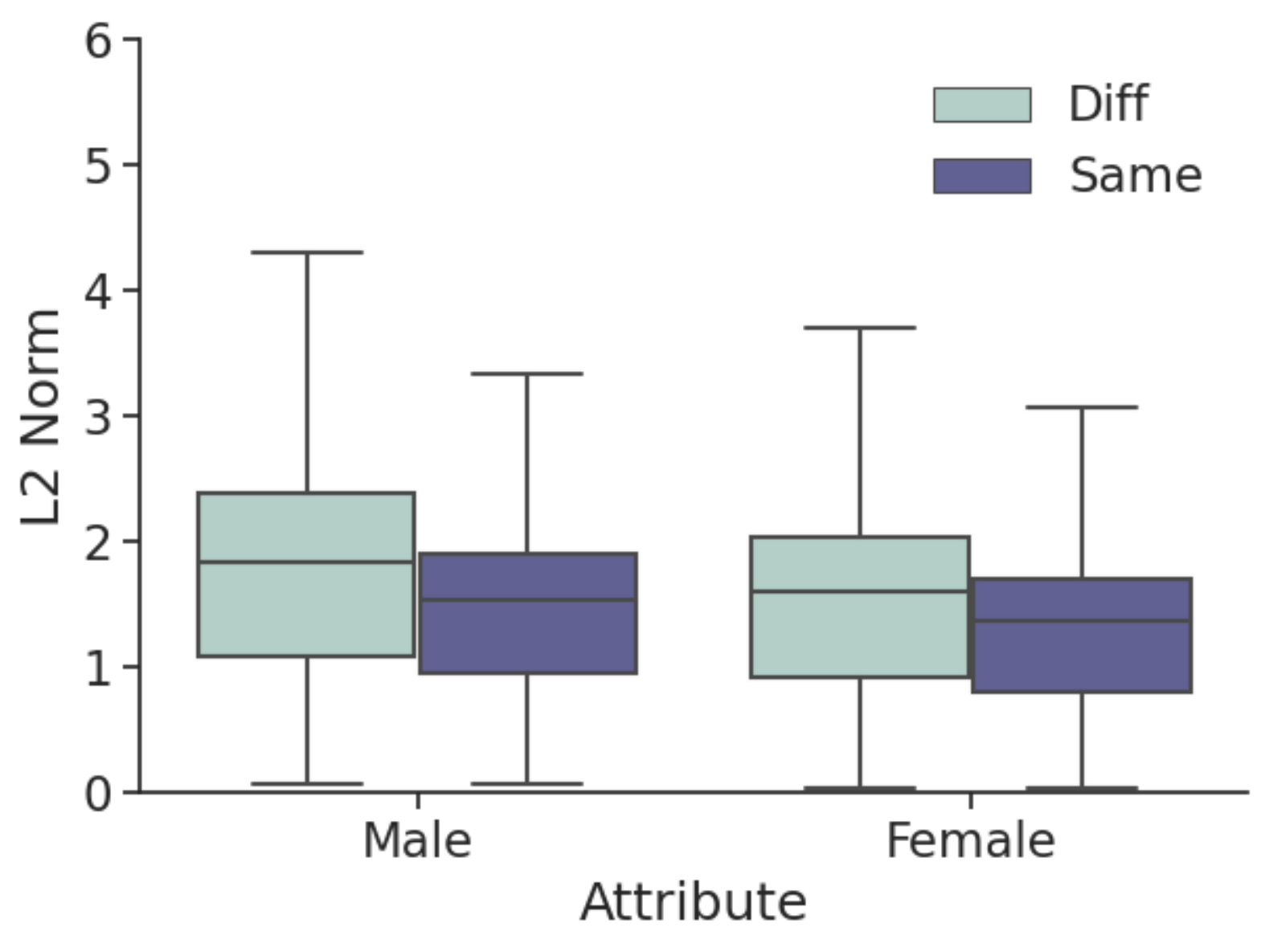}
         \caption{$\ell_2$ Norm: Sex}
         \label{fig:cdf_sex}
     \end{subfigure}
     \caption{Distribution of perturbation norms generated by CW~\cite{carlini2017towards} on LFW. 
    }
     \label{fig:cdf_pert_norm}
\end{figure}

\noindent
\textbf{Perturbation Norms:}
To detect disparities in the difficulty of face obfuscation, we examine the norms of perturbations $\norm{\perturbation}_2$, needed to successfully impersonate an identity. Using the adversarial examples generated by targeted attacks, we depict in \cref{fig:cdf_pert_norm} the distribution of perturbation norms conditioned by demographic. We observe that the perturbation strength necessary to successfully impersonate an identity depends on demographic. Further, we observe that for each demographic group, the strength of the perturbation necessary to impersonate an identity is larger if the target identity is in a demographic different from the source identity. We put forth \cref{null:demographicPerturbation} which formally addresses demographic disparities in perturbation strength. 

\begin{nullHyp} \label{null:demographicPerturbation}
    \item For each demographic group, the mean perturbation $\ell_2$ norm $\norm{\perturbation}_2$ necessary for targeted obfuscation is identical.
\end{nullHyp}

A formal analysis rejects \cref{null:demographicPerturbation}: Differences in perturbation strength for impersonation of identities in the same and different sex demographic group, and the same and different race demographic group are statistically significant: $p$-values do not exceed $3.34 \times 10^{-28}$. Furthermore, there is a statistically significant difference between perturbations targeted within the same demographic and perturbations targeted outside the demographic for the Male, Female, Asian and White  population groups with $p$-values of $6.61 \times 10^{-24}$, $2.00 \times 10^{-26}$, $0.00110$, and $0.00180$, respectively. These are the largest four demographic groups in our dataset. Compared to the Black and Indian minority groups, we expect a more significant difference in perturbation strength necessary to impersonate identities in the same demographic and that which is needed to impersonate identities in different demographics.

\smallskip

\begin{figure}[t]
     \centering
     \begin{subfigure}{0.6\columnwidth}
         \centering
         \includegraphics[width=\textwidth]{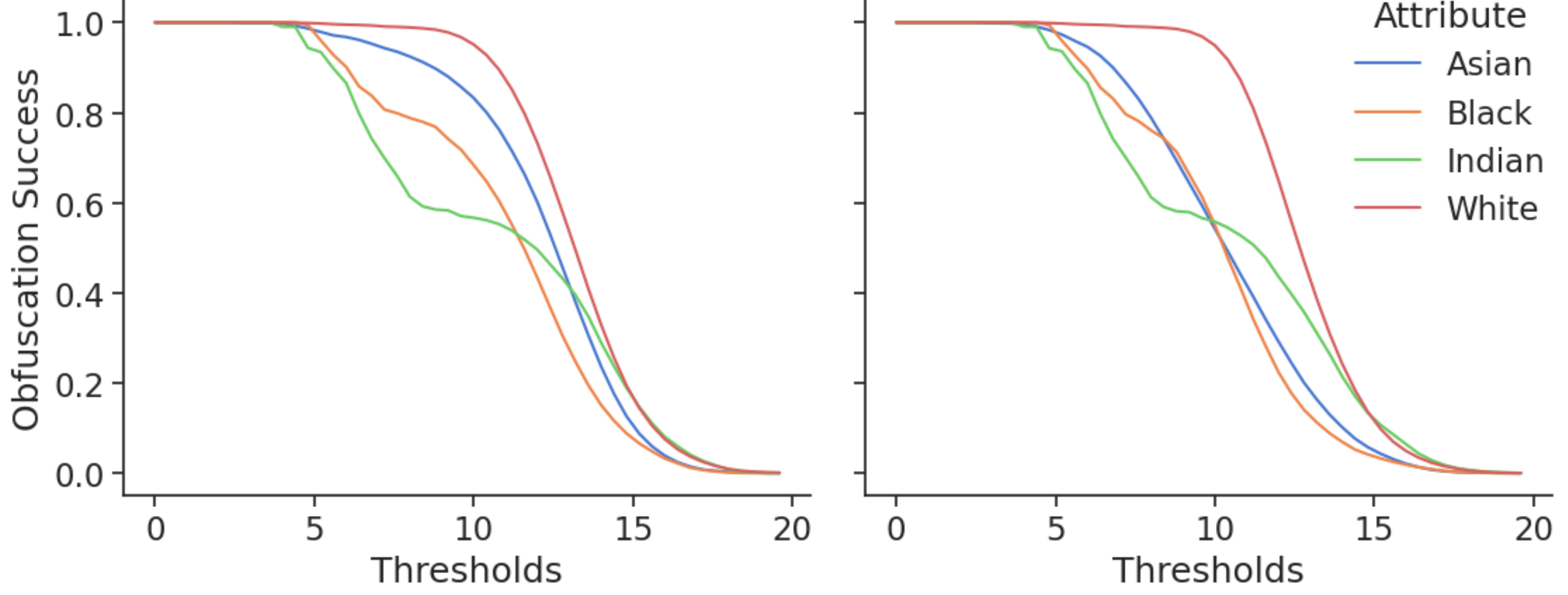}
         \caption{Targeted: Different Race (Left) vs. Same Race (Right)}
         \label{fig:targeted_race}
     \end{subfigure}
     \begin{subfigure}{0.6\columnwidth}
         \centering
         \includegraphics[width=\textwidth]{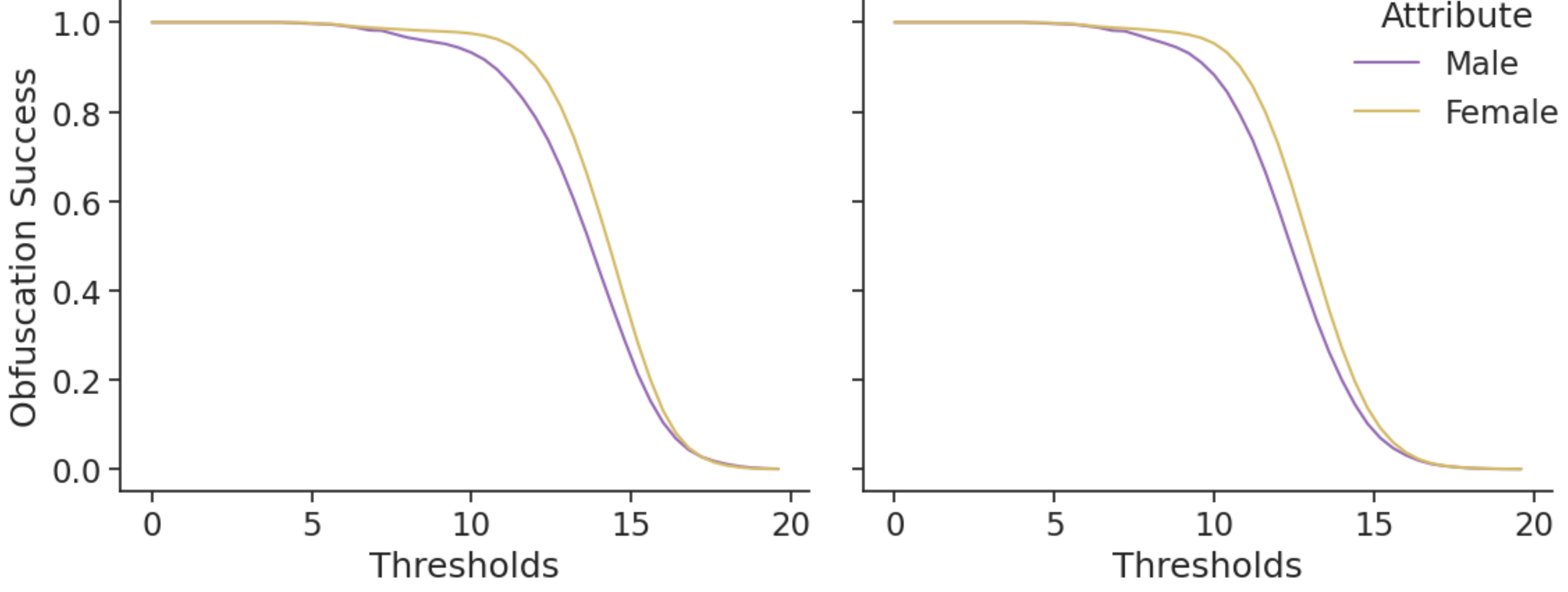}
         \caption{Targeted: Different Sex (Left) vs. Same Sex (Right)}
         \label{fig:targeted_sex}
     \end{subfigure}
     \caption{Targeted obfuscation success on FaceNet in a white-box setting. }
     \label{fig:adversarial_success}
\end{figure}

\begin{figure}[t]
     \centering
     \begin{subfigure}{0.48\columnwidth}
         \centering
         \includegraphics[width=\textwidth]{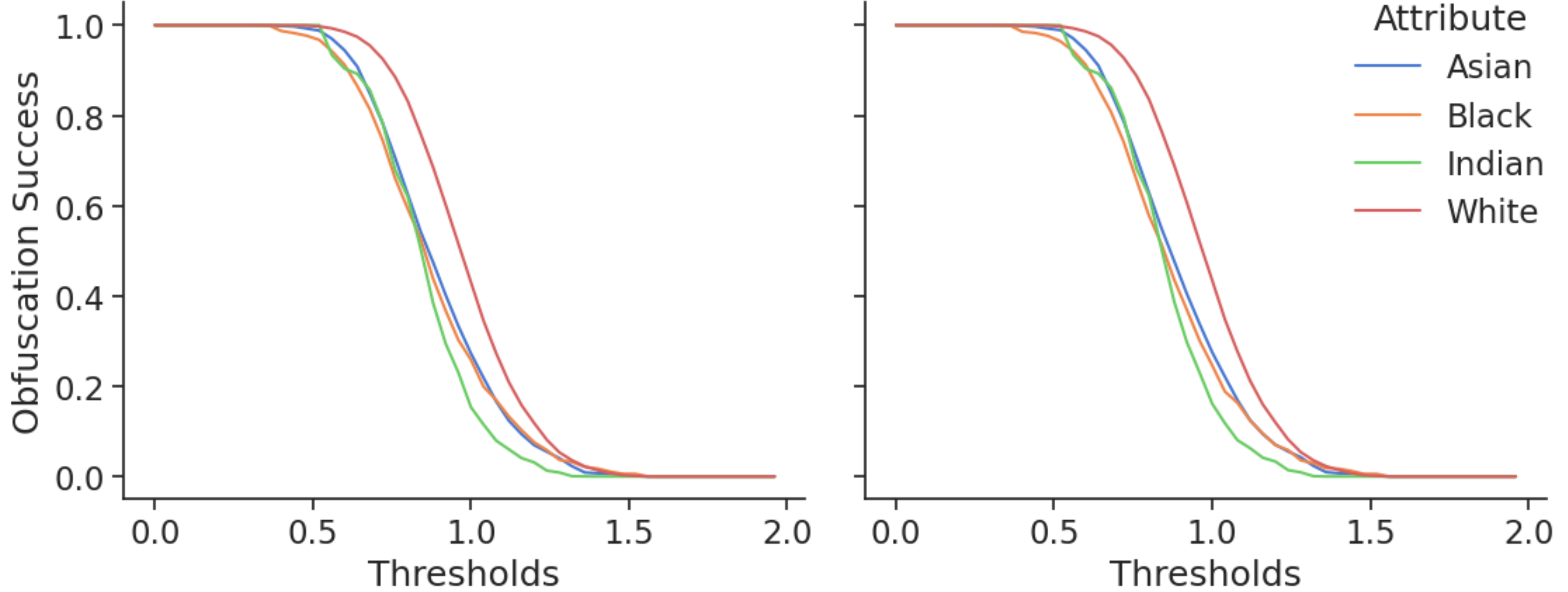}
         \caption{OpenFace: Different Race (Left) vs. Same Race (Right)}
         \label{fig:openface_race}
     \end{subfigure}
     
     \begin{subfigure}{0.48\columnwidth}
         \centering
         \includegraphics[width=\textwidth]{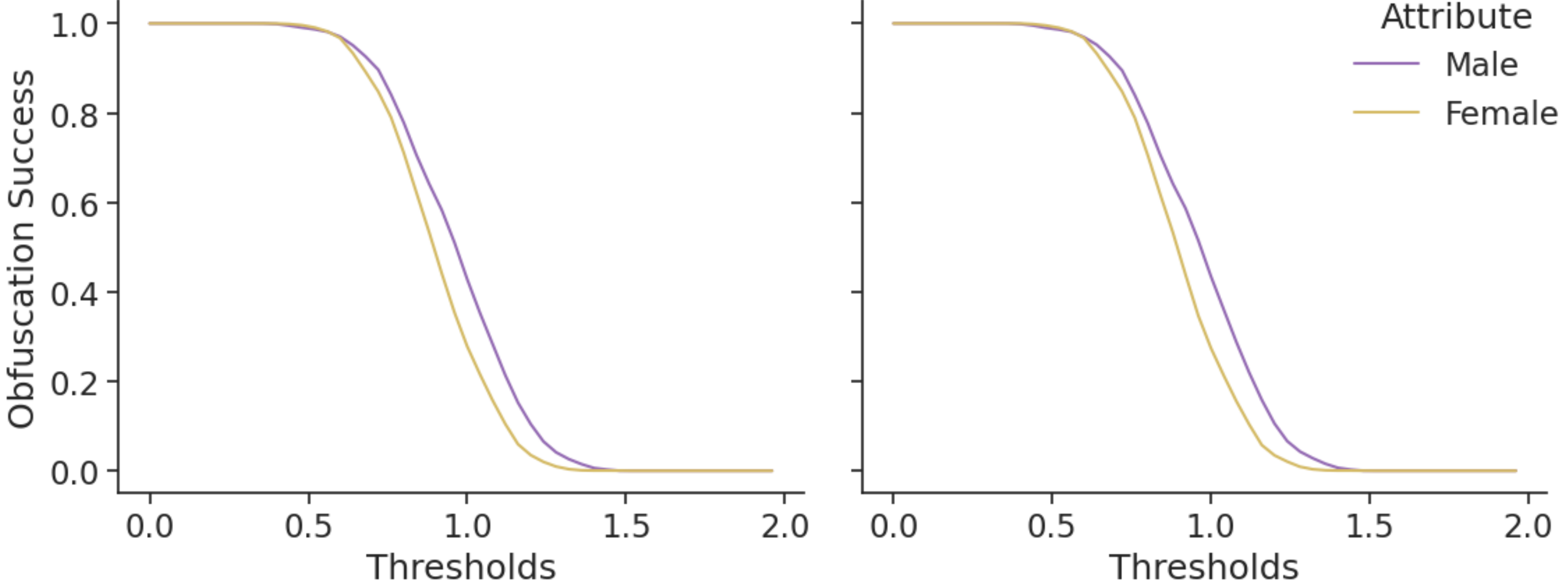}
         \caption{OpenFace: Different Sex (Left) vs. Same Sex (Right)}
         \label{fig:openface_sex}
     \end{subfigure}
      \caption{Targeted obfuscation success on OpenFace in a black-box setting.}
     \label{fig:transferability_success}
\end{figure}

\noindent
\textbf{Obfuscation Success:}
\Cref{fig:adversarial_success} shows that targeted perturbations enjoy success rate dependent on the demographic attribute. Faces with the White race attribute exhibit a higher obfuscation success rate. These results also hold in the black box settings: similar trends exist for embeddings generated on OpenFace, Deepface, ArcFace, and VGGFace models. Success rates for the OpenFace model are shown in \cref{fig:transferability_success}, and success rates for the remaining black box models may be found in \cref{fig:transferability_success_appendix} in \cref{subsec:Black-box_appendix}.  We observe examples generated on faces from the majority group transfer better than those of minority groups. We conjecture that the larger perturbation norms of such faces contribute to improved transferability rates. This observation is consistent with an observation from Face-off~\cite{DBLP:FaceOff}; increasing the norm of perturbations improved transferability to black-box models.

\begin{figure}[t]
\centering
     \begin{subfigure}{0.3\columnwidth}
         \centering
         \includegraphics[width=\textwidth]{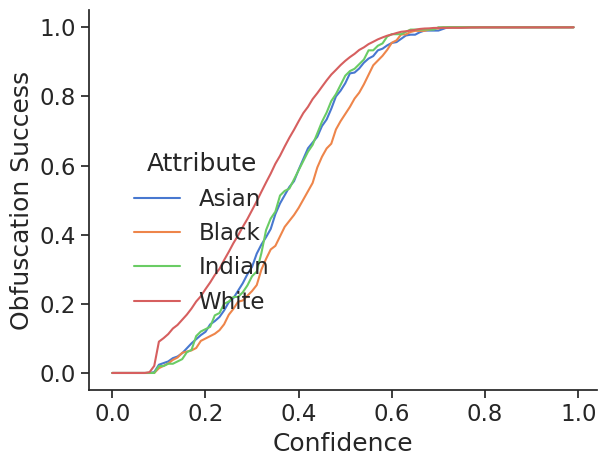}
         \caption{Untargeted Azure: Race}
         \label{fig:azure-lowkey-race}
     \end{subfigure}
     \begin{subfigure}{0.3\columnwidth}
         \centering
         \includegraphics[width=\textwidth]{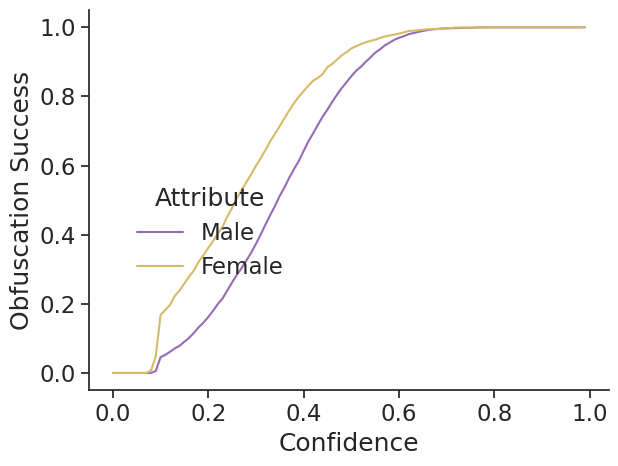}
         \caption{Untargeted Azure: Sex}
         \label{fig:azure-lowkey-sex}
     \end{subfigure}
     \caption{Untargeted obfuscation success on Microsoft Azure Face API.}
     \label{fig:azure-lowkey}
\end{figure}

In addition to offline models, this trend also holds for commercially available online face recognition APIs. We test the success of the perturbed faces against the Face++, Amazon AWS Rekognition, and Microsoft Azure Face APIs. In \cref{fig:azure-lowkey}, we present the CCDF of confidence scores for untargeted examples generated tested by Azure. For race, on a fixed obfuscation success rate we observe a generally higher confidence in impersonation for White faces.  In contrast, Black faces see a generally lower confidence in impersonation for that same obfuscation success rate.  Interestingly, for the sex attribute, we generally observe a slightly higher confidence for females. By examining the distance between an embedding and its target, we gain insight into the factor which directly controls obfuscation success rate. We put forth a null hypothesis to formally test our observations.

\begin{nullHyp}
    Across all source demographic groups, the distribution of distances between obfuscated embeddings and their targets is identical.
\end{nullHyp}

This null hypothesis is easily rejected:  Differences in distance between face embeddings and targets are statistically significant regardless of target demographic: The $p$-values do not exceed  $0.0395$. These results confirm the bias in obfuscation performance previously identified. From a practical perspective, our results may suggest that users should select an identity to impersonate of the same race or sex in order to optimize face obfuscation system utility. This is counterproductive to the user's privacy for two reasons: 1) adversarial perturbations with smaller $\ell_p$ norms will struggle in transferring to other models and 2) a user may leak demographic information to an adversary. \smallskip

\smallskip
\noindent
\textbf{Intuition from Analytical Model:}\label{subsubsec:analysisEffectivenessObfuscation}
From \cref{fig:cdf_pert_norm}, we observe demographic disparities in the strength of perturbation necessary to successfully obfuscate an image. We also showed it is generally easier to impersonate identities in the same demographic group than it is to impersonate identities in different demographic groups. This translated into the disparate success rates observed in \cref{fig:adversarial_success,fig:azure-lowkey,fig:transferability_success}. We use our analytical model to provide some mathematical intuition on this phenomenon.

Given an example image $\sample$ in group $\placeholderGroup$, we study how difficult it is to construct a perturbation $\perturbation$ such that $\sample + \perturbation$ successfully impersonates an identity outside of group $\placeholderGroup$. The analytical model is a natural medium in which quantifying the necessary strength of such a perturbation $\perturbation$ may occur.

For the sake of simplicity, consider the $1$-PCA embedding function $\embeddingFunc_1 : \realNumbers^{\dimension} \to \realNumbers$. Let sample $\sample$ be drawn from the overall synthetic data distribution $\sampleDist$ and without loss of generality assume this $\sample$ is a member of group $\groupA$. Denote by $\PDF_{\placeholderDist}$ the probability density function for a probability distribution $\placeholderDist$. We will assume group $\groupA$ is the minority group and so we assume that $\gamma \leq 1$, $\PDF_{\sampleDist_{\groupA}}[\vec{\mu}_{\groupA}] > \PDF_{\sampleDist_{\groupB}}[\vec{\mu}_{\groupA}]$, and $\PDF_{\sampleDist_{\groupB}}[\vec{\mu}_{\groupB}] > \PDF_{\sampleDist_{\groupA}}[\vec{\mu}_{\groupB}] $.  Given a perturbation $\perturbation$, we assume that perturbation $\perturbation$ is norm-bounded and in the direction of group $\groupB$, i.e.  $\frac{(\vec{\mu}_{\groupB}-\sample)}{\norm{\vec{\mu}_{\groupB}-\sample}_2}$. That is, we assume $\norm{\perturbation}_2 \leq \epsilon$ where $\epsilon$ is a non-negative real number.  
 We quantify the values of $\epsilon$ for which the following optimization problem is infeasible, thereby guaranteeing $\sample + \perturbation$  impersonates\footnote{Here, we assume it is possible for an example to impersonate itself.} an identity in group $\groupA$:

{\small
\begin{align}
\min_{\perturbation : \norm{\perturbation}_2 \leq \epsilon} \quad &  \norm{\perturbation}_2  & \mathrm{s.t.} \quad& \Pr_{\sample \drawnFrom \sampleDist_{\groupB}}\left[\embeddingFunc(\sample + \perturbation)\right] >  \Pr_{\sample \drawnFrom\sampleDist_{\groupA}}\left[\embeddingFunc(\sample + \perturbation)\right] \\
& &\quad \perturbation &= \eta \cdot  \frac{(\vec{\mu}_{\groupB}-\sample)}{\norm{\vec{\mu}_{\groupB}-\sample}_2} \text{ where } \eta \in \realNumbers
 \label{eq:theoryOptimizationProblem}
\end{align}
}

For notational compactness, we denote:
{\small
\begin{align*}
a=\embeddingFunc(\sample) \; \; \text{and} \; \;
b=\embeddingFunc\left(\frac{(\vec{\mu}_{\groupB}-\sample)}{\norm{\vec{\mu}_{\groupB}-\sample}_2}\right).
\end{align*}
}
This optimization objective in \cref{eq:theoryOptimizationProblem} guaranteed to be infeasible when: 
{\small
\begin{equation}
    \epsilon < \max\left\{0,\frac{2b(\gamma-1)\sqrt{\frac{a^2\gamma}{b^2(\gamma-1)^2}} +a\gamma +a + \embeddingFunc(\vec{\mu}_{\groupB})(1-\gamma)}{b(\gamma-1)}\right\} \label{eq:theoryCondition}
\end{equation}
}
This result is further detailed in \cref{sec:pca_math}.

Therefore we conclude that $\sample + \perturbation$ impersonates an identity in $\groupA$ when inequality \eqref{eq:theoryCondition} holds.
Furthermore, note that the bound in inequality \eqref{eq:theoryCondition} is not tight. The bound loosens as $\gamma$ approaches $0$. Within inequality \eqref{eq:theoryCondition}, we notice that for a fixed $\vec{\Sigma}_{\groupA}$, $\vec{\mu}_{\groupA}$, and $\sample$ which impersonates an identity in group $\groupA$, as $\gamma$ approaches $0$, the set of perturbations $\perturbation$ for which $\sample+\perturbation$ still impersonates an identity in group $\groupA$, decreases in size.

Relating to experiments in this section on existing face recognition datasets, it is the disparities in sampling which affect the strength of the perturbation necessary to impersonate an identity in a different demographic group. These disparities are captured by $\gamma$. This analysis agrees with perturbation norms in \cref{fig:cdf_pert_norm}: impersonating  an identity in a different demographic is more difficult than impersonating an identity in the same demographic group. \smallskip

\noindent
\textbf{Stability Properties of Face Obfuscation: }
\label{subsec:stability}
\Copy{copy:Pre_Lipschitz_geometry}{We further study the impact of $\gamma$ on the stability of networks. By examining estimates of the local Lipschitz constants, we investigate the stability of metric embedding networks in relation to the demographic distribution of their training sets. A classifier's margin scales inversely with the Lipschitz constant, making classifiers with high local Lipschitz constants less stable and easier to attack~\cite{bartlett2017spectrally,neyshabur2017exploring,scaman2018lipschitz,weng2018evaluating, Luxburg2003DistanceBasedCW}.} We use the RecurJac~\cite{zhang2018recurjac} and Fast-Lin~\cite{weng2018fastlin} bound algorithms to upper-bound local Lipschitz constants within small neural networks trained on datasets with uniform and non-uniform distributions of demographic groups. \Copy{copy:Lipschitz_geometry}{The level of  uniformity in the demographic distribution is captured by the parameter $\gamma$. When the demographics are sampled uniformly, then $\gamma=1$. If we assume the minority demographic group is $\groupA$, then greater disparities in sampling mean $\gamma$ tends to $0$. 
\Cref{fig:lipschitz} in \cref{subsec:Black-box_appendix} shows the distributions of the upper bounds on the estimated local Lipschitz constants for the non-uniform and uniform classifiers trained on 600 identities;  this distribution is plotted for the training set.  We observe larger upper bounds in non-uniform sampling of each demographic group. Further, identities corresponding to minority demographic groups have larger upper bounds on local Lipschitz constants than do majority identities.} These results suggest networks generalize worse for demographic groups which are a minority in the training set, thus networks are less robust to perturbation for certain demographics. \smallskip

\noindent
\textbf{Nearest Neighbors:} In addition to performance of source-target matching, we consider the generalized setting in which a source image must be matched to the best candidate among multiple targets.  This generalized scenario is studied by measuring accuracy in a nearest neighbors model. In the experiment, accuracy refers to the rate at which embeddings designed to impersonate are still classified as the original source identity. 
The training points for the nearest neighbor model are embedding centroids for each identity.

For each demographic group, when impersonating identities in the same demographic group, generated embeddings are accurate on at least $66.7\%$ of examples. For each demographic group, when impersonating identities in different demographic groups, the generated embeddings are accurate on at least $62.9\%$ of examples. 

As expected, the accuracy decreases or remains approximately the same for all demographic groups when impersonating identities from different demographics. The general trend is a decrease in accuracy.

\subsection{Bias Mitigation}\label{subsec:bias_mitigation}
Finally, we examine two bias mitigation techniques and study their impact on the demographic disparities of face obfuscation.
The first technique is a training procedure designed by Xu et al.~\cite{xu2021robustfair} to promote fairness, and the second technique is dataset balancing by demographic. Models of the FaceNet architecture, outputting embeddings in $\realNumbers^{512}$ are trained. Because we are interested in characterizing the best-case obfuscation performance in the presence of bias mitigation strategies, we focus exclusively on the white-box setting.

\noindent
\subsubsection{Training Procedure of Xu et al.} \label{subsubsec:Xu_et_al_model}
Xu et al. propose a technique, that within the context of face obfuscation, is designed to mitigate disparate susceptibility of individuals to impersonation. The training technique decomposes overall error into the natural error and boundary error. \emph{Natural error} refers to the error on examples prior to the addition of a perturbation, and \emph{boundary error} refers to the net increase in error induced when perturbing the natural example. The Xu et al. training procedure incentivizes training of an accurate model such that natural error and boundary error are each roughly equivalent across all identities. 


\begin{figure}[t]
     \centering
     \begin{subfigure}[b]{0.3\textwidth}
         \centering
         \includegraphics[width=\textwidth]{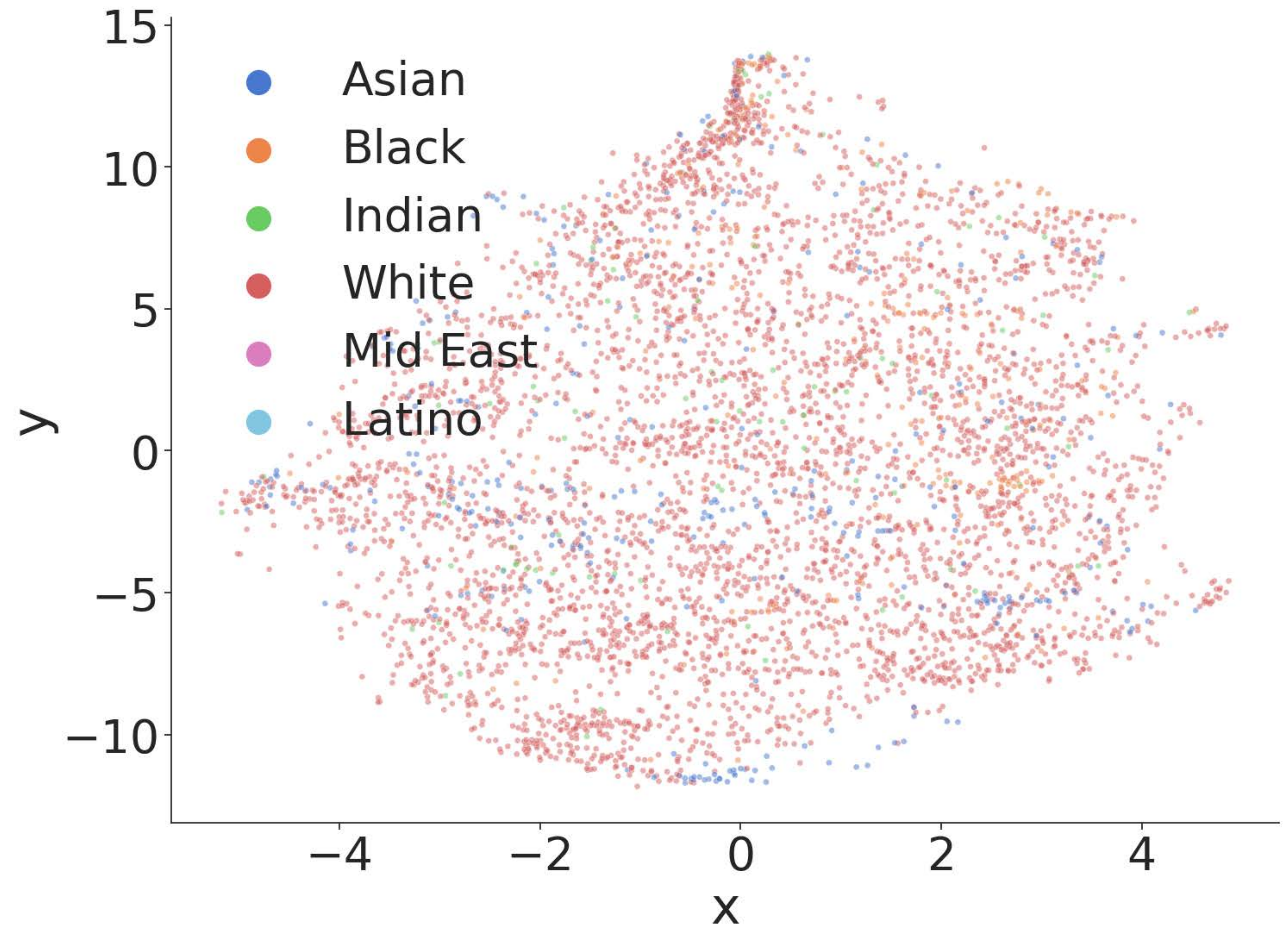}
         \caption{LFW - Race}
         \label{fig:tsne_lfw_race_xu}
     \end{subfigure}
     \begin{subfigure}[b]{0.3\textwidth}
         \centering
         \includegraphics[width=\textwidth]{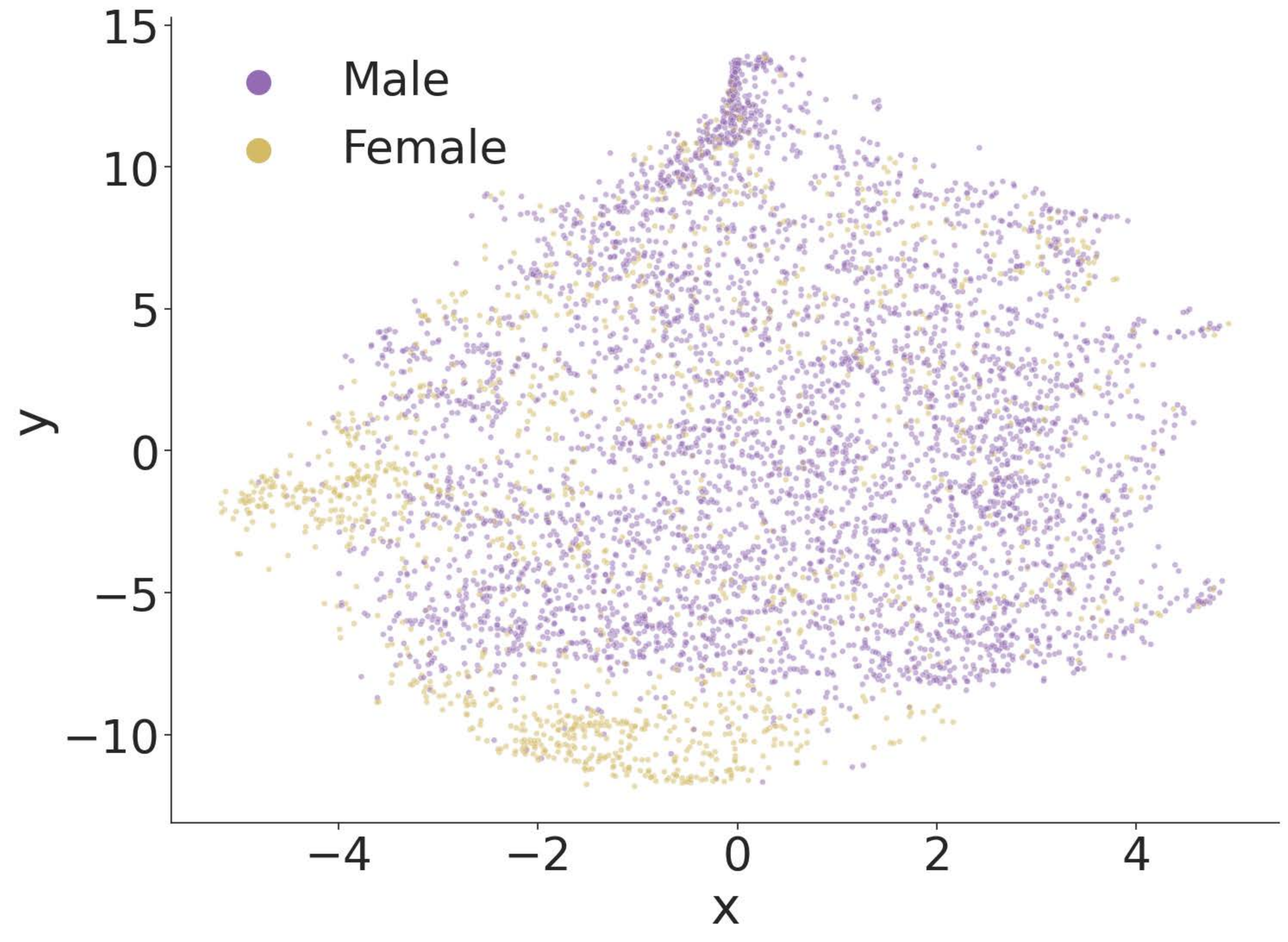}
         \caption{LFW - Sex}
         \label{fig:tsne_lfw_sex_xu}
     \end{subfigure}
     \caption{\Copy{copy:tsne_xu_caption}{t-SNE of the embeddings for the LFW dataset. Embeddings are generated on a FaceNet model trained by Xu et al. procedure.  
     }}
     \label{fig:tsne_xu}
\end{figure}

\begin{figure}[t]
      \centering
     \begin{subfigure}{0.3\columnwidth}
         \centering
         \includegraphics[width=\columnwidth]{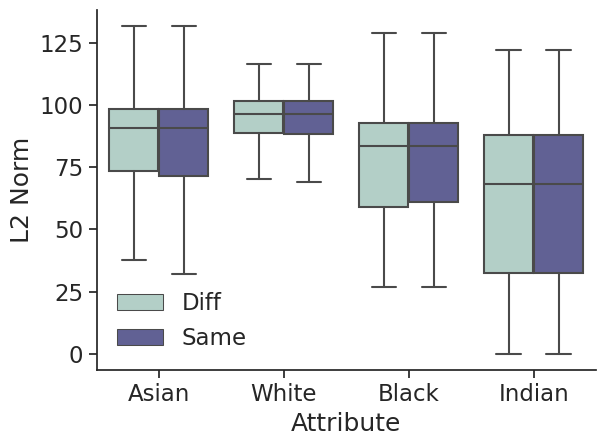}
         \caption{$\ell_2$ Norm: Race}
         \label{fig:pert_norm_xu_race}
     \end{subfigure}
     \begin{subfigure}{0.3\columnwidth}
         \centering
         \includegraphics[width=\columnwidth]{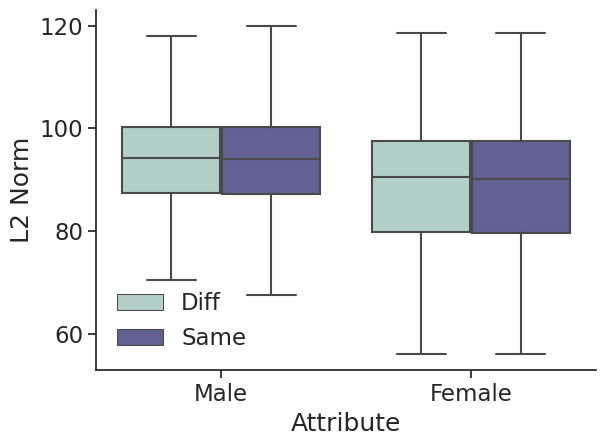}
         \caption{$\ell_2$ Norm: Sex}
         \label{fig:pert_norm_xu_sex}
     \end{subfigure}
        \caption{Distribution of perturbation norms generated by the CW attack on FaceNet model trained with Xu et al. procedure. }
        \label{fig:pert_norm_xu}
\end{figure}

\smallskip
\noindent
\textbf{Visualizing Embeddings:} 
\Copy{copy:XuEmbeddingDiscuss}{
Upon examining the t-SNE plots in \cref{fig:tsne_xu}, we observe the training procedure of Xu et al. does mitigate sources of bias in the embedding space geometry: Demographic clusters are less distinct and embeddings for each demographic group are somewhat interspersed amongst each other. However, clustering is not eliminated entirely; there is a cluster of females in \cref{fig:tsne_lfw_sex_xu}, though many female embeddings are interspersed among embeddings corresponding to males.

}

\smallskip
\noindent
\textbf{Analyzing Obfuscation:}
Embedding space geometry impacts the perturbation norms needed for impersonation: Compared to the pre-trained  model, we see less disparity between the perturbation norms needed to impersonate identities in the same and different demographic groups in \cref{fig:pert_norm_xu}. We put forth a formal null hypothesis to assess our observations. 

\begin{nullHyp} \label{null:demographicPerturbationXu}
    \Copy{copy:null_demographicPerturbationXu}{For the model trained by the Xu et al. procedure, the mean perturbation $\ell_2$ norm $\norm{\perturbation}_2$ necessary to impersonate an identity in the same demographic is identical to the mean perturbation $\ell_2$ norm $\norm{\perturbation}_2$ necessary to impersonate an identity in a different demographic group.}
\end{nullHyp}

\Copy{copy:null_demographicPerturbationXu_explain}{Though we perceive a significant reduction in bias due to the Xu et al. training procedure, we can partially reject \cref{null:demographicPerturbationXu}. There is a statistically significant difference between perturbations targeted within the same demographic and perturbations targeted outside the demographic for only the White demographic, with a $p$-value of $9.61\times 10^{-78}$.}

\begin{figure}[t]
     \centering
     \begin{subfigure}{0.6\columnwidth}
         \centering
         \includegraphics[width=\textwidth]{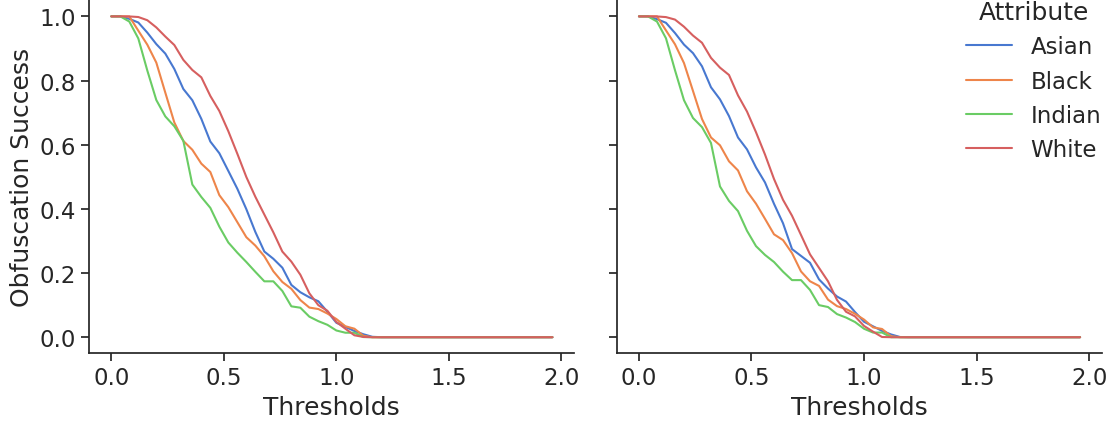}
         \caption{
         Different (Left) Race vs. Same (Right) Race}
         \label{fig:success_Xu_race}
     \end{subfigure}
     \begin{subfigure}{0.6\columnwidth}
         \centering
         \includegraphics[width=\textwidth]{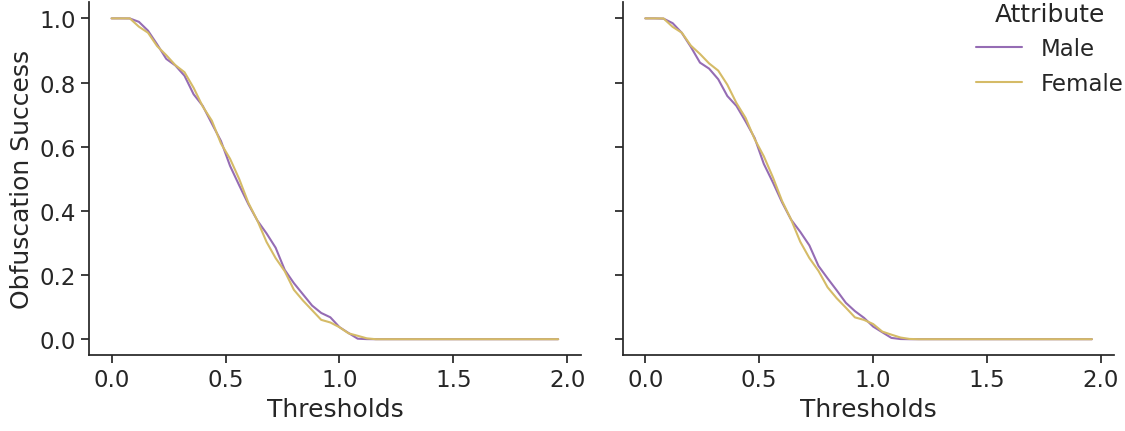}
         \caption{
         Different (Left) Sex vs. Same (Right) Sex }
         \label{fig:success_Xu_sex}
     \end{subfigure}
     \caption{Targeted white-box obfuscation success on a FaceNet model trained with Xu et al. procedure.}
     \label{fig:success_Xu}
\end{figure}

We also observe that the perturbation strength necessary to impersonate an identity has a dependence on demographic. We put forth a formal null hypothesis to test this observation:

\begin{nullHyp} \label{null:crossDemographicPerturbationXu}
    For the model trained by the Xu et al. procedure, the mean perturbation $\ell_2$ norm $\norm{\perturbation}_2$ necessary to impersonate an identity is identical for each source demographic group.
\end{nullHyp}

As anticipated, we can reject this null hypothesis. Differences in perturbation strength between demographic groups are statistically significant for impersonation targeting identities in the same and different sex demographic groups, and the same and different race demographic groups. The $p$-values do not exceed $2.15 \times 10^{-90}$.

\Copy{copy:success_xu_discussion_text}{The significance of such disparities manifests itself in obfuscation obfuscation success rates as depicted in \cref{fig:success_Xu}, especially for success rates conditioned by race. Much like the pre-trained FaceNet, the White race group tends to have the highest obfuscation success rate. Looking at the sex demographic groups, obfuscation success rates appear to be similar. It is unclear what structurally about Xu et al procedure yields a model with obfuscation success rates disparate with respect to race, yet obfuscation success rates conditioned by sex are similar.} For completeness, we put forth a null hypothesis to test the significance of this observation:

\begin{nullHyp}
    For the model trained by the Xu et al. procedure: Across all source demographic groups, the distribution of distance between each obfuscated face embedding and its target is identical.
\end{nullHyp}

We partially reject this null hypothesis: Only when targeting identities outside the identity of source identity demographic group, differences in distance between face embeddings and targets are statistically significant. The $p$-values are  $2.51\times10^{-24}$ and $2.39\times10^{-7}$  when targeting identities in a different race and sex, respectively. Given these results, it is apparent that the Xu et al. training procedure does mitigate bias to some extent, but is far from a perfect solution.

\smallskip
\noindent
\textbf{Accuracy Trade-off:}
\Copy{copy:discuss_xu_accuracy_tradeoff}{
The observed bias mitigation also comes at the cost of model accuracy. We measure these costs by examining the matching performance and through accuracy of a nearest neighbors model.
The matching performance, $\mathrm{TPR}_{0.05}$, of a model trained with the Xu et al. procedure is depicted in the middle subtable of \cref{table:k_fold_confusion_xu}. Matching performance of the reference FaceNet is depicted in the upper subtable of \cref{table:k_fold_confusion_xu}. The matching performance across most demographics is reduced by 1-4\% compared to the reference model. When measuring training accuracy of a nearest neighbors classifier, conditioned by demographic, the lowest accuracy in a model trained with the Xu et al. procedure is 55.2\%. This compares poorly to the reference FaceNet model, for which the worst natural accuracy, when conditioning by demographic, is 95.4\%. Full results for nearest neighbors accuracy can be found in \cref{table:benign_lfw_nearest_neighbor_accuracy}.}

\noindent
\subsubsection{Dataset Balancing}\label{subsubsec:dataset_balancing}
The second bias mitigation strategy we study is that of training face recognition models on demographically balanced datasets. 
To do so, we train three metric embedding networks of the FaceNet architecture. Each training procedure is identical, with the exception of the training dataset.  The network trained on the first dataset acts as the reference:  It is trained on the entirety of the VGGFace2~\cite{caovggface2} training split, a total of 8631 identities. Experiments for this model appear in the appendix.  

The remaining two datasets are used to train comparison models. One such dataset is Sex-Balanced VGGFace2, which contains a total of 4866 identities. The Sex-Balanced VGGFace2 dataset was created by removing original identites to create a training set which contains an equal number identities for the demographic of interest.  Instead of performing data augmentation, we remove examples from VGGFace2 so as to not introduce any artifacts. Similarly, we construct a third dataset, which consists of $307$ identities for each of the race subgroups. We call this dataset, containing 1842 total identities, Race-Balanced VGGFace2.  Race-Balanced VGGFace2 and Sex-Balanced VGGFace2 are collectively referred to as ``(demographically) balanced datasets''. Race-Balanced VGGFace2 and Sex-Balanced VGGFace2 are the training sets used to obtain Race-Balanced FaceNet and Sex-Balanced FaceNet, respectively. We observe that the Balanced FaceNets have less demographic-wise disparity in both face recognition and face obfuscation performance. 

\begin{figure}[t]
     \centering
     \begin{subfigure}[b]{0.3\textwidth}
         \centering
         \includegraphics[width=\textwidth]{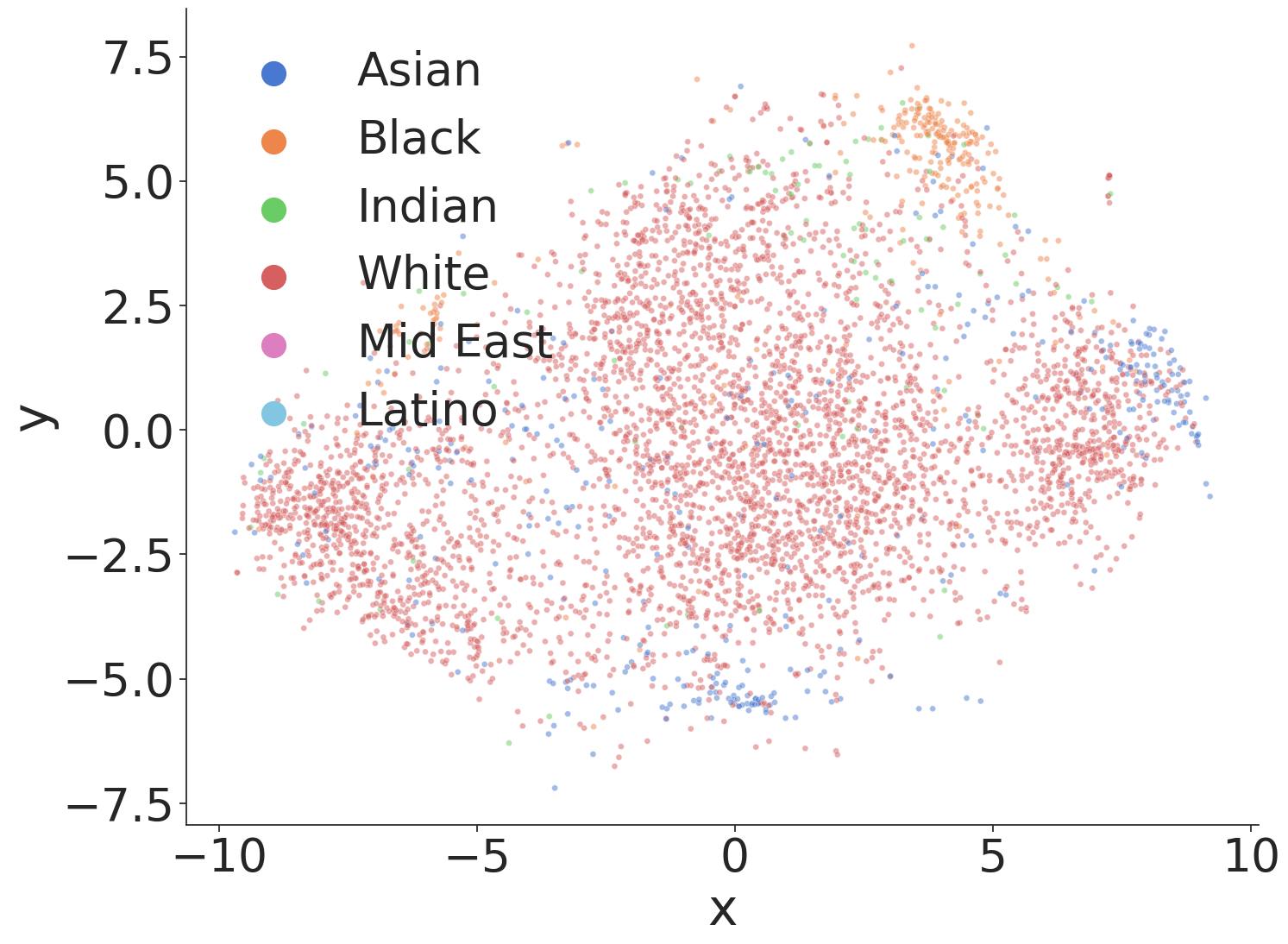}
         \caption{LFW - Race}
         \label{fig:tsne_lfw_race_balanced}
     \end{subfigure}
     \begin{subfigure}[b]{0.3\textwidth}
         \centering
         \includegraphics[width=\textwidth]{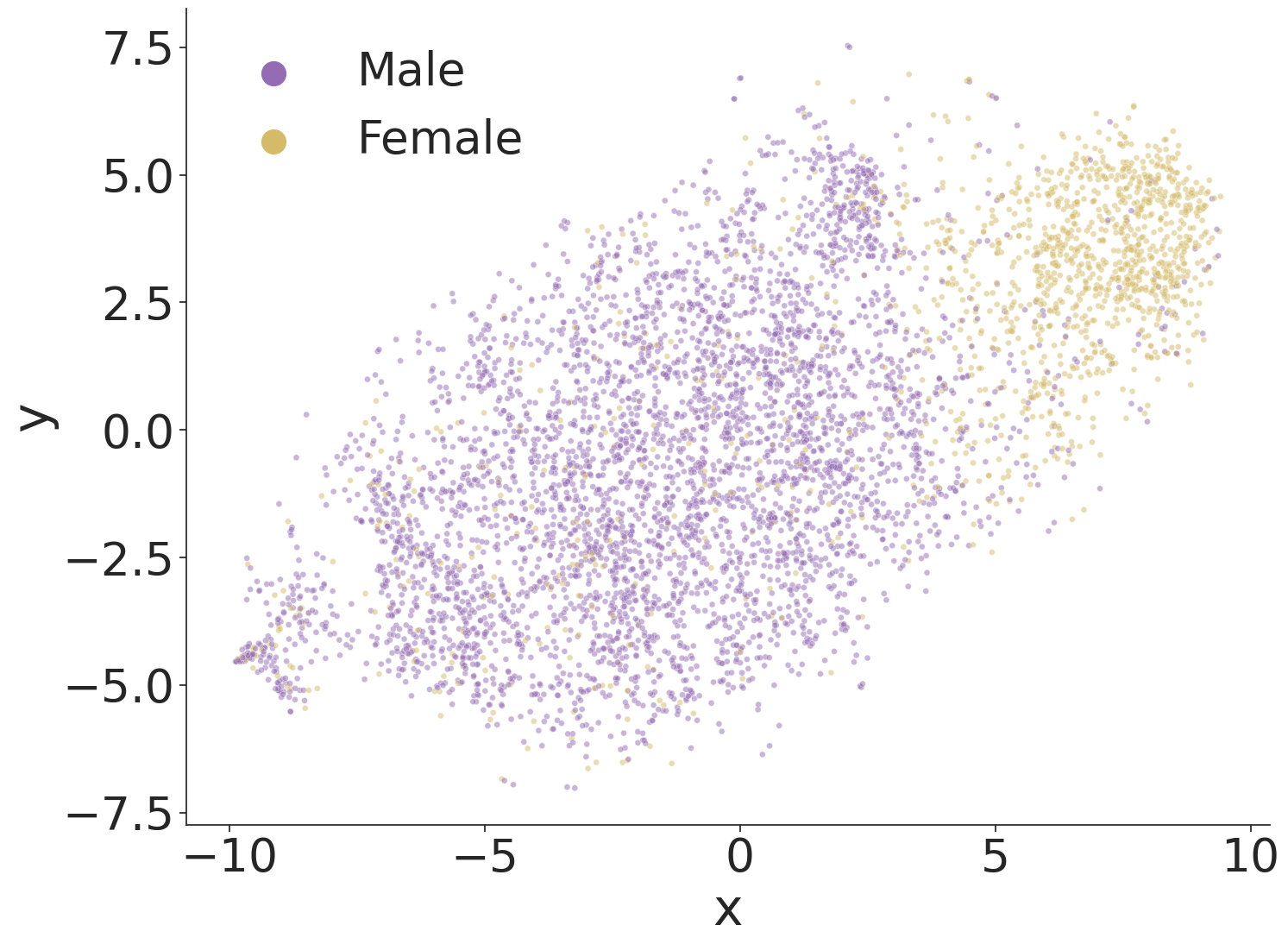}
         \caption{LFW - Sex}
         \label{fig:tsne_lfw_sex_balanced}
     \end{subfigure}
     \caption{\Copy{copy:tsne_balanced_facenet_caption}{t-SNE of embeddings for the LFW dataset. \Cref{fig:tsne_lfw_race_balanced} is generated on Race-Balanced FaceNet. \cref{fig:tsne_lfw_sex_balanced} is generated on Sex-Balanced FaceNet.}}
     \label{fig:tsne_balanced}
\end{figure}

\smallskip
\noindent
\textbf{Visualizing Embeddings from Models:} Upon examining the t-SNE plots in \cref{fig:tsne_balanced}, we notice that
minority demographic groups have larger clusters than do the clusters for the same demographic groups generated on the reference model. Compared to embeddings generated by the model trained with the Xu et al. procedure, demographic clusters generated by Sex-Balanced and Race-Balanced  are more separate and distinct. Further, each demographic group generally appears more distinct.

\begin{figure}[t]
      \centering
     \begin{subfigure}{0.48\columnwidth}
         \centering
         \includegraphics[width=0.8\textwidth]{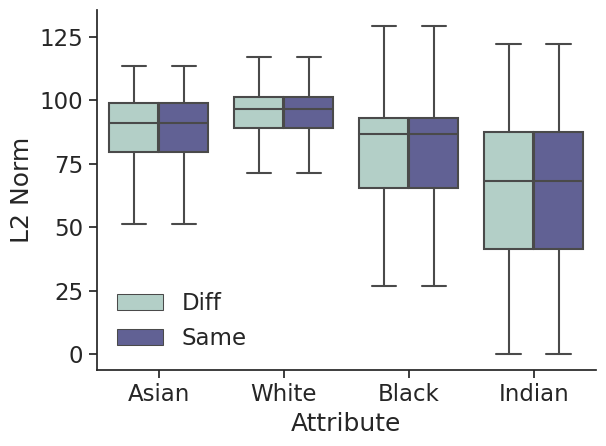}
         \caption{Race-Balanced FaceNet on race data}
         \label{fig:pert_norm_race_balanced}
     \end{subfigure}
     \begin{subfigure}{0.48\columnwidth}
         \centering
         \includegraphics[width=0.8\textwidth]{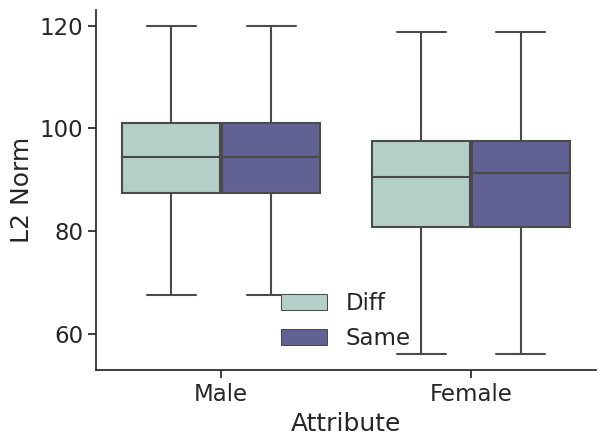}
         \caption{Sex-Balanced FaceNet on on sex data}
         \label{fig:pert_norm_sex_balanced}
     \end{subfigure}
        \caption{\Copy{copy:Pert_norm_balanced_caption}{The distribution of adversarial perturbation sizes generated using the CW attack on FaceNet trained on demographically balanced datasets. }}
        \label{fig:pert_norm_balanced}
\end{figure}


\smallskip
\noindent
\textbf{Analyzing Obfuscation:} Given the distinct demographic clusters we see in balanced  embedding space geometry, we do not expect to see much bias mitigation. Before drawing such a conclusion, we formally test a null hypothesis:
 
\begin{nullHyp} \label{null:demographicPerturbationBalanced}
    For  models trained on balanced datasets, the mean perturbation $\ell_2$ norm $\norm{\perturbation}_2$ necessary to impersonate an identity in the same demographic is identical to the mean perturbation $\ell_2$ norm $\norm{\perturbation}_2$ necessary to impersonate an identity in a different demographic group.
\end{nullHyp}

We are able to partially reject \cref{null:demographicPerturbationBalanced}. There is a statistically significant difference between perturbations targeted within the same demographic and perturbations targeted outside the demographic for the only male, female and white demographic groups with $p$-values of $3.20 \times 10^{-6}$, $5.28 \times 10^{-5}$, and $1.65 \times 10^{-30}$, respectively. 

We observe disparities in the perturbation norms required for faces in each demographic group to be successfully obfuscated. We propose a null hypothesis to test this observation:

\begin{nullHyp} \label{null:crossDemographicPerturbationBalanced}
    For models trained on balanced datasets, the mean perturbation $\ell_2$ norm $\norm{\perturbation}_2$ necessary to impersonate an identity is identical for each source demographic group.
\end{nullHyp}

We are also able to reject this null hypothesis: For models trained on both trained on sex-balanced and race-balanced data, differences in perturbation strength between demographic groups are statistically significant for same sex demographic group, different sex demographic group, same race demographic group, and different race demographic group. The $p$-values do not exceed $2.04 \times 10^{-44}$. 

\begin{figure}[h]
\centering
     \begin{subfigure}[b]{0.6\columnwidth}
         \centering
         \includegraphics[width=\textwidth]{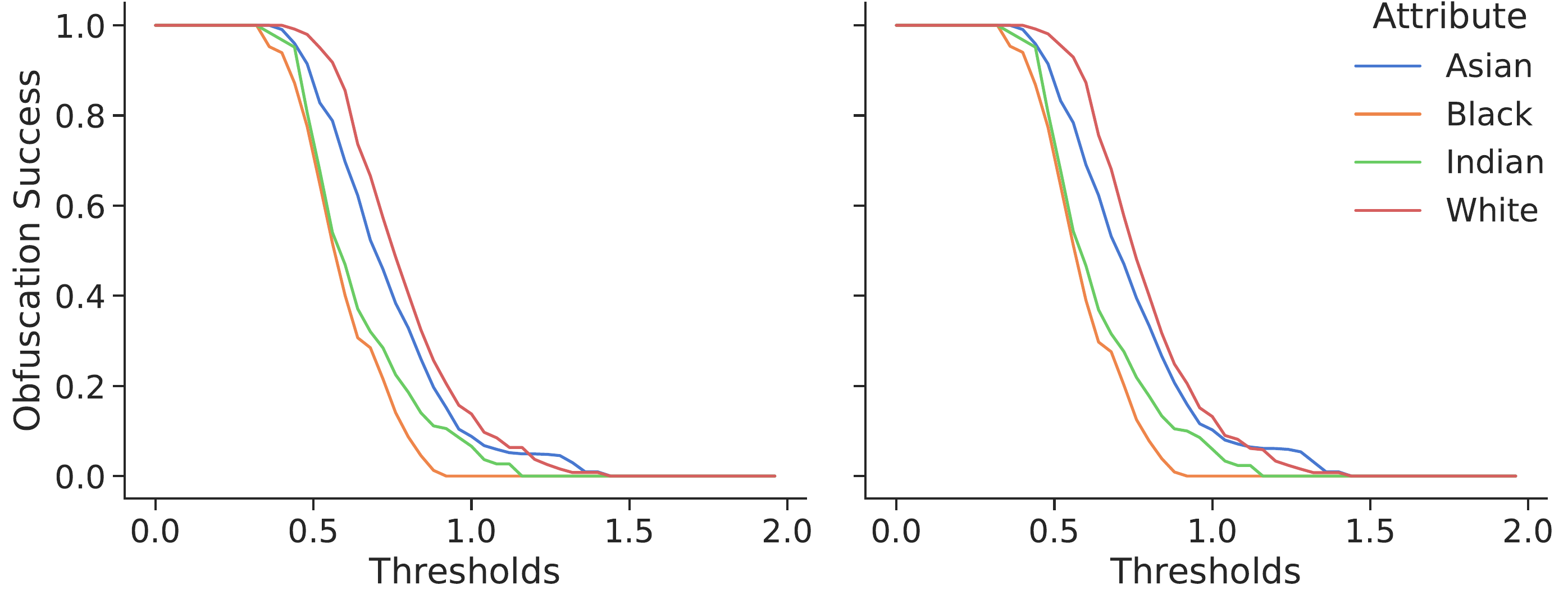}
         \caption{ Different (Left) Race vs. Same (Right) Race. Race-Balanced FaceNet}
         \label{fig:targeted_race_balance}
     \end{subfigure}
     
     \begin{subfigure}[b]{0.6\columnwidth}
         \centering
         \includegraphics[width=\textwidth]{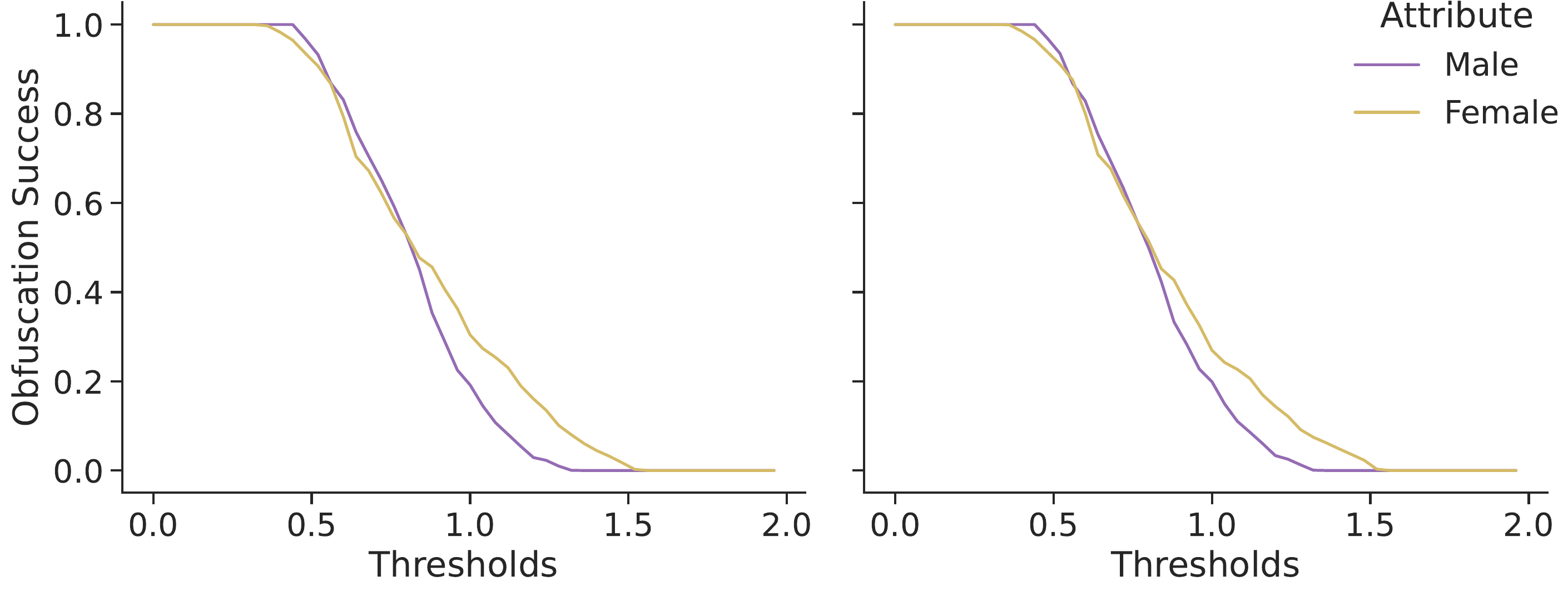}
         \caption{Different (Left) Sex vs. Same (Right) Sex. Sex-Balanced FaceNet}
         \label{fig:targeted_sex_balance}
     \end{subfigure}
     \caption{Targeted obfuscation success evaluated on Demographically Balanced FaceNets in a white-box setting.}
     \label{fig:targeted_balance}
\end{figure}

The disparities in perturbation norm also manifest themselves in obfuscation success rates as depicted in \cref{fig:targeted_balance}. Such success rates are disparate when conditioned by both sex and race. Much like the pre-trained FaceNet model, when conditioned by race, the White demographic group has the highest success rate. Unlike the pre-trained FaceNet, the Female demographic group has higher obfuscation success rate than the male demographic group. It is unclear why the Female demographic group now has higher obfuscation success rates. We put forth a null hypothesis to formally test our observations:

\begin{nullHyp}
    For models trained on balanced data: Across all source demographic groups, the distribution of distance between each obfuscated face embedding and its target is identical.
\end{nullHyp}

The null hypothesis is easily rejected:  Differences in distance between face embeddings and targets are statistically significant in all settings. The $p$-values do not exceed  $1.13 \times 10^{-17}$. Given the significant disparities in face obfuscation performance, it is apparent dataset balancing is ineffective at mitigating bias in face obfuscation.

\smallskip
\noindent
\textbf{Natural Accuracy Trade-off:}
The matching performance, $\mathrm{TPR}_{0.05}$, of a models trained on balanced datasets is the lowest sub-table in \cref{table:k_fold_confusion_xu}. The matching performance across most demographics is reduced by 0.5-3\% compared to the reference model.   When measuring training accuracy of a nearest neighbors classifier, conditioned by demographic, the lowest accuracy in a model trained with demographically balanced training data  94.5\%. This is only a very slight performance reduction relative to the reference FaceNet model.



\newcommand{\STAB}[1]{\begin{tabular}{@{}c@{}}#1\end{tabular}}

\begin{table}[t]

\begin{center}
  \begin{tabular}{g gggggg}
  \toprule
    \multicolumn{7}{c}{\textbf{Reference FaceNet Model}} \\ 
    \midrule
     $\mathrm{TPR}_{0.05}$ & .9046 & .8620 $\;\;\;$ & $\;\;\;$ .9164 & .9152 & .8565 & .9000 \\
    AUC & .9808 & .9864 $\;\;\;$ & $\;\;\;$ .9826 & .9820 & .9711 & 1.000 \\
    \hline
    \rowcolor{white}
    $\mathrm{TPR}_{0.05}$ & .9164 & .9328 $\;\;\;$ & $\;\;\;$ .9294 & .9720 & .9919 & .8000 \\
    \rowcolor{white}
    AUC & .9828 & .9864 $\;\;\;$ & $\;\;\;$ .9847 & .9938 & .9967 & 1.000 \\
    \toprule
    \multicolumn{7}{c}{\textbf{Training Procedure of Xu et al.}} \\ 
    \midrule
    
     $\mathrm{TPR}_{0.05}$ & .8984 & .8180 $\;\;\;$ & $\;\;\;$ .8952 & .8776 & .8258 & .6000 \\
    AUC & .9774 & .9631 $\;\;\;$ & $\;\;\;$ .9782 & .9796 & .9644 & 1.000 \\
    \hline
    \rowcolor{white}
    $\mathrm{TPR}_{0.05}$ & .9036 & .9408 $\;\;\;$ & $\;\;\;$ .9104 & .9768 & .9806 & .9000 \\
    \rowcolor{white}
    AUC & .9806 & .9866 $\;\;\;$ & $\;\;\;$ .9816 & .9945 & .9955 & 1.000 \\
    
     \toprule
     \multicolumn{7}{c}{\textbf{Balanced Training}} \\
     \midrule
     \multicolumn{1}{c}{} & \multicolumn{2}{c}{\textbf{Sex-Balanced}$\;\;\;$}
      &  \multicolumn{4}{c}{$\;\;\;$\textbf{Race-Balanced}} \\
     \midrule
         $\mathrm{TPR}_{0.05}$ & .9184 & .8804 $\;\;\;$ & $\;\;\;$ .8716 & .9160 & .8726 & .8000 \\
    AUC & .9837 & .9733 $\;\;\;$ & $\;\;\;$ .9719 & .9825 & .9738 & .9800 \\
    \hline
    \rowcolor{white}
    $\mathrm{TPR}_{0.05}$ & .9330 & .9244 $\;\;\;$ & $\;\;\;$ .8978 & .9672 & .9823 & .7000 \\
    \rowcolor{white}
    AUC & .9867 & .9842 $\;\;\;$ & $\;\;\;$ .9772 & .9938 & .9950 & 1.000 \\
    \midrule
    \rowcolor{white}
    $N$ & 10000 & 5000 $\;\;\;$ & $\;\;\;$ 10000 & 2500 & 1240 & 20 \\
    \midrule
    \rowcolor{white}
    {} & {\bf Male} & {\bf Female} $\;\;\;$ & $\;\;\;$ {\bf White} & {\bf Asian} & {\bf Black} & {\bf Indian} \\
    \bottomrule
    \end{tabular}
    \vspace{-0.15in}
    \phantom{\begin{tabular}{g gggggg}
    \hline
    \multicolumn{7}{c}{\textbf{Fairness Promoting Adversarial Training}}
    \end{tabular}}
\fbox{\begin{tabular}{llll}
\textcolor{Gray}{$\blacksquare$} & Same Demographic & $\square$ & Any Demographic
\end{tabular}
} 
\end{center}
\caption{The matching performance on LFW on a reference FaceNet model (top), FaceNet trained with Xu et al. procedure (model), and FaceNet models on demographically balanced data.}
\label{table:k_fold_confusion_xu}
\end{table}


\section{Discussion}\label{sec:discussion}

\noindent
\textbf{Potential Remedies:}
In \cref{subsec:bias_mitigation}, we discussed the impact of two bias mitigation strategies on face obfuscation. First, we examined a model training procedure by Xu et al. This procedure does not explicitly optimize for fairness, but instead implicitly relies on a documented connection between adversarial learning and fairness ~\cite{hashimoto18a}. The second bias mitigation procedure involved training FaceNet models on balanced datasets. The Xu et al. procedure mitigated more bias than the balanced Facenet models, but neither procedure completely mitigated disparities in face obfuscation performance. A training procedure which better mitigates performance disparities in face obfuscation efficiently, is a topic for future work.


Beyond bias mitigation in the manner we studied in this paper, balancing performance across combinations of multiple demographic attributes might be of interest. There exists work addressing this problem: Serna et al. propose Sensitive Loss, a ``discrimination-aware'' Triplet Loss tuning procedure for pre-trained models~\cite{serna2020sensitiveloss}.  This tuning procedure involves adding a layer to the network, then tweaking this layer using only identities drawn from the same demographic group. \smallskip

\noindent
\textbf{Threats to Validity:}
 The intent of \cref{subsec:analyticalModel} is the creation of a tractable analytical model for embedding spaces and dimensionality reduction in general.  Perhaps the most significant limitation of this analytical model is that PCA is a linear embedding function and is incapable of capturing non-linear effects present in metric embedding networks. 
 Another threat to validity is the scarcity of publicly available face recognition datasets; datasets with labeled demographic attributes are even more rare.  Such datasets are rather small, as such, our results may not generalize to datasets orders of magnitude larger than the available face recognition datasets.

\section{Related Work}
\label{sec:related}


 Buolamwini and Gebru study commercially available face recognition datasets and classifiers~\cite{DBLP:Gendershades}.  Their findings indicate that prominent commercial face classifiers exhibit disparate performance across demographic groups. Follow-up research has attempted to address these demographic disparities through balanced datasets~\cite{karkkainen19fairface, thomee2016yfcc}. Such datasets improve model generalization but do not resolve disparities completely. These findings are consistent with results in \cref{sec:AdditionalExperiments}.

Demographic disparities seem to contribute to the phenomenon of overlearning, a term coined by Song and Shmatikov \cite{song2019overlearning}.  Overlearning refers to the phenomenon where models implicitly learn to recognize sensitive patterns not part of the original learning objective. Metric embedding networks trained on faces do overlearn;  they learn a demographically-aware dimensionality reduction on faces.

\Copy{copy:Cherepanova_cite}{
More related to our research, Cherepanova et al.~\cite{cherepanova2022deep} study the relationship between training data, test data, population demographics and face recognition performance. The authors, independent of our results, observe that models trained on demographically  datasets are not bias-free. They discuss neither face obfuscation, nor impacts on privacy and security.}

\Copy{copy:Nanda_cite}{
Nanda et al. discuss the relationship between fairness and robustness of face recognition~\cite{nanda20umdfairness}.  
Our work differs in several respects: First, there is a distinction between the study of robust metric learning. Our paper explores bias present in the embedding space and how it affects recently proposed face obfuscation systems. Through an analytical model in \cref{subsec:analyticalModel}, we show the amount of tolerable perturbation in a face recognition system depends on class imbalances. Second, we examine face obfuscation in the black-box setting; these experiments are performed on both open-source and commercial models as depicted in \cref{fig:azure-lowkey,fig:facepp-faceoff-appendix}. The results demonstrate transferability of biases in adversarial perturbations; we note the negatively-affected demographics can differ between the white-box and black-box settings depending on the perturbation strength. Third, we investigate root causes for demographic-wise performance disparities via TCAV.  We also explore dataset balancing and the Xu et al. training procedure as bias mitigation techniques.
}

\Copy{copy:Cilloni_cite}{
Cilloni et al. study a targeted face obfuscation which attempts to find the minimum strength successful perturbation via optimization procedure regularized by the a well-established image similarity scoring function known as DSSIM. Modulo the regularizer, our Carlini-Wagner based experiments achieve the same purpose.
}
\Copy{copy:Rajabi_cite}{
Rajabi et al~\cite{RajabiFacePrivacy}. consider two methods of face obfuscation: the first is a universal ensemble perturbation, an untargeted obfuscatory method designed to be transferable; we explore transferability of untargeted perturbations in \cref{subsec:obfuscationEffectiveness} and \cref{fig:transferability_success}. The second technique is cryptographic in nature.}
Finally, a Master's thesis by Qin~\cite{qin2021thesis} evaluates discrepancies in a face obfuscation system called FAWKES~\cite{DBLP:Fawkes}, conditioned on skin tones. Qin finds differences in perturbation visibility for certain demographics.  In comparison, our work characterizes these discrepancies and their impact on face obfuscation systems.

\section{Conclusion}
Face recognition systems have seen increased usage in online settings at the cost of heightened privacy concerns. Researchers have proposed face obfuscation systems that leverage evasion attacks against metric embedding networks. Our results show that, in an effort to mitigate such privacy concerns, face obfuscation systems have performance characteristics that depend on demographic information, thereby creating a new privacy incursion.  Such performance characteristics can not only leak demographic membership information and decrease the performance of face obfuscation among underrepresented demographic groups.  Imbalances in training set demographics are only partially to blame for this privacy leak: remaining causes are yet to be discovered. To mitigate the effects of this incidental privacy leak, we must not only develop loss functions for training fair metric embedding networks, but also develop techniques to characterize if such privacy leaks will occur. 


\newpage
    
\bibliographystyle{IEEEtran}
\bibliography{references}
\newpage
\newpage
\newpage
\appendix


\section{Supplemetal Mathematics for the Analytical Model}\label{sec:pca_math}

To solve the optimization problem posed in \cref{eq:theoryOptimizationProblem}, we examine the likelihood function. Let $\PDF_{\sampleDist_{\placeholderGroup}}$ now denote the PDF of the image of distribution $\sampleDist_{\placeholderGroup}$ as it appears in the 1-D PCA embedding space.  We assume $\gamma \leq 1$.  We aim to find the strength of the perturbation necessary to push an example $\sample$ in $\groupA$ $\frac{\PDF_{\sampleDist_{\groupA}}\left[\embeddingFunc(\sample+\perturbation) \right]}{\PDF_{\sampleDist_{\groupB}}\left[\embeddingFunc(\sample+\perturbation) \right]}$ exceeds one, where:

{\small
\begin{align}
    1&\leq\left(\frac{\PDF_{\sampleDist_{\groupA}}\left[\embeddingFunc(\sample+\perturbation) \given \perturbation \propto \frac{(\vec{\mu}_{\groupB}-\sample)}{\norm{\vec{\mu}_{\groupB}-\sample}_2}\right]}{\PDF_{\sampleDist_{\groupB}}\left[\embeddingFunc(\sample+\perturbation) \given  \perturbation \propto \frac{(\vec{\mu}_{\groupB}-\sample)}{\norm{\vec{\mu}_{\groupB}-\sample}_2} \right]}\right) \label{eq:initialLikelihoodRatio} \\
    \begin{split} 
    &\qquad=\left((2\pi)^{-1/2}(\embeddingFunc(\vec{q}_1))^{-1}\right)\\
    &\qquad \quad \times\exp\left\{-\frac{1}{2}\left[\embeddingFunc\left(\sample+\eta\frac{(\vec{\mu}_{\groupB}-\sample)}{\norm{\vec{\mu}_{\groupB}-\sample}_2} -\vec{\mu}_{\groupA}\right)\right]^2 (\embeddingFunc(\vec{q}_1))^{-2}\right\}\\
     &\qquad \quad \Bigg[\left((2\gamma\pi)^{-1/2}(\embeddingFunc(\vec{q}_1))^{-1}\right)\\
    &\qquad \quad \times \exp\left\{-\frac{\gamma}{2}\cdot\left[\embeddingFunc\left(\sample+\eta\frac{(\vec{\mu}_{\groupB}-\sample)}{\norm{\vec{\mu}_{\groupB}-\sample}_2} -\vec{\mu}_{\groupB}\right)\right]^2 (\embeddingFunc(\vec{q}_1))^{-2}\right\}\Bigg]^{-1}
     \end{split}\\
      \begin{split}
    &\qquad\leq\exp\left\{-\frac{1}{2}\left[\embeddingFunc\left(\sample+\eta\frac{(\vec{\mu}_{\groupB}-\sample)}{\norm{\vec{\mu}_{\groupB}-\sample}_2}\right) +\embeddingFunc\left(\vec{\mu}_{\groupB}\right)\right]^2 \right\}\\
    &\qquad \quad \times \exp\left\{-\frac{\gamma}{2}\cdot\left[\embeddingFunc\left(\sample+\eta\frac{(\vec{\mu}_{\groupB}-\sample)}{\norm{\vec{\mu}_{\groupB}-\sample}_2}\right) -\embeddingFunc\left(\vec{\mu}_{\groupB}\right)\right]^2 \right\}\Bigg]^{-1}
     \end{split}\label{eq:likelihoodRatio}
\end{align}
}
We now solve the following the following inequality for $\eta$
{\small
\begin{align}
    \begin{split}
    1 &\leq\exp\left\{-\frac{1}{2}\left[\embeddingFunc\left(\sample+\eta\frac{(\vec{\mu}_{\groupB}-\sample)}{\norm{\vec{\mu}_{\groupB}-\sample}_2}\right) +\embeddingFunc\left(\vec{\mu}_{\groupB}\right)\right]^2 \right\}\\
    &\qquad \quad \times \exp\left\{-\frac{\gamma}{2}\cdot\left[\embeddingFunc\left(\sample+\eta\frac{(\vec{\mu}_{\groupB}-\sample)}{\norm{\vec{\mu}_{\groupB}-\sample}_2}\right) -\embeddingFunc\left(\vec{\mu}_{\groupB}\right)\right]^2 \right\}\Bigg]^{-1}
     \end{split}\label{eq:solveInequality}
\end{align}
}

After some algebra, we arrive at the following interval solution for $\eta$.

\begin{align}
    \eta &> \frac{2b(1-\gamma)\sqrt{\frac{a^2\gamma}{b^2(\gamma-1)^2}} +a\gamma +a + \embeddingFunc(\vec{\mu}_{\groupB})(1-\gamma)}{b(\gamma-1)}\label{eq:eta_lower}
    \intertext{AND}
    \eta &< \frac{2b(\gamma-1)\sqrt{\frac{a^2\gamma}{b^2(\gamma-1)^2}} +a\gamma +a + \embeddingFunc(\vec{\mu}_{\groupB})(1-\gamma)}{b(\gamma-1)} \label{eq:eta_upper}
\end{align}

Where, for notational compactness, we denoted the following:
{\small
\begin{align}
a&=\embeddingFunc(\sample) \\
b&=\embeddingFunc\left(\frac{(\vec{\mu}_{\groupB}-\sample)}{\norm{\vec{\mu}_{\groupB}-\sample}_2}\right)
\end{align}
}

Since the right-side of inequality \eqref{eq:solveInequality} upper bounds the right-side of inequality \eqref{eq:initialLikelihoodRatio}, we know that any solution for inequality \eqref{eq:solveInequality} is also a solution for inequality \eqref{eq:initialLikelihoodRatio}.  

Since inequality \eqref{eq:eta_upper} is a bound on $\epsilon$ which provides a guarantee on when \cref{eq:theoryOptimizationProblem} is infeasible. Hence we conclude that \cref{eq:theoryOptimizationProblem} may be feasible only when

{\small
\begin{align}
    \epsilon &\geq \max\left\{0,\frac{2b(\gamma-1)\sqrt{\frac{a^2\gamma}{b^2(\gamma-1)^2}} +a\gamma +a + \embeddingFunc(\vec{\mu}_{\groupB})(1-\gamma)}{b(\gamma-1)}\right\} 
\end{align}
}

\section{Supplemental Information for Analytical Model}\label{subsec:pcaAdversarialAttack}

\begin{figure}[t]
     \centering
     \begin{subfigure}[t]{0.48\columnwidth}
         \centering
         \includegraphics[width=0.8\textwidth]{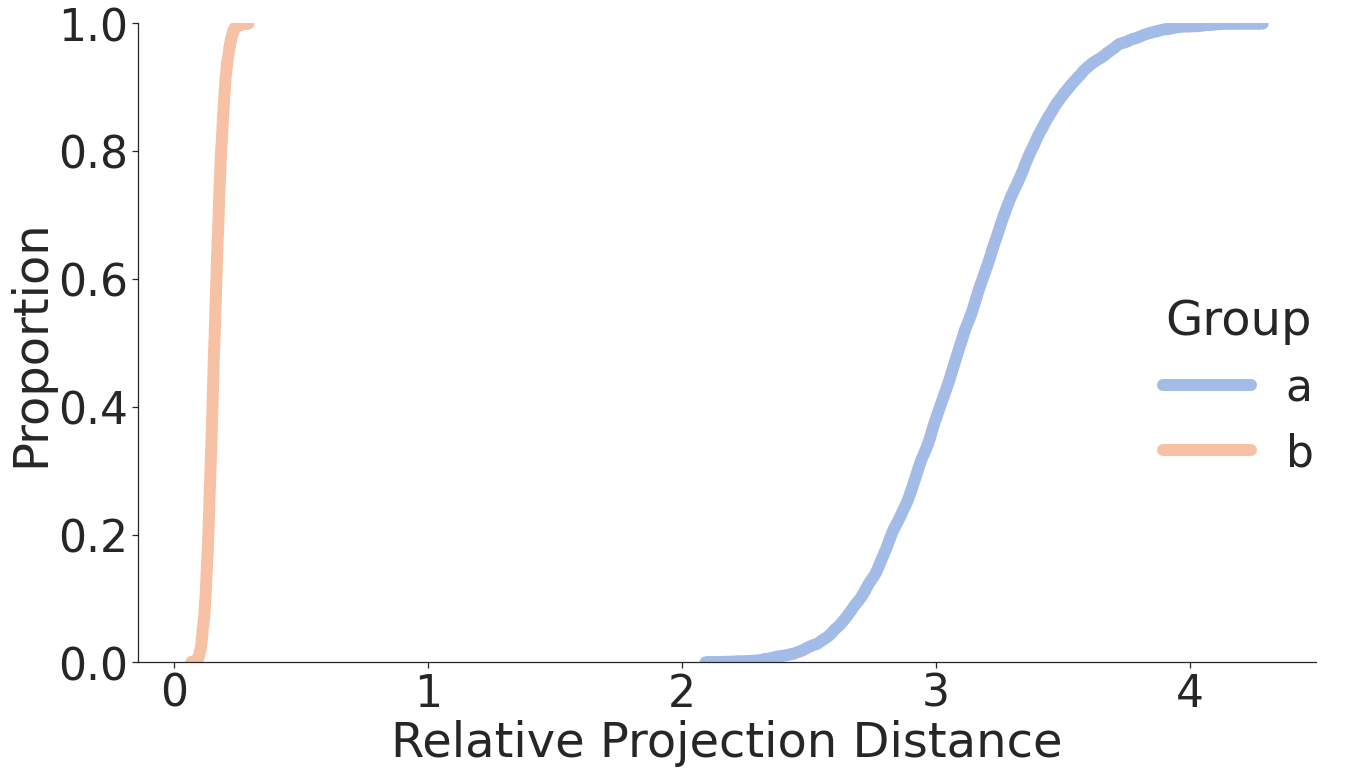}
         \caption{Histogram of relative projection distance is plotted for groups $\groupA$ and $\groupB$ with $\gamma = \frac{1}{100}$.}
         \label{fig:pca_imbalance}
     \end{subfigure}
     \hfill
     \begin{subfigure}[t]{0.48\columnwidth}
         \centering
            \includegraphics[width=0.8\textwidth]{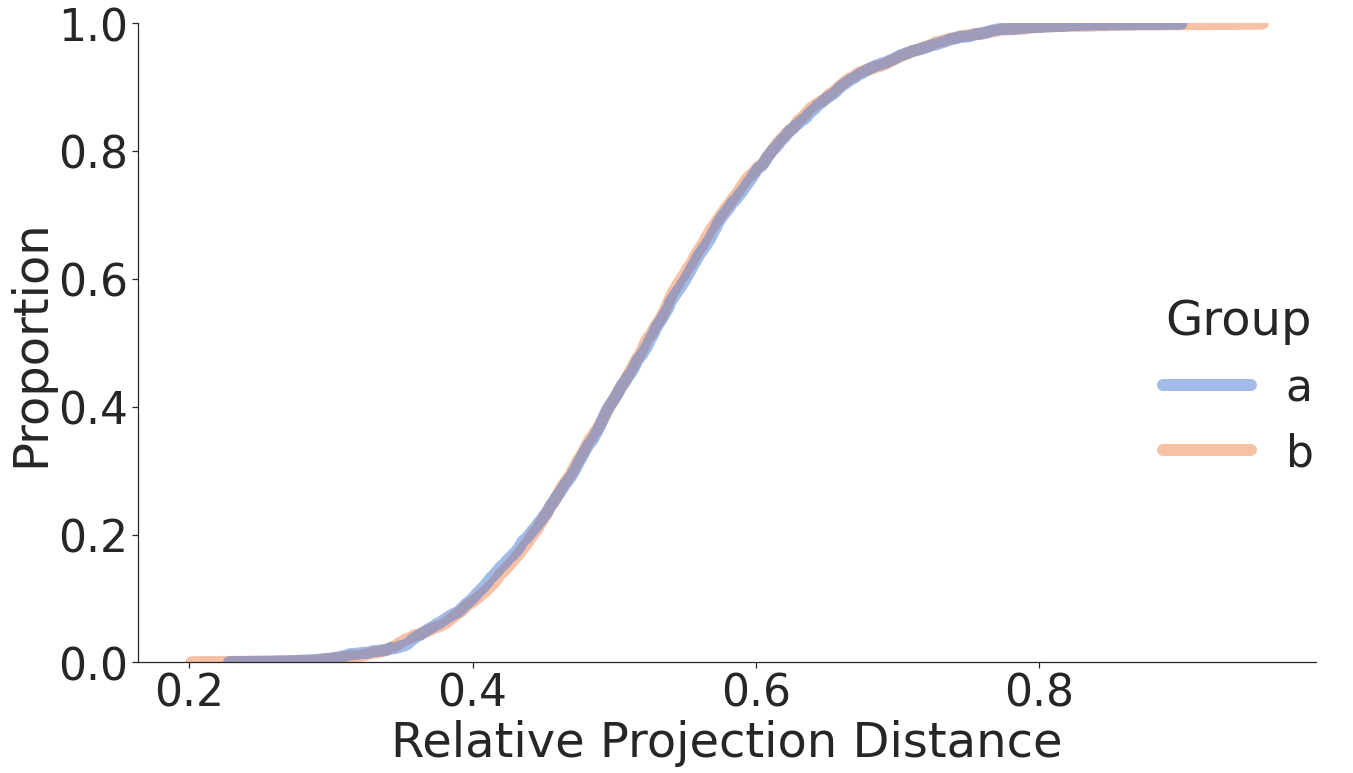}
            \caption{Histogram of relative projection distance is plotted for groups $\groupA$ and $\groupB$ with $\gamma = 1$.}
         \label{fig:pca_balance}
     \end{subfigure}
        \caption{The empirical cumulative distribution function of relative projection distance is plotted for $\gamma = \frac{1}{100}$ and $\gamma = 1$. $2500$ examples per group.}
        \label{fig:plotRelProjectionDistance}
\end{figure}

Given our interest in studying the impact of the frequency of each group in a training set on the efficacy of a learned embedding network, we focus on how the relative projection distance relates to local Lipschitz constant for faces in each group ($\groupA$ and $\groupB$): \Cref{fig:plotRelProjectionDistance} portrays the groupwise distribution of relative projection distances. The case in which  the local Lipschitz constants for each group are equal is captured by \cref{fig:pca_balance}.  \Cref{fig:pca_imbalance} captures the setting in which the local Lipschitz constants for each group are non-equal.  This discrepancy in Lipschitz constant manifests itself as $\gamma$ tending away from $1$.  In particular, each subfigure has $5000$ total identities with $\placeholderCount = 50$ images sampled per identity. The number of examples in each group is $2500$. \Cref{fig:pca_imbalance} is constructed such that $\vec{\Sigma}_{\groupA} = \frac{1}{100}\vec{\Sigma}_{\groupB}$ and  \Cref{fig:pca_balance} is constructed such that $\vec{\Sigma}_{\groupA} = \vec{\Sigma}_{\groupB}$.   Furthermore, observe how in \cref{fig:pca_imbalance}, the distribution of relative projection distances for group~$\groupA$ is shifted right and more dispersed than the distribution of relative projection distances for group~$\groupB$.

 Relating relative projection distance to the experiments on face datasets: minority groups tend to have larger local Lipschitz constants (\cref{fig:lipschitz}), meaning their distribution of relative projection distances is more dispersed than the distribution of relative projection distances for a more frequent population group. We conclude that greater dispersion in relative projection distances contributes to the performance disparities incurred by minority groups in face recognition.  
 
\section{Additional Experiments}\label{sec:AdditionalExperiments}

This section contains additional experiments which further our understanding of fairness issues present in both face recognition and face obfuscation systems. Results from these experiments are discussed.

\subsection{Face Obfuscation and Demographics}\label{subsec:obfuscation-impact-demographic_appendix}

\begin{figure}[h]
\centering
     \begin{subfigure}{0.6\columnwidth}
         \centering
         \includegraphics[width=\textwidth]{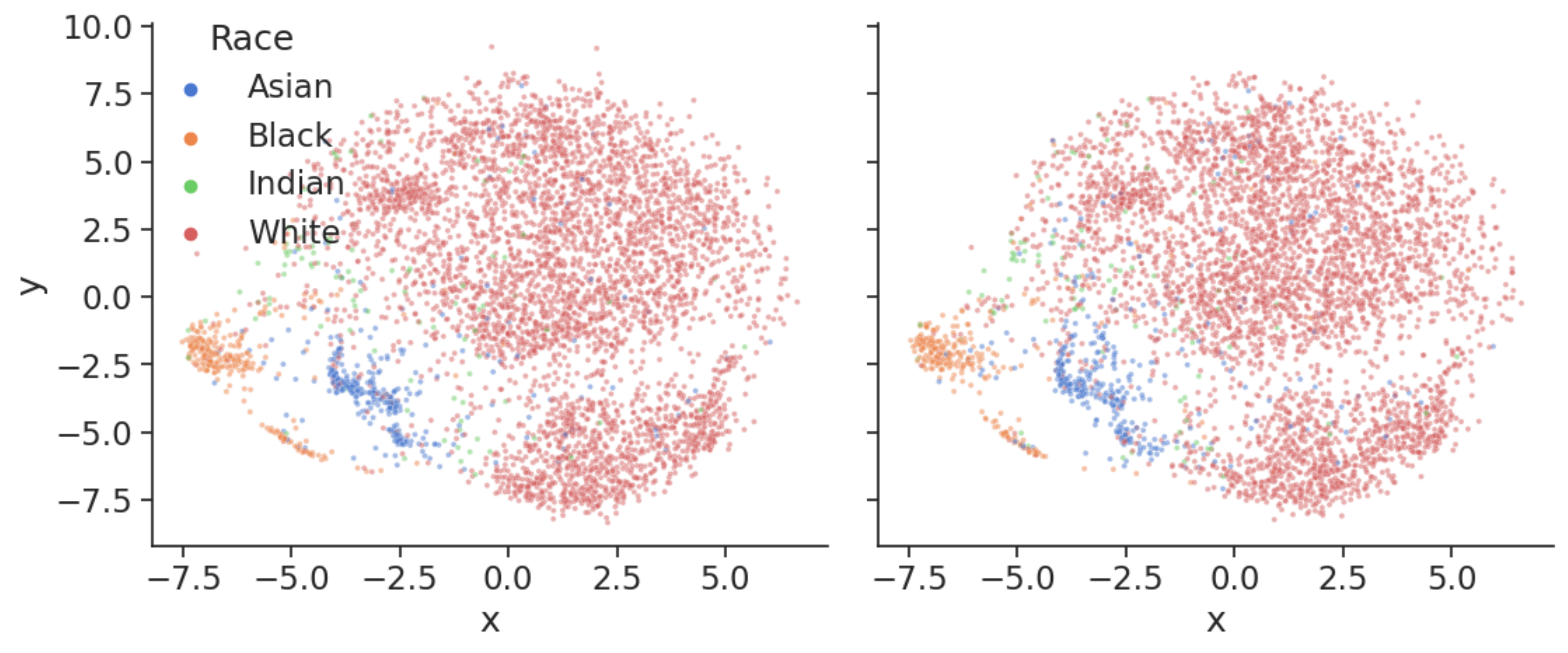}
         \caption{Race: Natural (Left) vs. Adversarial (Right)}
         \label{fig:tsne_adv_race}
     \end{subfigure}
     
     \begin{subfigure}{\columnwidth}
         \centering
         \includegraphics[width=0.6\textwidth]{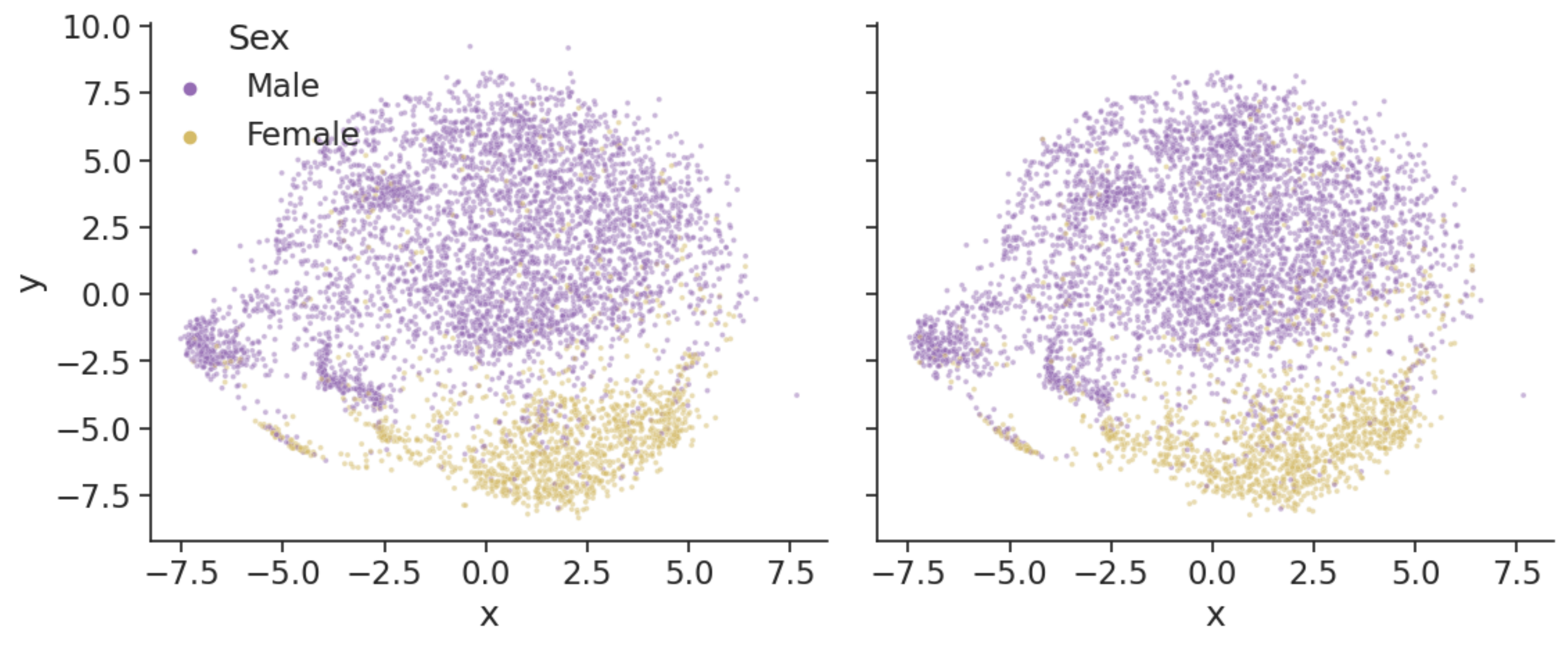}
         \caption{Sex: Natural (Left) vs. Adversarial (Right)}
         \label{fig:tsne_adv_sex}
     \end{subfigure}
     \caption{t-SNE of the embedding spaces generated using the non-adversarial and untargeted adversarial LFW datasets. Embeddings of identities are colored by race and sex. Note that the clusters do not change, identities are still rooted within their own demographic.}
     \label{fig:tsne_nat_adv}
\end{figure}

While adversarial examples cause a misclassification on a particular identity, these adversarial examples do not necessarily cause their demographic attributes to change. To validate this claim, we generate untargeted adversarial examples on the LFW dataset and classify these perturbed images using a classifier that predicts the race attribute from a face~\cite{serengil2020lightface}. Of each identity in the dataset, only 8\% of the identities' race attribute change after adversarial perturbations are added. We visualize this in \cref{fig:tsne_nat_adv}, where the embeddings of the natural examples and adversarial examples are plotted according to their ground truth demographics using t-SNE. Comparing the plots within \cref{fig:tsne_adv_race,fig:tsne_adv_sex}, a heavy overlap is observed between the embeddings of both natural and adversarial examples of the same race and the same sex. These results further addressed confirm our intuition from \cref{subsubsec:analysisEffectivenessObfuscation}; an obfuscated face is unlikely to depart it's demographic group within the embedding space. This is especially true when the groups are clustered in the embedding space.

\begin{table}[t]
\centering
\begin{tabular}{c c c c c c} 
 \toprule
 I & II & III & IV & V & VI \\ 
 \midrule
  \cellcolor[HTML]{F4F2F5} 0xF4F2F5 & \cellcolor[HTML]{FAF0EF} 0xFAF0EF & \cellcolor[HTML]{FFF9E2} 0xFFF9E2 & \cellcolor[HTML]{E4C567} 0xE4C567 & \cellcolor[HTML]{8F573C} {\color{white} 0x8F573C} & \cellcolor[HTML]{2D2024} {\color{white}0x2D2024}\\
 \cellcolor[HTML]{EDECEA} 0xEDECEA & \cellcolor[HTML]{F3EBE6} 0xF3EBE6 & \cellcolor[HTML]{F1E8C4} 0xF1E8C4 & \cellcolor[HTML]{E2C26A} 0xE2C26A & \cellcolor[HTML]{7A4C2C} {\color{white}0x7A4C2C} & \cellcolor[HTML]{14152A} {\color{white}0x14152A}\\ 
 \cellcolor[HTML]{FAF9F7} 0xFAF9F7 & \cellcolor[HTML]{F4F1EB} 0xF4F1EB & \cellcolor[HTML]{F0E3AE} 0xF0E3AE & \cellcolor[HTML]{E0C27C} 0xE0C27C & \cellcolor[HTML]{642D0E} {\color{white}0x642D0E} & \\
 \cellcolor[HTML]{FDFBE7} 0xFDFBE7 & \cellcolor[HTML]{FBFCF4} 0xFBFCF4 & \cellcolor[HTML]{E1D394} 0xE1D394 & \cellcolor[HTML]{DFB978} 0xDFB978 & \cellcolor[HTML]{652C1A} {\color{white}0x652C1A} & \\ 
 \cellcolor[HTML]{FDF6E7} 0xFDF6E7 & \cellcolor[HTML]{FCF8EE} 0xFCF8EE & \cellcolor[HTML]{F2E398} 0xF2E398 & \cellcolor[HTML]{C8A664} 0xC8A664 & \cellcolor[HTML]{602D1B} {\color{white}0x602D1B} & \\ 
 \cellcolor[HTML]{FEF7E6} 0xFEF7E6 & \cellcolor[HTML]{FEF6E2} 0xFEF6E2 & \cellcolor[HTML]{ECD7A0} 0xECD7A0 & \cellcolor[HTML]{BD9862} 0xBD9862 & \cellcolor[HTML]{562E24} {\color{white}0x562E24} & \\ 
  & \cellcolor[HTML]{FFF9E2} 0xFFF9E2 & \cellcolor[HTML]{ECDA86} 0xECDA86 & \cellcolor[HTML]{9D6B41} 0x9D6B41 & \cellcolor[HTML]{3E1A0D} {\color{white} 0x3E1A0D} & \\ [1ex] 
 \bottomrule
\end{tabular}
\caption{This chart depicts colors in the Fitzpatrick scale~\cite{skin_tones}, which is derived from Von Luschan's chromatic scale~\cite{von1897beitrage}. Colors are in hexadecimal.}
\label{table:Fitzpatrick}
\end{table}

\subsection{Black-Box Obfuscation Disparities}\label{subsec:Black-box_appendix}

\begin{figure}[t]
    \centering
    \includegraphics[width=0.6\columnwidth]{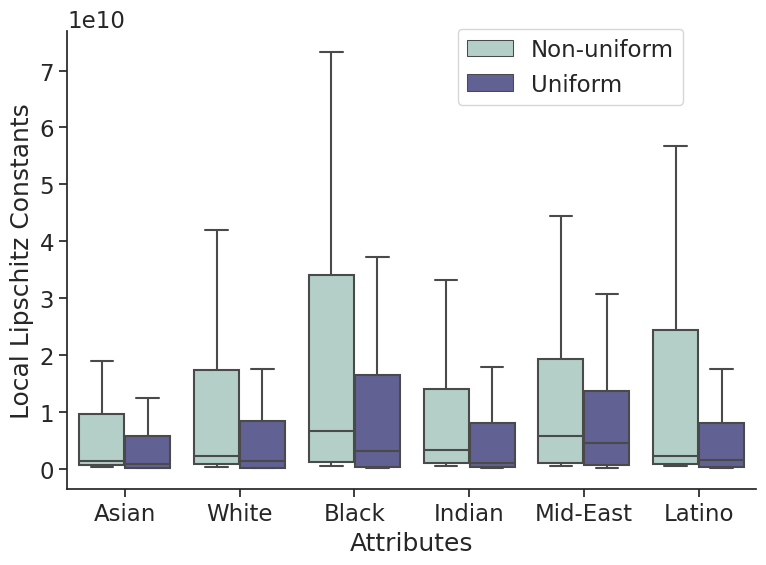} 
     \caption{Upper bounds on local Lipschitz constants estimated using Fast-Lin and RecurJac.  Models trained on non-uniform demographic distributions suffer from higher instability.  The maximum constants for each demographic is 2-3 times larger in the non-uniform model than in the uniform model.}
     \label{fig:lipschitz}
\end{figure}

To understand whether the performance disparities between demographics (as discussed in \cref{subsec:obfuscationEffectiveness}) manifest in commercial face recognition systems, we tested the success of the perturbed faces against three commercially available face recognition systems. These three Face Recognition APIs are Face++, Microsoft Azure Face API, and Amazon AWS Rekognition . In \cref{fig:azure-lowkey}, we observe large differences in obfuscation success rates dependent on the race demographic. This performance disparity can be attributed to the larger perturbation sizes associated with Asian and White identities (\cref{fig:cdf_pert_norm}). The Black demographic was also observed to have smaller perturbations, and higher local Lipschitz constants (\cref{fig:lipschitz}), suggesting that lower obfuscation success rates can stem from disparities in the robustness of each demographic. 

\begin{figure}[h]
\centering
     \begin{subfigure}{0.3\columnwidth}
         \centering
         \includegraphics[width=\textwidth]{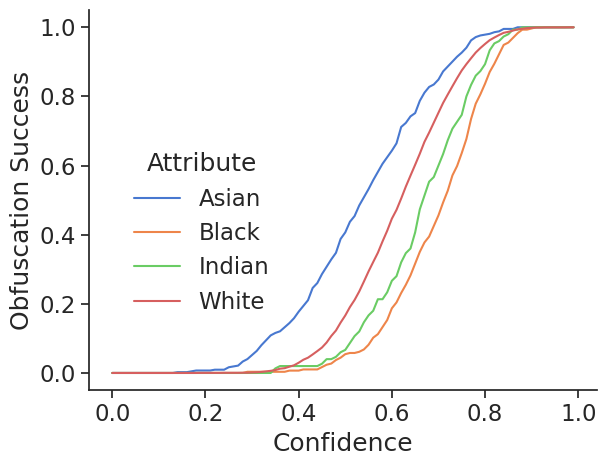}
         \caption{Untargeted Face++: Race}
         \label{fig:facepp-lowkey-race-appendix}
     \end{subfigure}
     \begin{subfigure}{0.3\columnwidth}
         \centering
         \includegraphics[width=\textwidth]{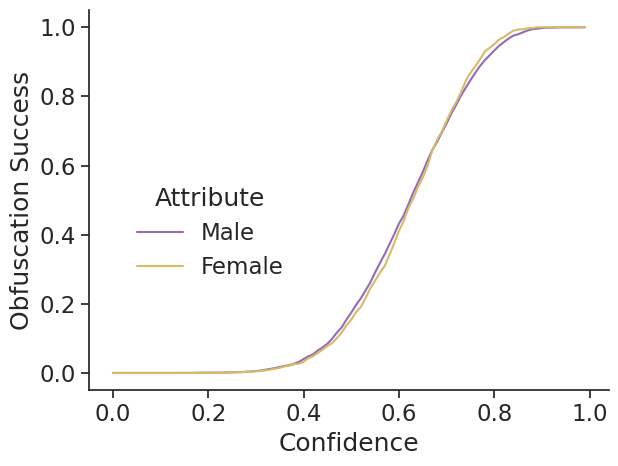}
         \caption{Untargeted Face++: Sex}
         \label{fig:facepp-lowkey-sex-appendix}
     \end{subfigure}
    
     \begin{subfigure}{0.3\columnwidth}
         \centering
         \includegraphics[width=\textwidth]{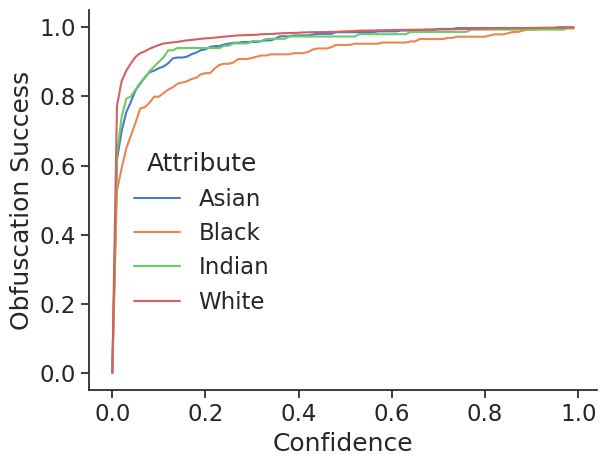}
         \caption{Untargeted AWS: Race}
         \label{fig:aws-lowkey-race-appendix}
     \end{subfigure}
     \begin{subfigure}{0.3\columnwidth}
         \centering
         \includegraphics[width=\textwidth]{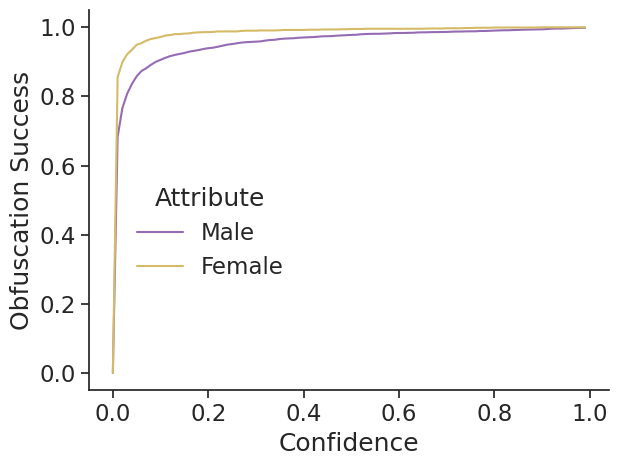}
         \caption{Untargeted AWS: Sex}
         \label{fig:aws-lowkey-sex-appendix}
     \end{subfigure}
     \caption{Untargeted adversarial examples generated using LowKey~\cite{DBLP:LowKey} and evaluated on Face++ and AWS.}
     \label{fig:api-lowkey-appendix}
\end{figure}

\begin{figure}[h]
\centering
     \begin{subfigure}{0.3\columnwidth}
         \centering
         \includegraphics[width=\textwidth]{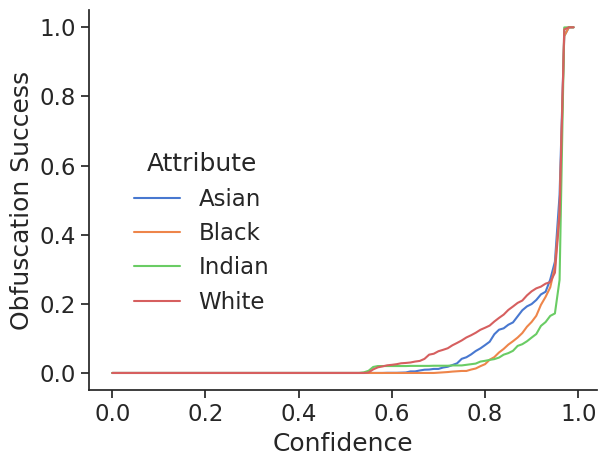}
         \caption{Targeted Face++: Race}
         \label{fig:targeted-facepp-lowkey-race-appendix}
     \end{subfigure}
     \begin{subfigure}{0.3\columnwidth}
         \centering
         \includegraphics[width=\textwidth]{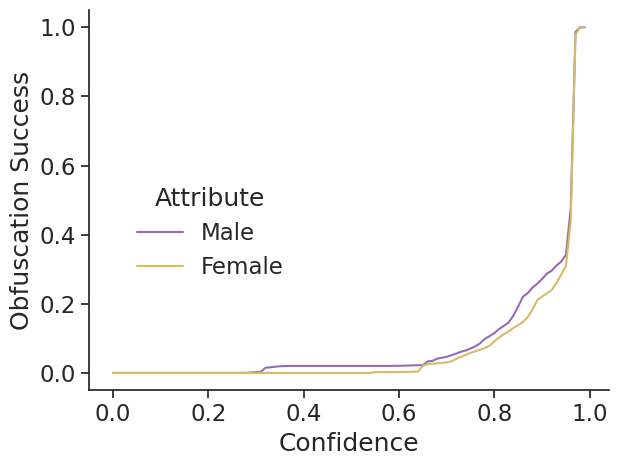}
         \caption{Targeted Face++: Sex}
         \label{fig:targeted-facepp-lowkey-sex-appendix}
     \end{subfigure}
     \begin{subfigure}{0.3\columnwidth}
         \centering
         \includegraphics[width=\textwidth]{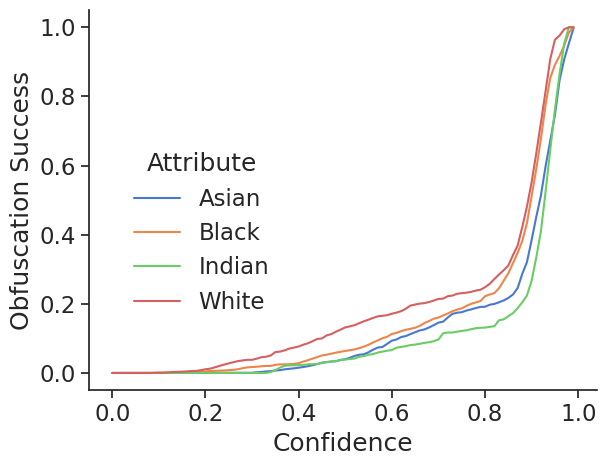}
         \caption{Targeted Azure: Race}
         \label{fig:targeted-azure-lowkey-race-appendix}
     \end{subfigure}
     
     \begin{subfigure}{0.3\columnwidth}
         \centering
         \includegraphics[width=\textwidth]{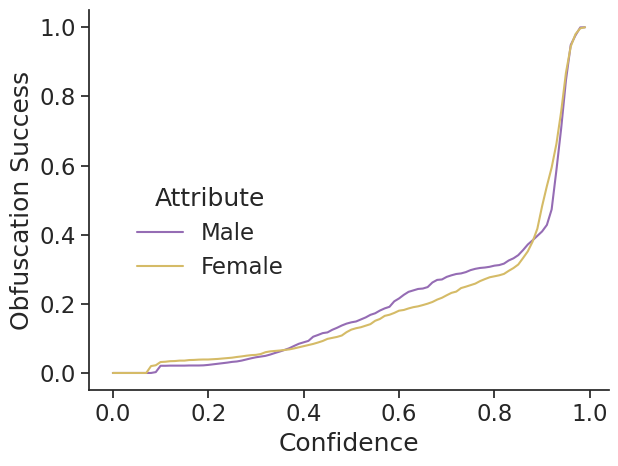}
         \caption{Targeted Azure: Sex}
         \label{fig:targeted-azure-lowkey-sex-appendix}
     \end{subfigure}
     \begin{subfigure}{0.3\columnwidth}
         \centering
         \includegraphics[width=\textwidth]{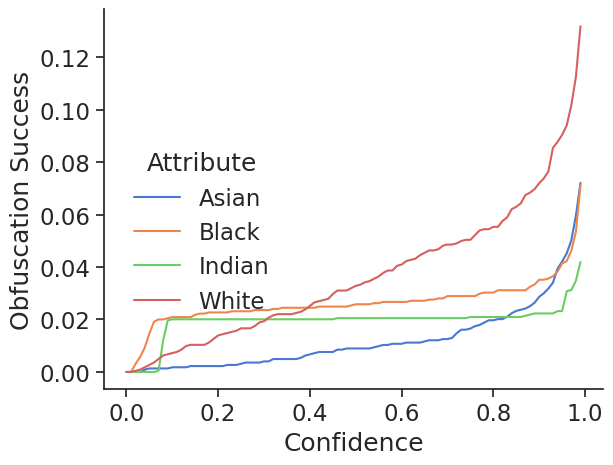}
         \caption{Targeted AWS: Race}
         \label{fig:targeted-aws-lowkey-race-appendix}
     \end{subfigure}
     \begin{subfigure}{0.3\columnwidth}
         \centering
         \includegraphics[width=\textwidth]{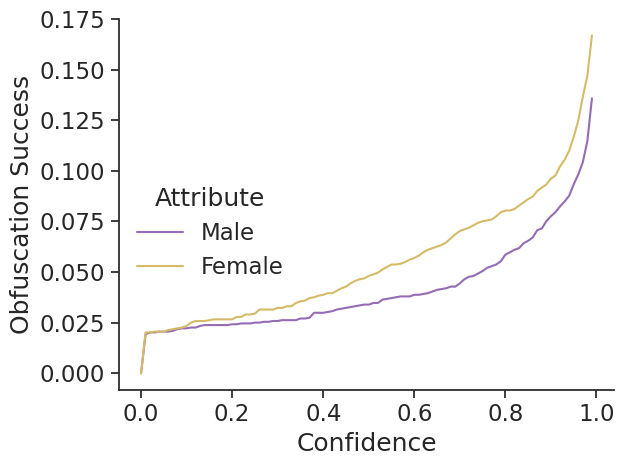}
         \caption{Targeted AWS: Sex}
         \label{fig:targeted-aws-lowkey-sex-appendix}
     \end{subfigure}
     \caption{Targeted adversarial examples generated using Face-Off~\cite{DBLP:LowKey} and evaluated on Face++, Azure, and AWS.}
     \label{fig:facepp-faceoff-appendix}
\end{figure}

\begin{figure}[t]
     \centering
     \begin{subfigure}{0.48\columnwidth}
         \centering
         \includegraphics[width=\textwidth]{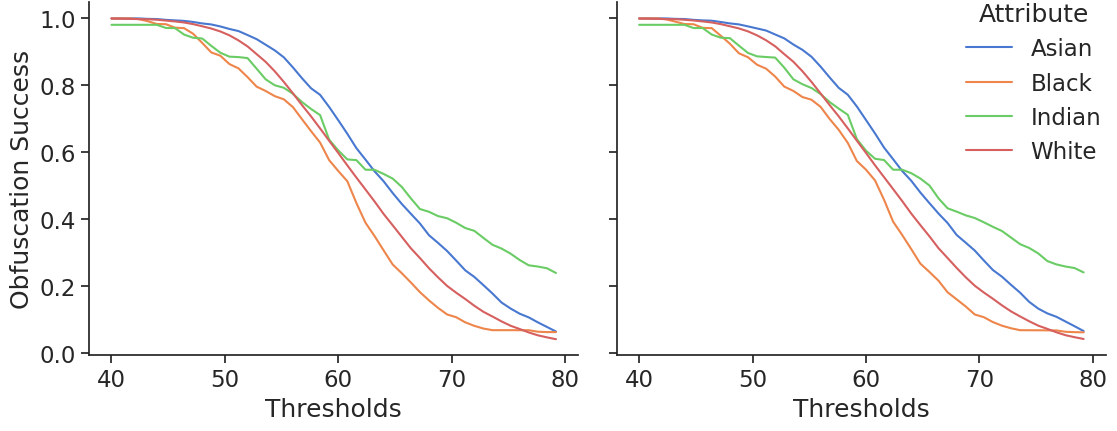}
         \caption{DeepFace: Different Race (Left) vs. Same Race (Right)}
         \label{fig:deepface_race}
     \end{subfigure}
     \begin{subfigure}{0.48\columnwidth}
         \centering
         \includegraphics[width=\textwidth]{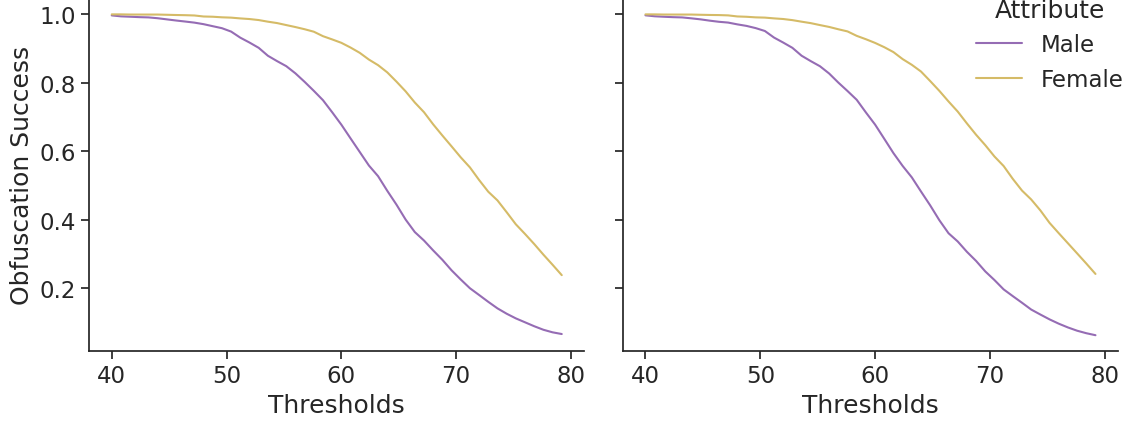}
         \caption{DeepFace: Different Sex (Left) vs. Same Sex (Right)}
         \label{fig:deepface_sex}
     \end{subfigure}
     \begin{subfigure}{0.48\columnwidth}
         \centering
         \includegraphics[width=\textwidth]{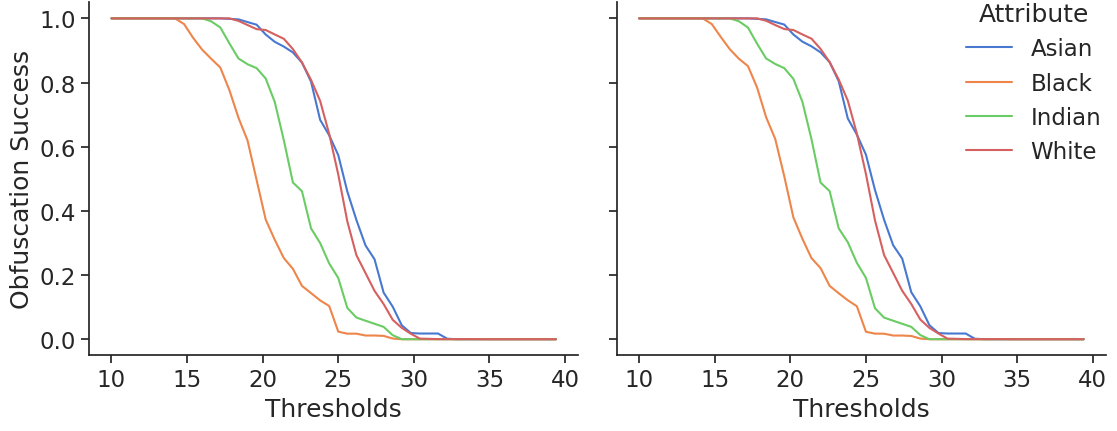}
         \caption{ArcFace: Different Race (Left) vs. Same Race (Right)}
         \label{fig:arcface_race}
     \end{subfigure}
     \begin{subfigure}{0.48\columnwidth}
         \centering
         \includegraphics[width=\textwidth]{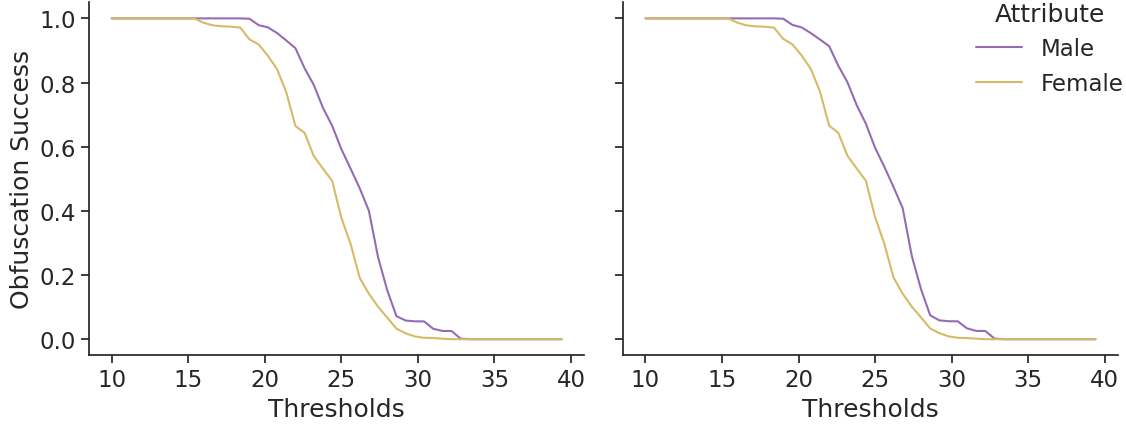}
         \caption{ArcFace: Different Sex (Left) vs. Same Sex (Right)}
         \label{fig:arcface_sex}
     \end{subfigure}
     \begin{subfigure}{0.48\columnwidth}
         \centering
         \includegraphics[width=\textwidth]{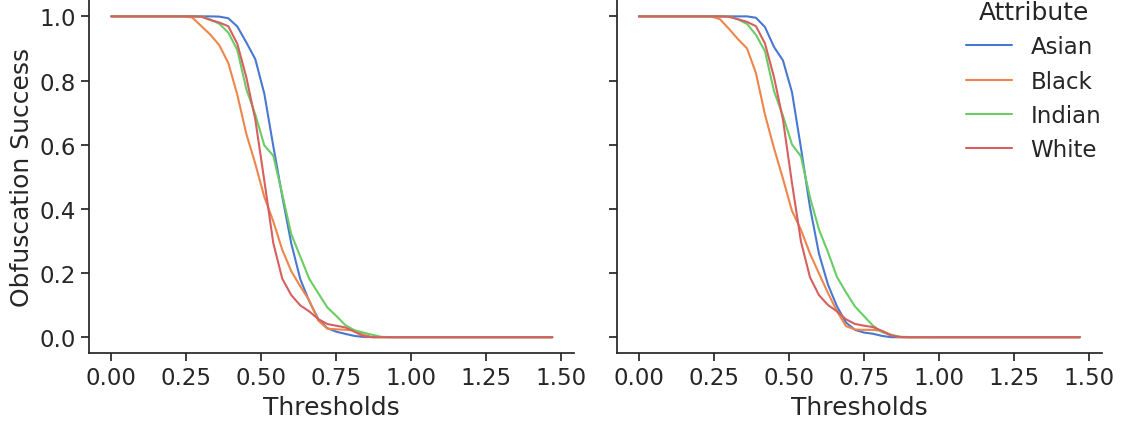}
         \caption{VGGFace: Different Race (Left) vs. Same Race (Right)}
         \label{fig:vggface_race}
     \end{subfigure}
     \begin{subfigure}{0.48\columnwidth}
         \centering
         \includegraphics[width=\textwidth]{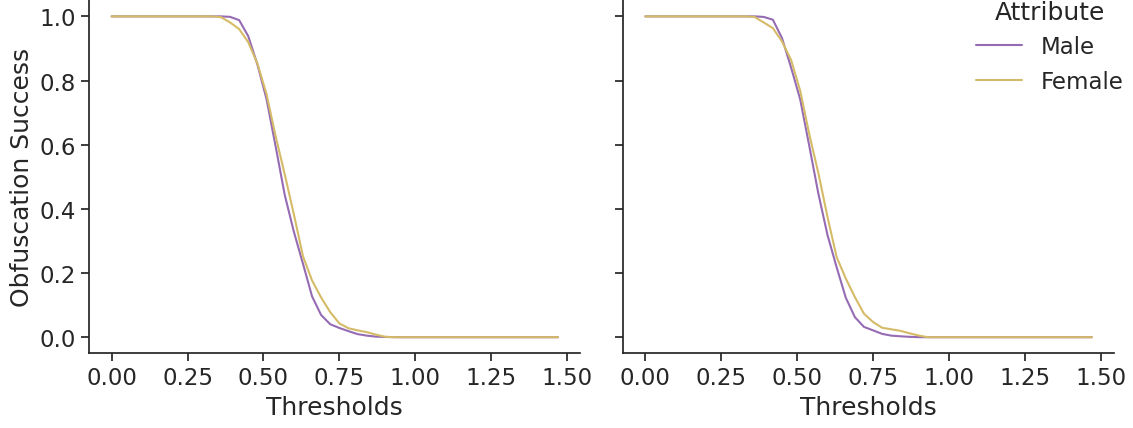}
         \caption{VGGFace: Different Sex (Left) vs. Same Sex (Right)}
         \label{fig:vggface_sex}
     \end{subfigure}
     \caption{Targeted obfuscation success evaluated on 3 metric embedding networks in a black-box setting.}
     \label{fig:transferability_success_appendix}
\end{figure}

\subsection{Targeted and Untargeted Obfuscation Success Rates}

\begin{figure}[h]
\centering
     \begin{subfigure}[b]{0.6\columnwidth}
         \centering
         \includegraphics[width=\textwidth]{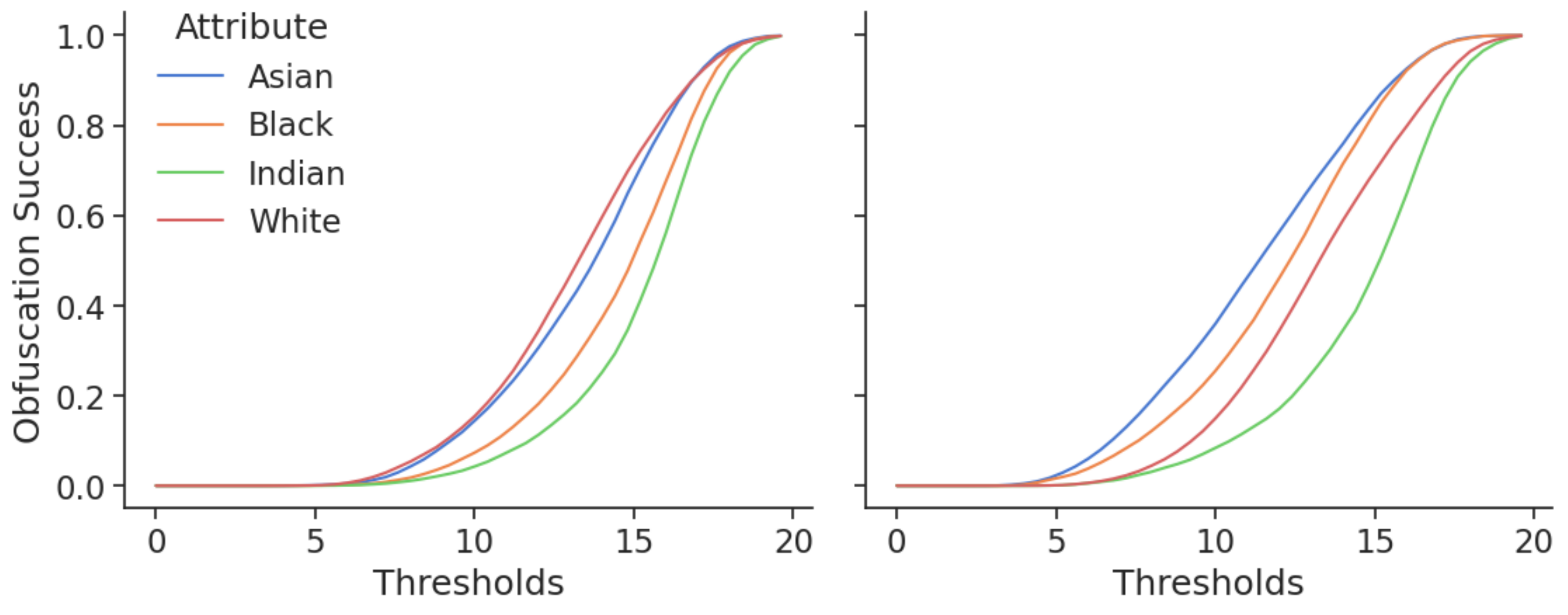}
         \caption{Obfuscation success rates on FaceNet: Different (Left) vs. Same (Right) Race}
         \label{fig:untargeted_race}
     \end{subfigure}
     
     \begin{subfigure}[b]{0.6\columnwidth}
         \centering
         \includegraphics[width=\textwidth]{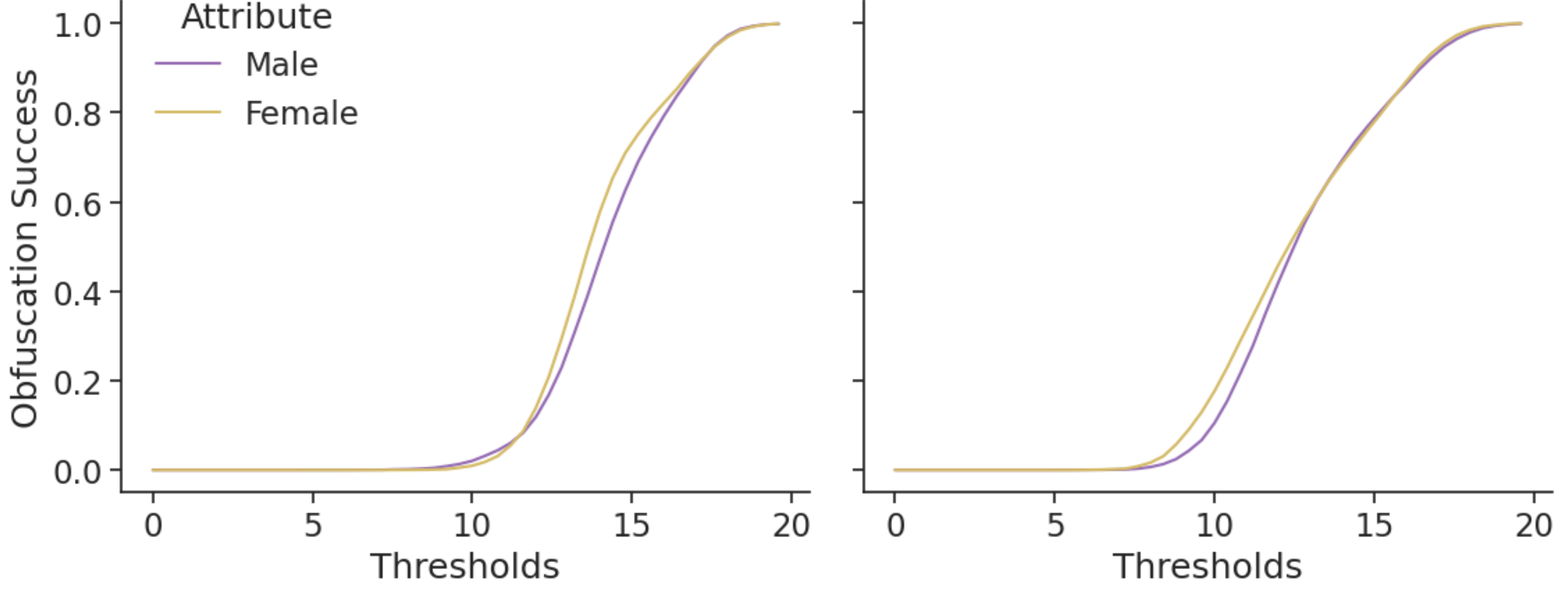}
         \caption{Obfuscation success rates on FaceNet: Different (Left) vs. Same (Right) Sex}
         \label{fig:untargeted_sex}
     \end{subfigure}
     \caption{Untargeted obfuscation success evaluated on the FaceNet metric embedding network in a white-box setting.}
     \label{fig:untargeted_success}
\end{figure}

\begin{figure}[h]
\centering
     \begin{subfigure}[b]{0.6\columnwidth}
         \centering
         \includegraphics[width=\textwidth]{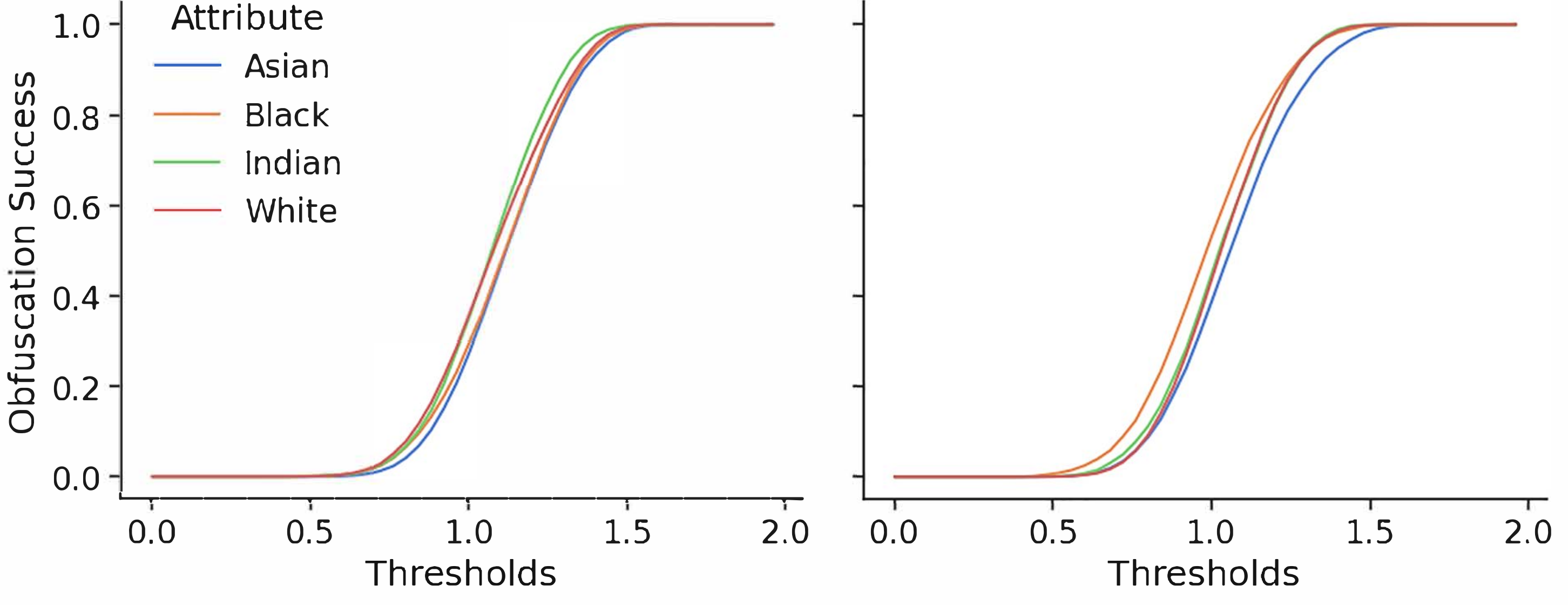}
         \caption{Obfuscation success rates on OpenFace: Different (Left) vs. Same (Right) Race}
         \label{fig:untargeted_race_openface}
     \end{subfigure}
     
     \begin{subfigure}[b]{0.6\columnwidth}
         \centering
         \includegraphics[width=\textwidth]{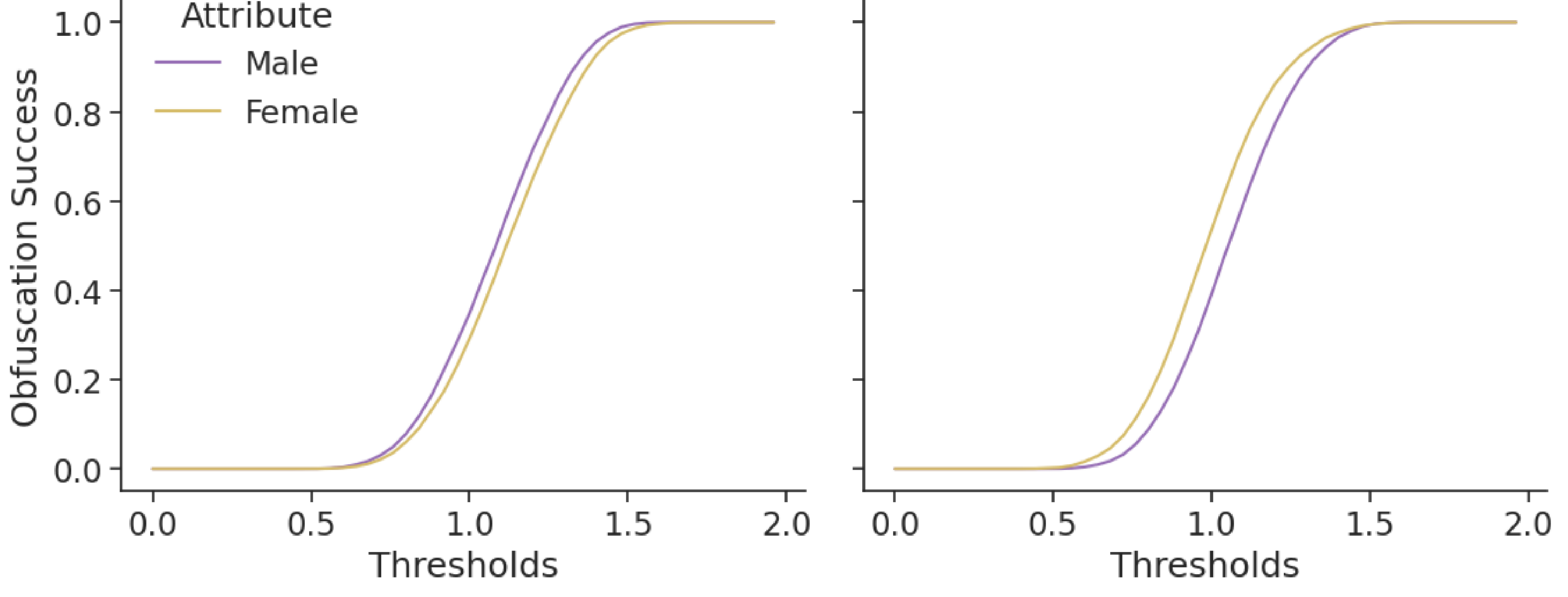}
         \caption{Obfuscation success rates on OpenFace: Different (Left) vs. Same (Right) Sex}
         \label{fig:untargeted_sex_openface}
     \end{subfigure}
     \caption{Untargeted obfuscation success evaluated on the OpenFace metric embedding network in a black-box setting.}
     \label{fig:untargeted_success_openface}
\end{figure}

The results depicted in \cref{fig:untargeted_success} portray the untargeted obfuscation success rates for the pre-trained Facenet model. These untargeted obfuscation success rates are also provided for the black-box setting with the OpenFace model (\cref{fig:untargeted_success_openface}). In \cref{fig:targeted_balance}, the targeted success rates are provided for the Race-Balanced and Sex-Balanced Facenet models. We observe similar trends as discussed in \cref{subsec:obfuscationEffectiveness,subsec:bias_mitigation}.

\begin{figure}[t]
     \centering
     \begin{subfigure}[b]{0.23\textwidth}
         \centering
         \includegraphics[width=\textwidth]{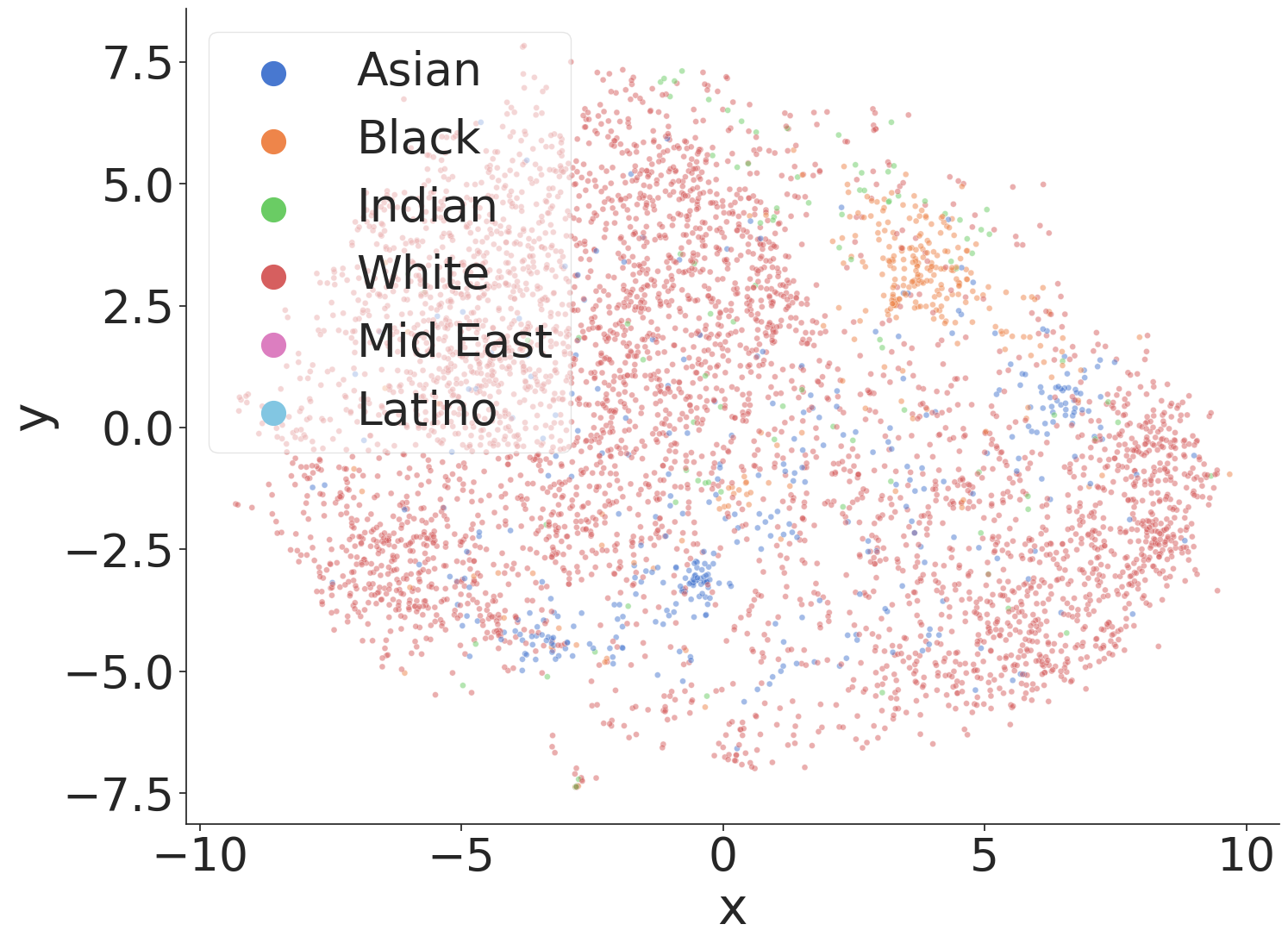}
         \caption{LFW - Race}
         \label{fig:tsne_lfw_race_reference}
     \end{subfigure}
     \begin{subfigure}[b]{0.23\textwidth}
         \centering
         \includegraphics[width=\textwidth]{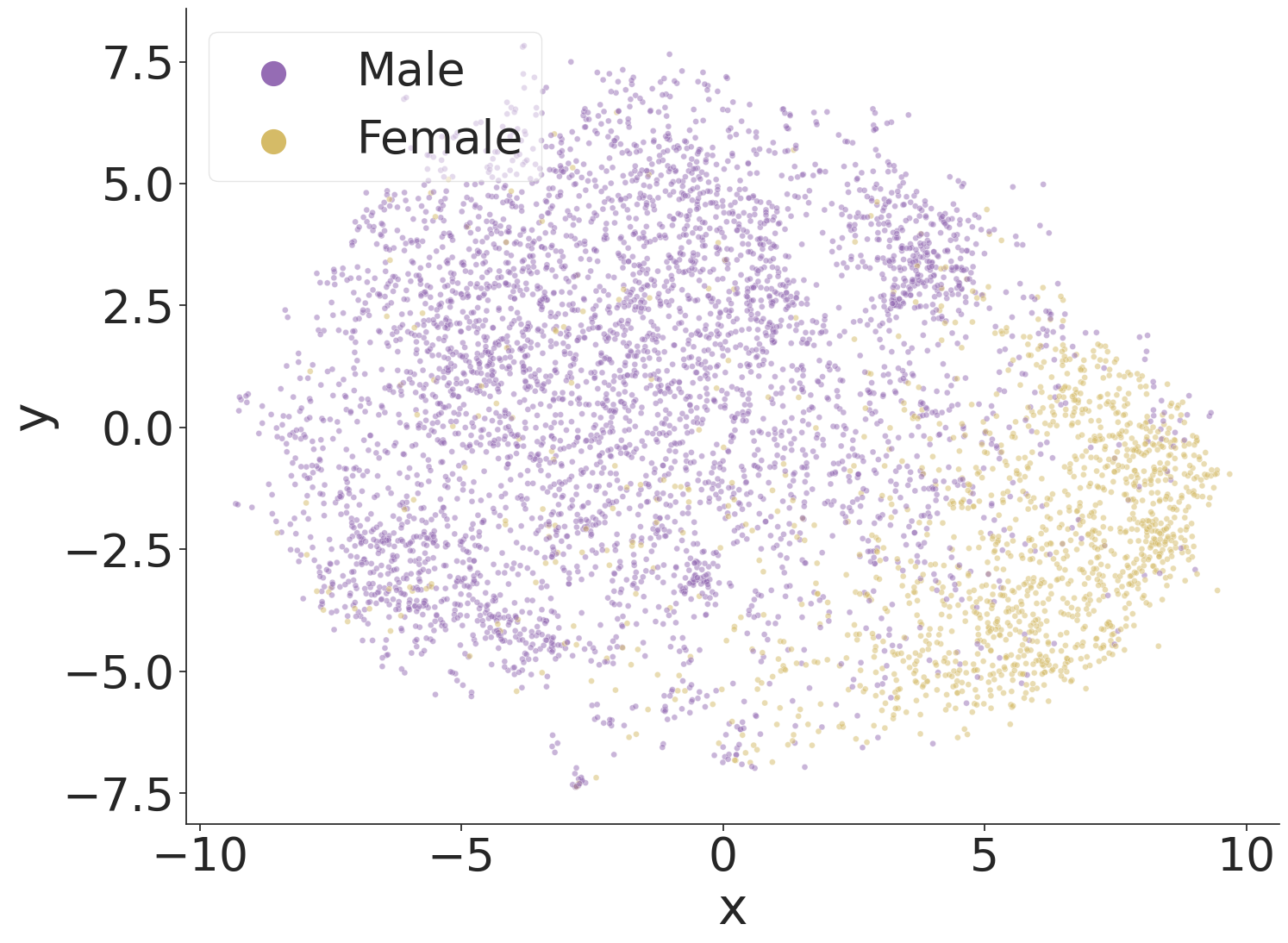}
         \caption{LFW - Sex}
         \label{fig:tsne_lfw_sex_reference}
     \end{subfigure}
     \begin{subfigure}[b]{0.23\textwidth}
         \centering
         \includegraphics[width=\textwidth]{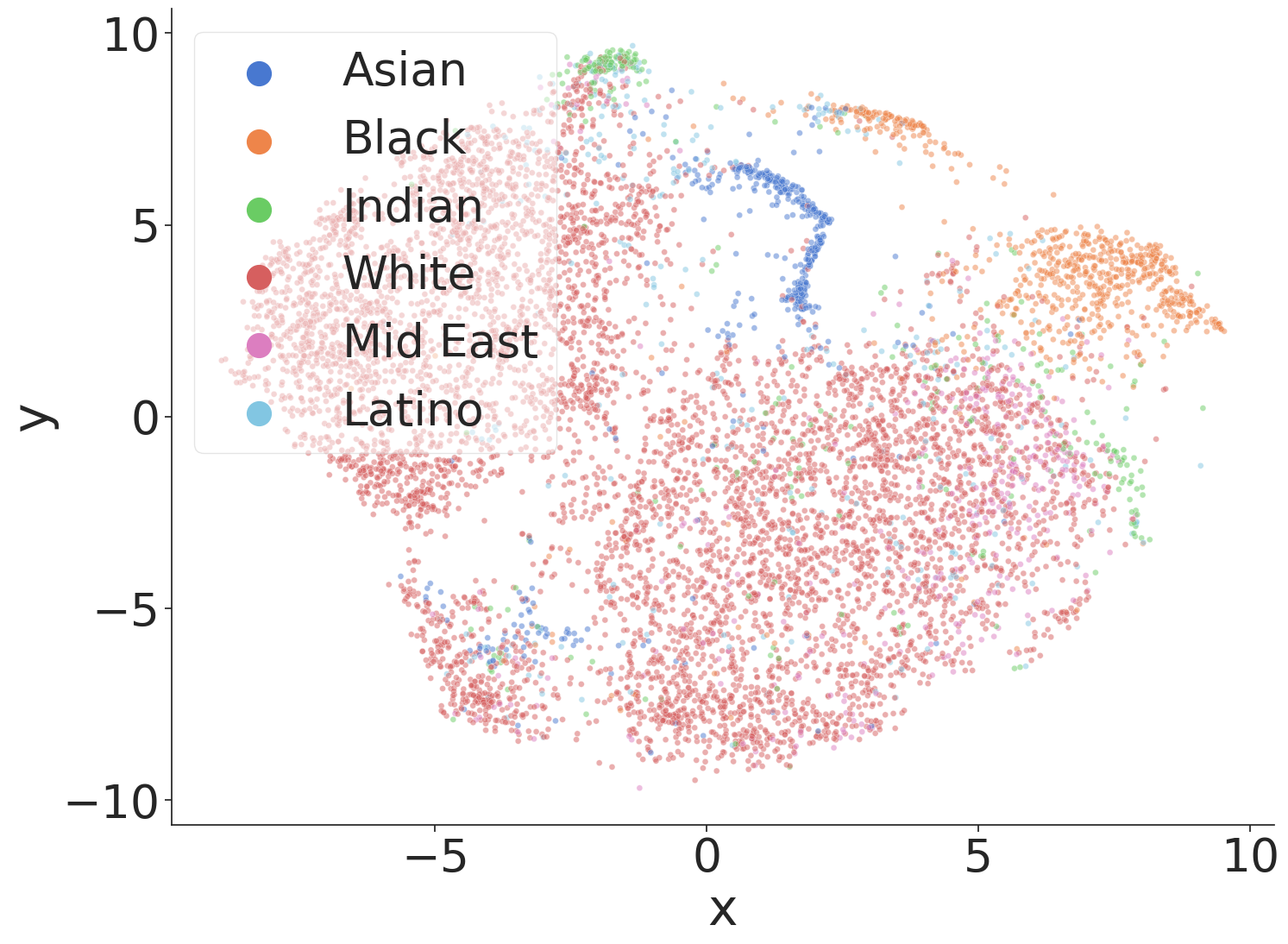}
         \caption{VGGFace2 - Race}
         \label{fig:tsne_vggface2_race_reference}
     \end{subfigure}
     \begin{subfigure}[b]{0.23\textwidth}
         \centering
         \includegraphics[width=\textwidth]{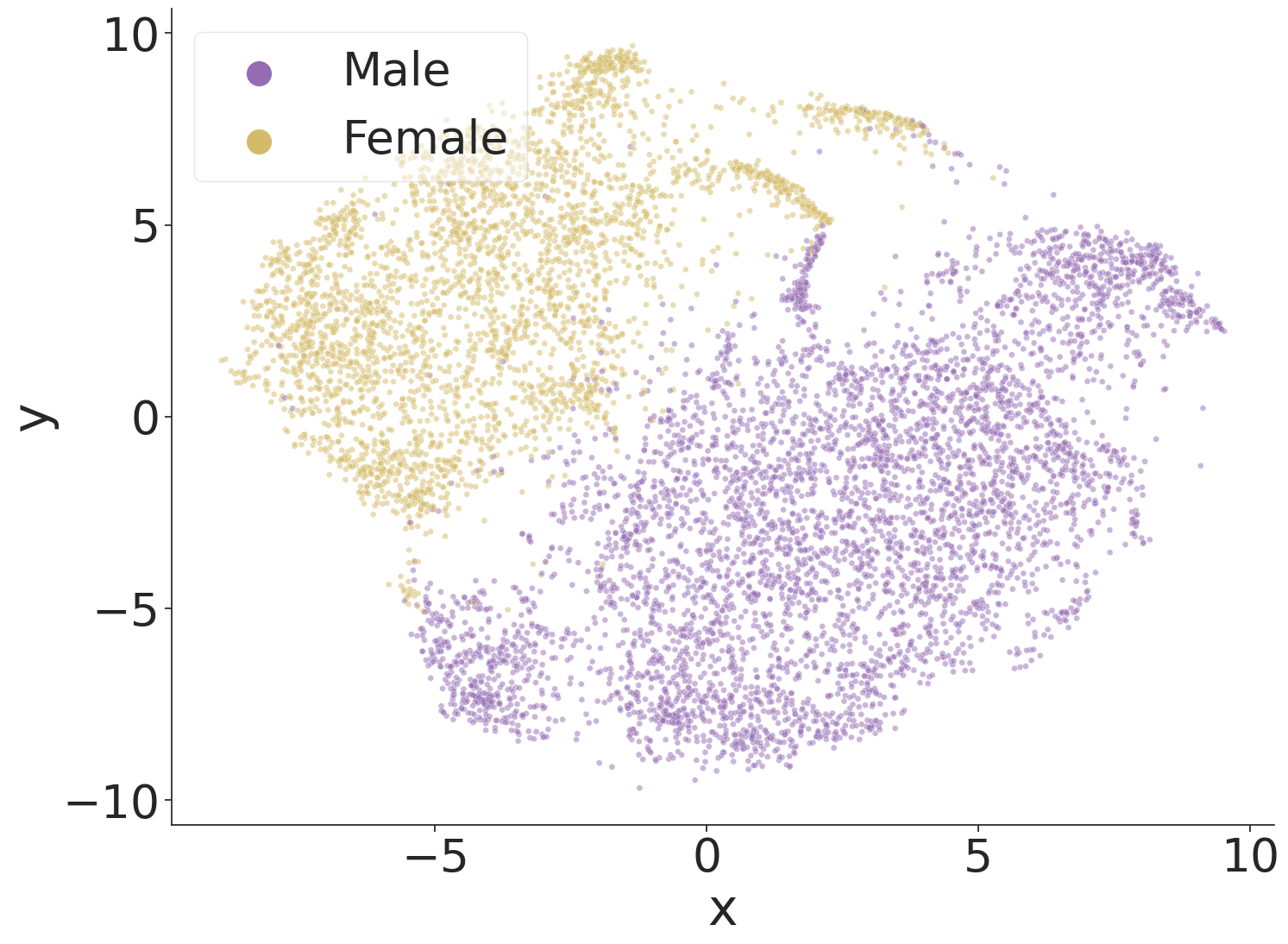}
         \caption{VGGFace2 - Sex }
         \label{fig:tsne_vggface2_sex_reference}
     \end{subfigure}
     \caption{t-SNE~\cite{maaten2008tsne} for the Reference FaceNet. These figures visualize embedding spaces generated of LFW~\cite{LFWTech} and VGGFace2~\cite{caovggface2} datasets. Embeddings of identities are colored by demographic group. The clustering behavior observed in \cref{fig:tsne} less disparate clusters for each demographic group, some clusters are more spread out than others.}
     \label{fig:tsne_reference}
\end{figure}

\begin{figure}[h]
\centering
     \begin{subfigure}[b]{0.6\columnwidth}
         \centering
         \includegraphics[width=\textwidth]{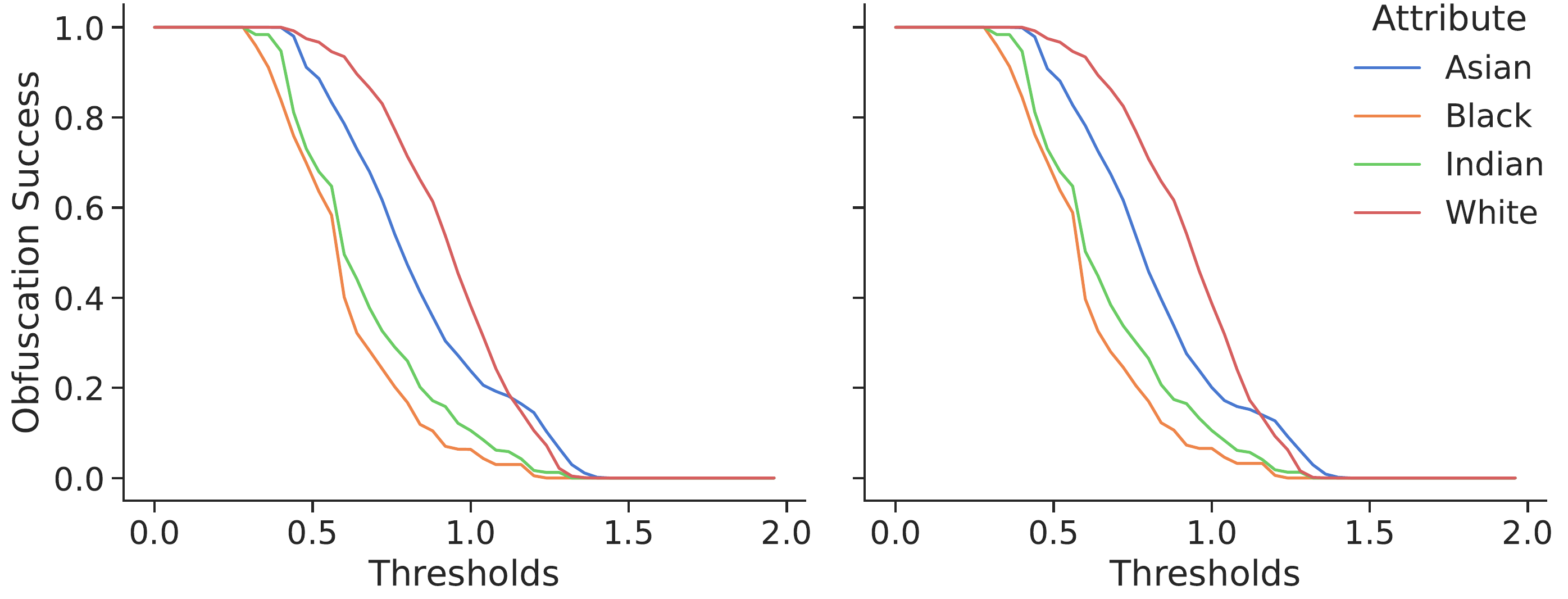}
         \caption{Obfuscation success rates on the Reference FaceNet: Different (Left) vs. Same (Right) Race}
         \label{fig:targeted_race_imbalance}
     \end{subfigure}
     
     \begin{subfigure}[b]{0.6\columnwidth}
         \centering
         \includegraphics[width=\textwidth]{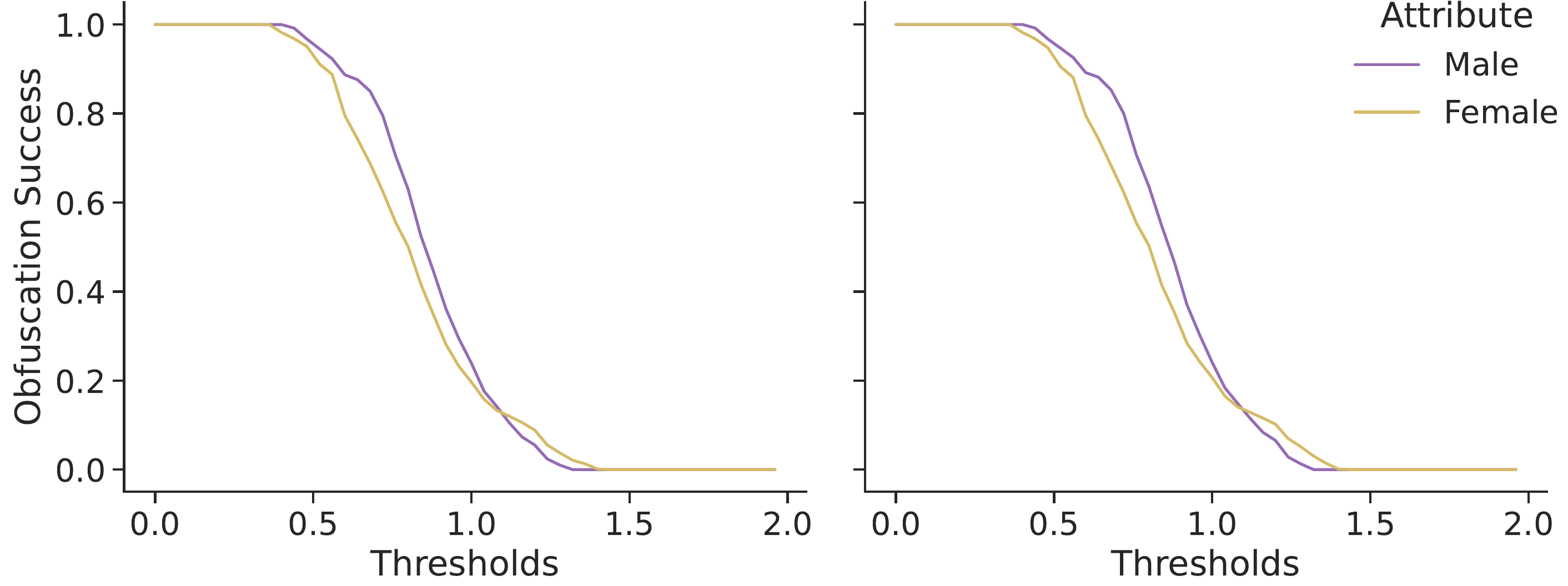}
         \caption{Obfuscation success rates on the Reference FaceNet: Different (Left) vs. Same (Right) Sex}
         \label{fig:targeted_sex_imbalance}
     \end{subfigure}
     \caption{Targeted obfuscation success evaluated on the reference FaceNet in a white-box setting.}
     \label{fig:targeted_imbalance}
\end{figure}

\begin{figure}[t]
      \centering
     \begin{subfigure}{0.48\columnwidth}
         \centering
         \includegraphics[width=0.8\textwidth]{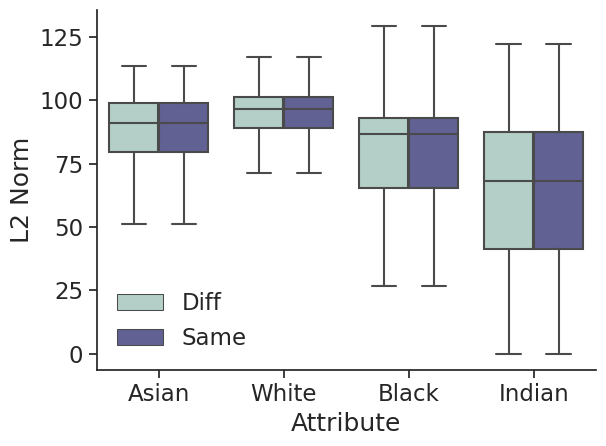}
         \caption{Reference FaceNet on race data}
         \label{fig:pert_norm_imbalance_race}
     \end{subfigure}
     \begin{subfigure}{0.48\columnwidth}
         \centering
         \includegraphics[width=0.8\textwidth]{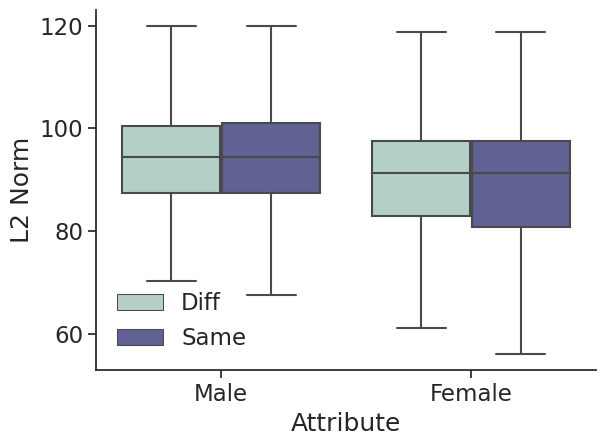}
         \caption{Reference FaceNet on sex data}
         \label{fig:pert_norm_imbalance_sex}
     \end{subfigure}
        \caption{The distribution of adversarial perturbation sizes generated using the CW~\cite{carlini2017towards} attack on the reference FaceNet.}
        \label{fig:pert_norm_imbalance}
\end{figure}

\subsection{Additional TCAV Experiments}\label{subsec:tcav_appendix}

We also evaluate TCAV on a larger subset of VGGFace2 with 4102 identities. However, most identities fall under Type 4 or Type 5 skin tones, a result of the uneven distribution of VGGFace2. Our results indicate that skin tone is a significant concept used in predicting the identities of faces. 


\begin{figure}[b]
     \centering
     \begin{subfigure}[b]{0.3\textwidth}
         \centering
         \includegraphics[width=\textwidth]{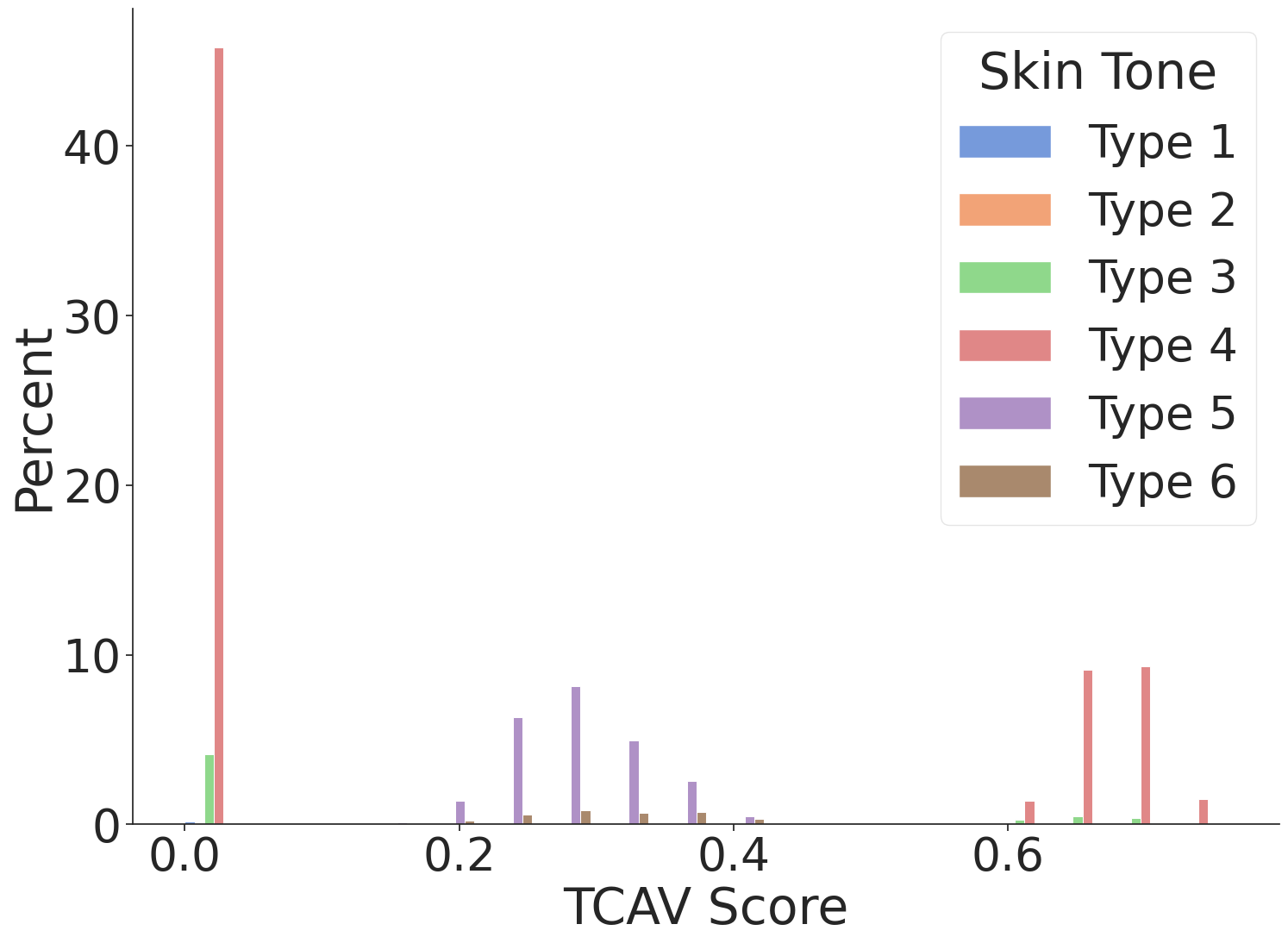}
         \caption{Block 17 Activation 9}
         \label{fig:tcav_big_1_appendix}
     \end{subfigure}
     \begin{subfigure}[b]{0.3\textwidth}
         \centering
         \includegraphics[width=\textwidth]{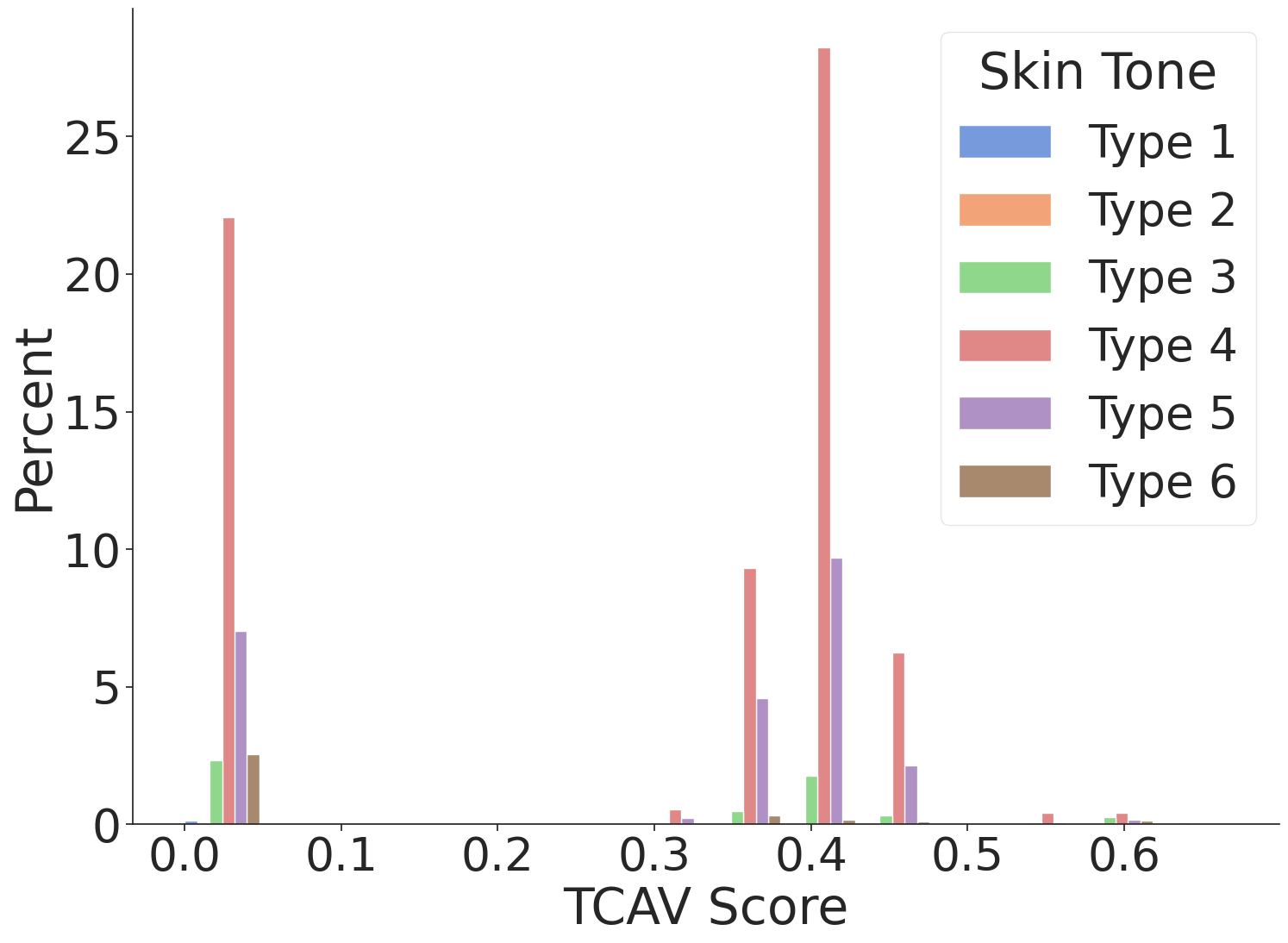}
         \caption{Block 35 Activation 3}
         \label{fig:tcav_big_2_appendix}
     \end{subfigure}
     
     \begin{subfigure}[b]{0.3\textwidth}
         \centering
         \includegraphics[width=\textwidth]{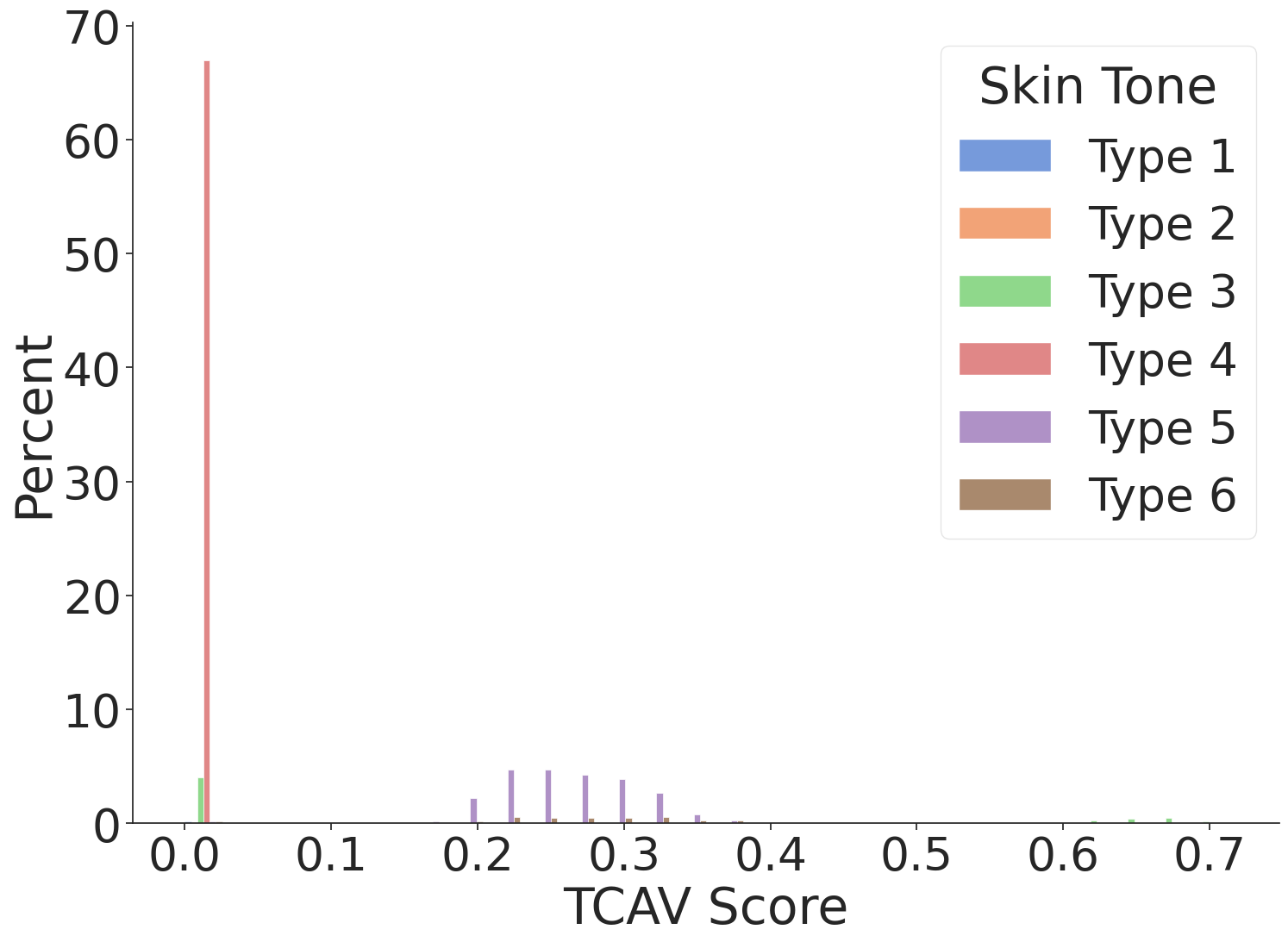}
         \caption{Block 35 Activation 5}
         \label{fig:tcav_big_3_appendix}
     \end{subfigure}
     \begin{subfigure}[b]{0.3\textwidth}
         \centering
         \includegraphics[width=\textwidth]{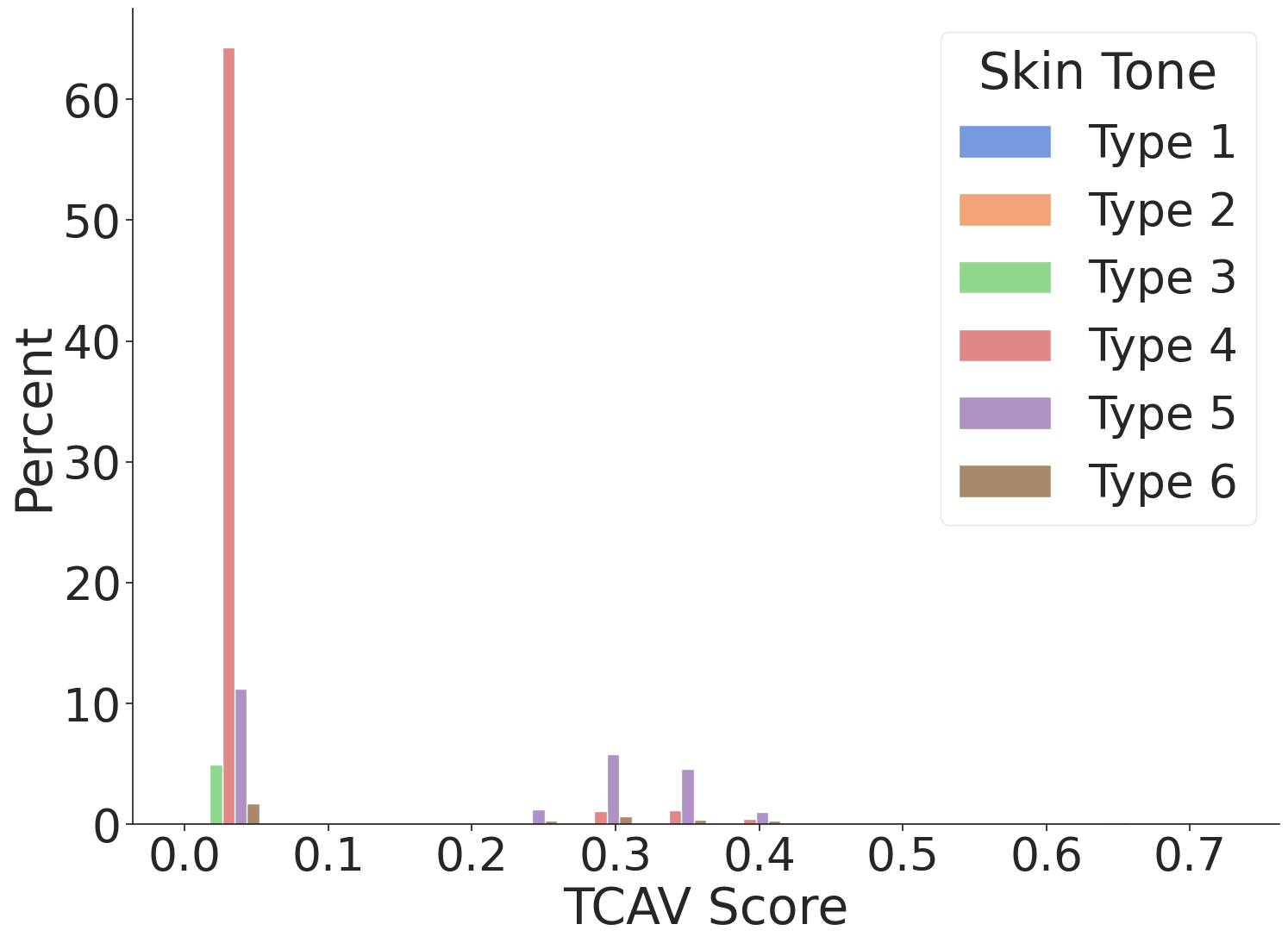}
         \caption{Conv2d 2a 3x3 Activation}
         \label{fig:tcav_big_4_appendix}
     \end{subfigure}
     \caption{Distribution of TCAV scores for 4102 identities from the VGGFace2~\cite{caovggface2} dataset. A TCAV score of above zero indicates that the concept is used for the prediction of the face with a statistically significant impact on the output prediction. Results for 4 layers of the Facenet network are reported.}
     \label{fig:tcav_big}
\end{figure}

\subsection{Additional t-SNE Experiments}\label{subsec:tsne_appendix}

\begin{figure}[t]
     \centering
     \begin{subfigure}[b]{0.3\textwidth}
         \centering
         \includegraphics[width=\textwidth]{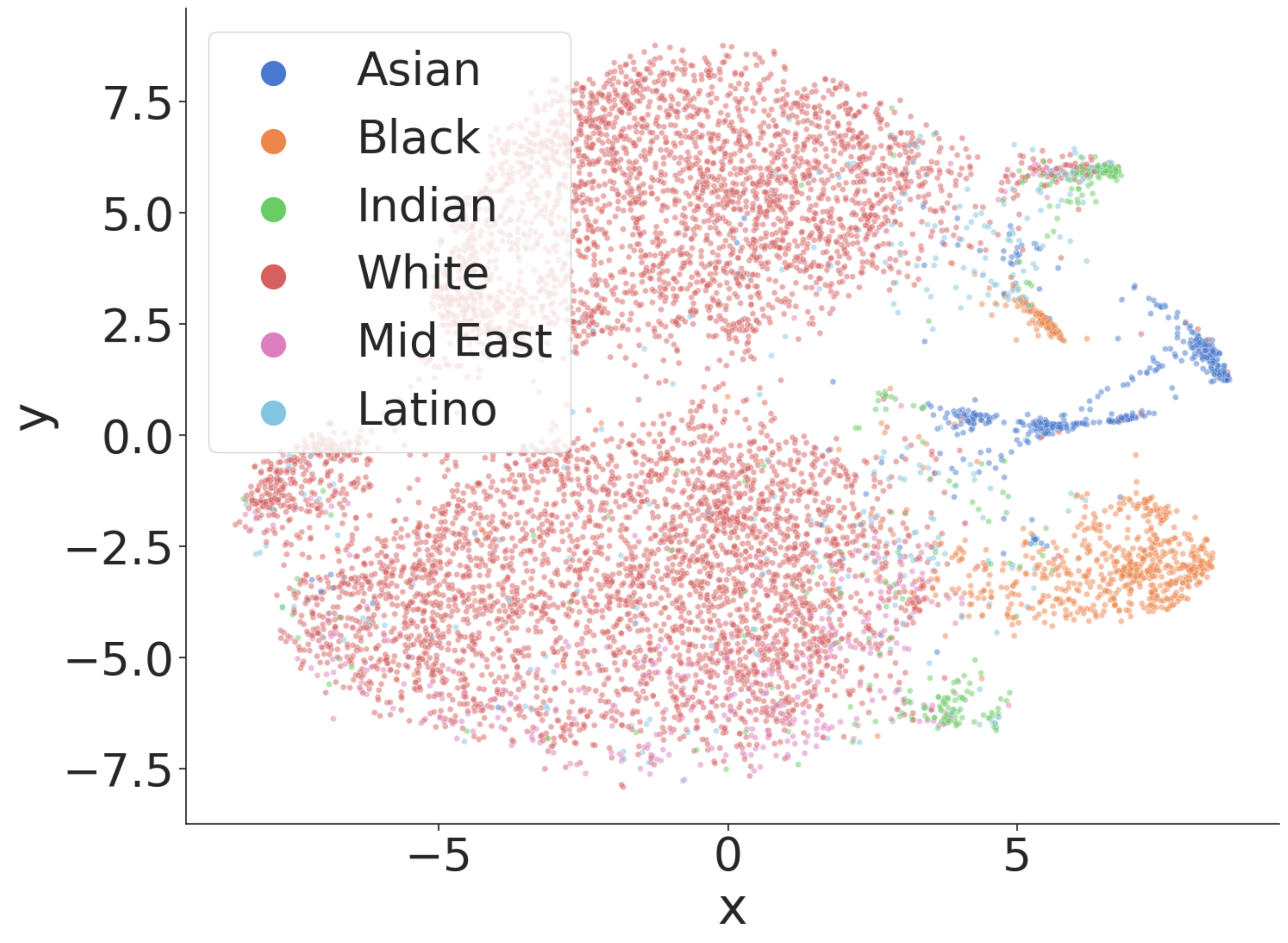}
         \caption{VGGFace2 - Race}
         \label{fig:tsne_vgg_race_appendix}
     \end{subfigure}
     \begin{subfigure}[b]{0.3\textwidth}
         \centering
         \includegraphics[width=\textwidth]{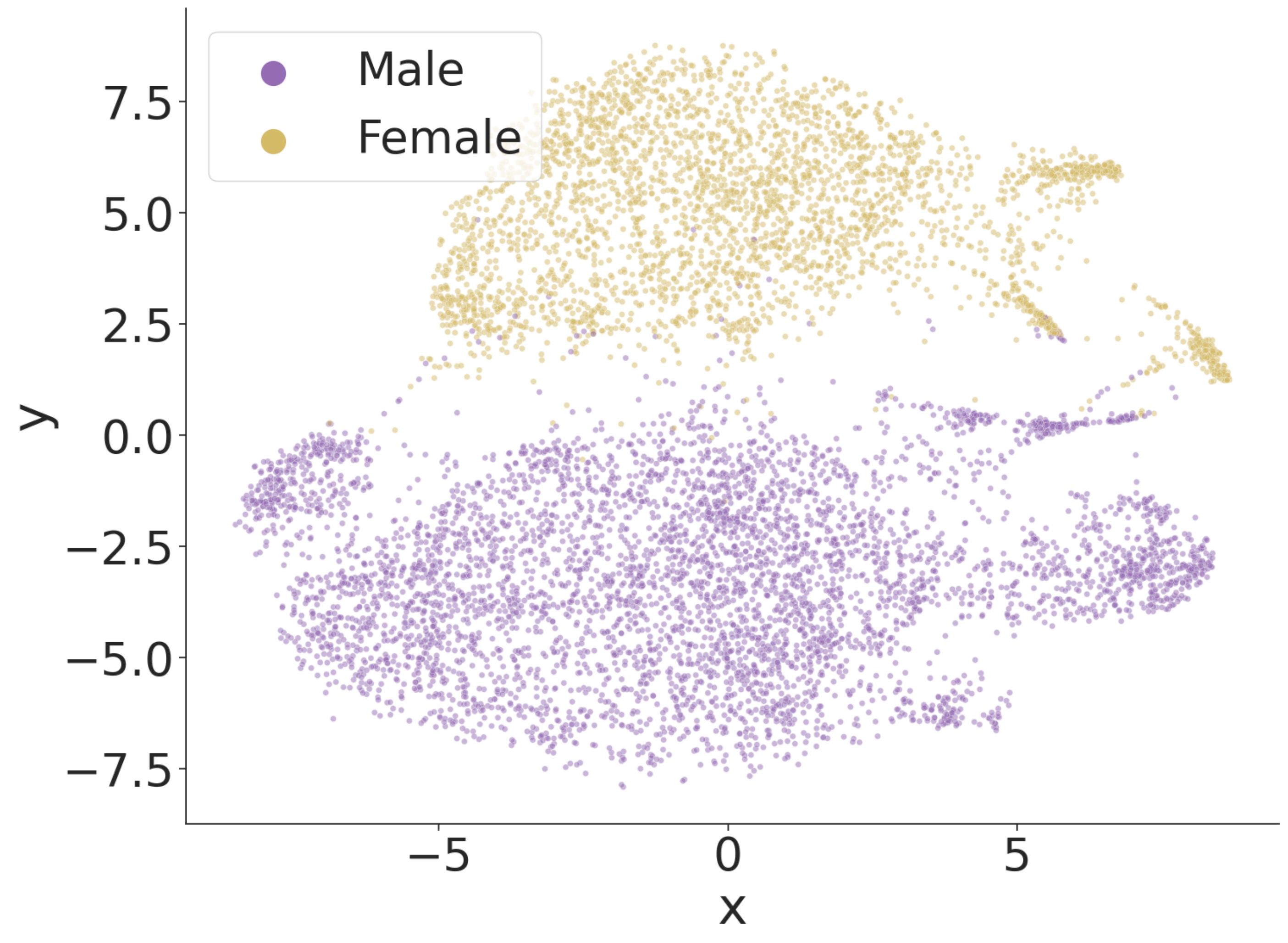}
         \caption{VGGFace2 - Sex}
         \label{fig:tsne_vgg_sex_appendix}
     \end{subfigure}
     \caption{t-SNE of the embedding spaces generated using the VGGFace2 dataset. Embeddings of identities are colored by race/sex. Distinct clusters exist for each demographic group.}
     \label{fig:tsne_appendix}
\end{figure}

\begin{figure}[t]
     \centering
     \begin{subfigure}[b]{0.3\textwidth}
         \centering
         \includegraphics[width=\textwidth]{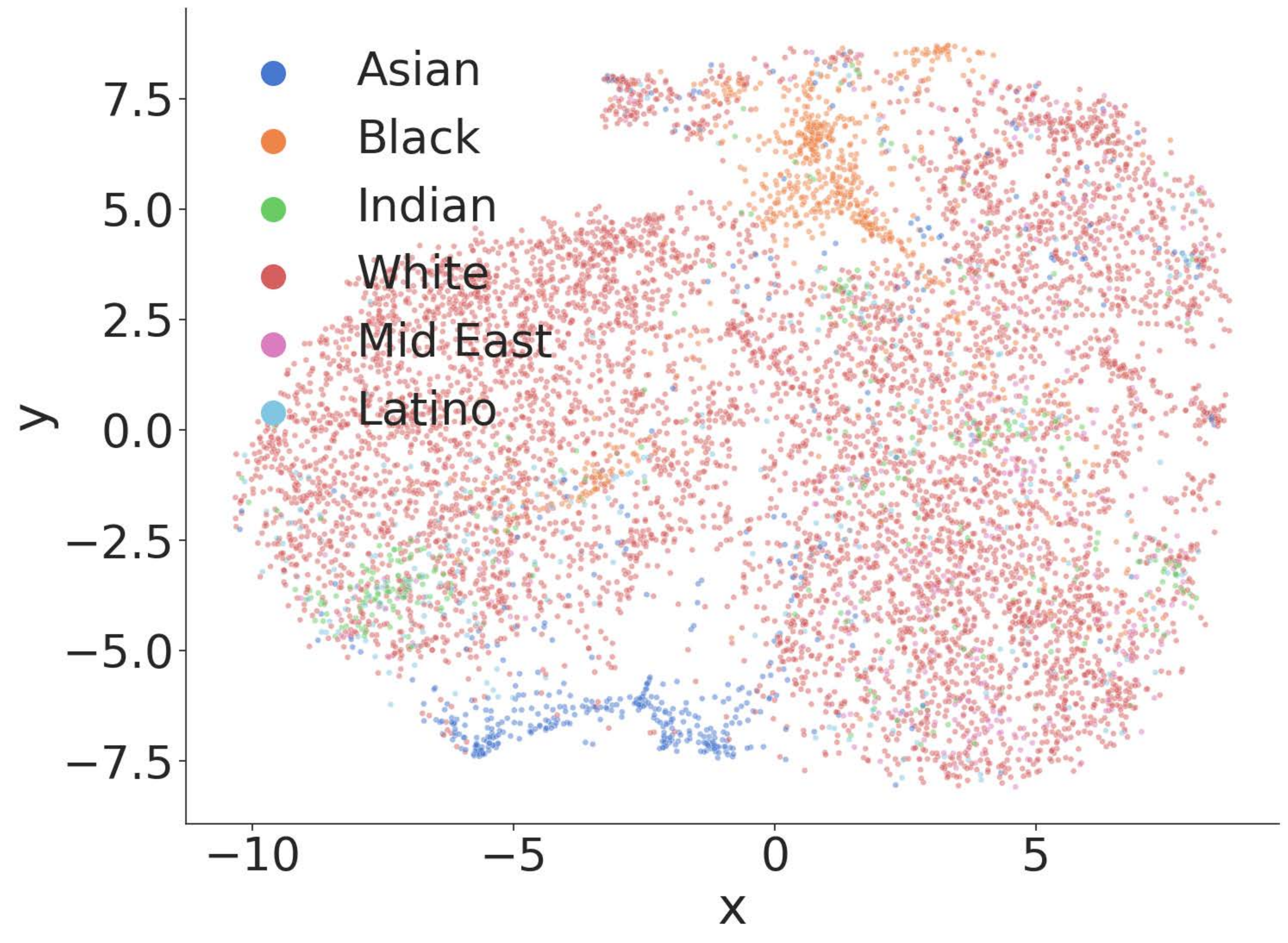}
         \caption{VGGFace2 - Race}
         \label{fig:tsne_vgg_race_xu_appendix}
     \end{subfigure}
     \begin{subfigure}[b]{0.3\textwidth}
         \centering
         \includegraphics[width=\textwidth]{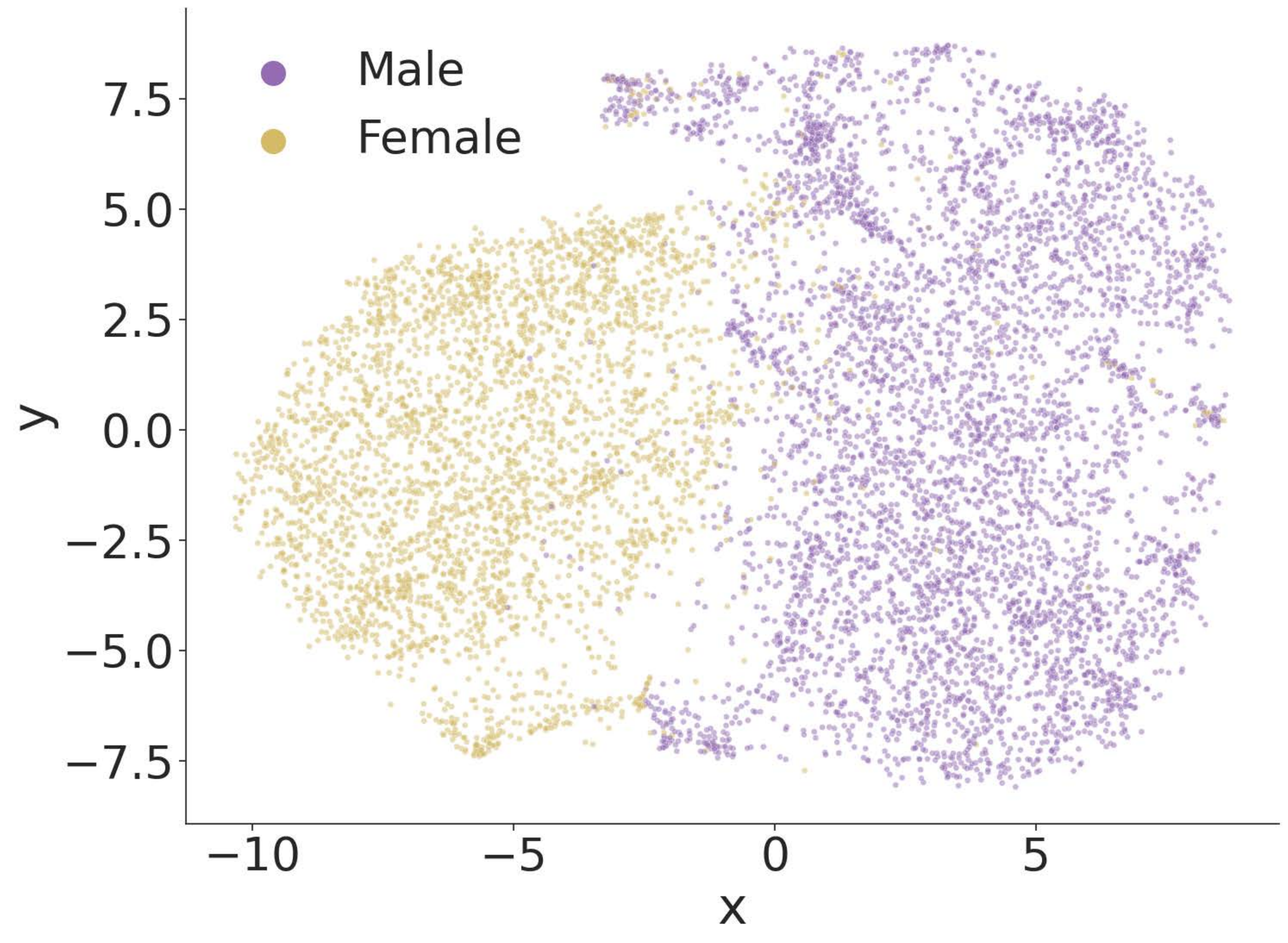}
         \caption{VGGFace2 - Sex}
         \label{fig:tsne_vgg_sex_xu_appendix}
     \end{subfigure}
     \caption{t-SNE for models trained with Xu et al. procedure.  Embeddings of VGGFace2 are depicted. Embeddings of identities are colored by demographic. The clusters are less disparate than those observed in \cref{fig:tsne_appendix}.}
     \label{fig:tsne_xu_appendix}
\end{figure}

\begin{figure}[t]
     \centering
     \begin{subfigure}[b]{0.3\textwidth}
         \centering
         \includegraphics[width=\textwidth]{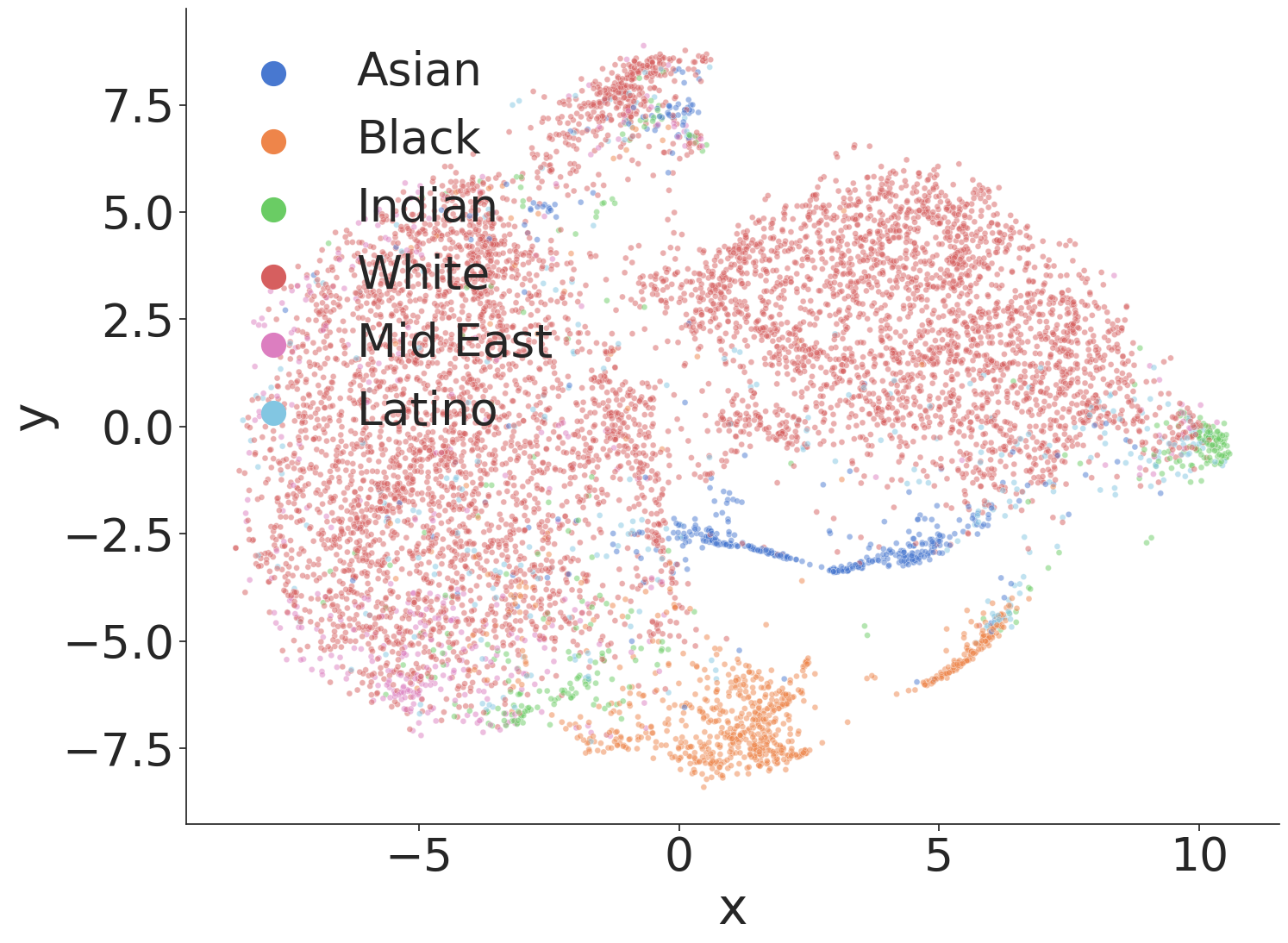}
         \caption{VGGFace2 - Race}
         \label{fig:tsne_vggface2_race_balanced_appendix}
     \end{subfigure}
     \begin{subfigure}[b]{0.3\textwidth}
         \centering
         \includegraphics[width=\textwidth]{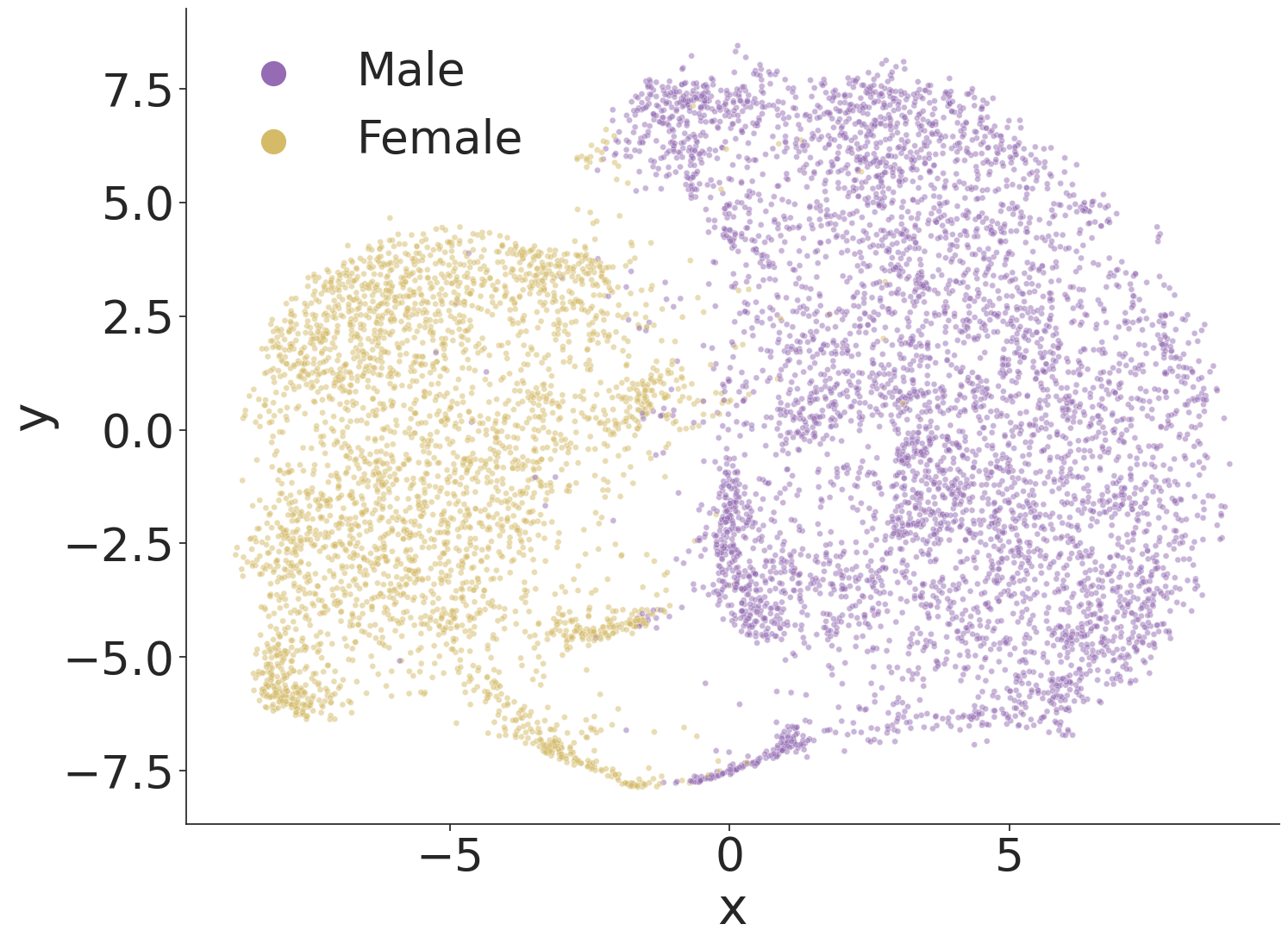}
         \caption{VGGFace2 - Sex }
         \label{fig:tsne_vggface2_sex_balanced_appendix}
     \end{subfigure}
     \caption{t-SNE for models trained with demographically balanced data and visualized for embeddings from VGGFace2. \Cref{fig:tsne_vggface2_race_balanced_appendix} is generated via race-balanced VGGFace2, and \cref{fig:tsne_vggface2_sex_balanced_appendix} is generated via  sex-balanced VGGFace2.}
     \label{fig:tsne_balanced_appendix}
\end{figure}

In addition to visualizing the embedding spaces generated using the LFW dataset on the various models we evaluate, we also generate t-SNE plots for the VGGFace2 dataset. These embedding spaces are depicted in \cref{fig:tsne_appendix,fig:tsne_xu_appendix,fig:tsne_balanced_appendix}. These embeddings demonstrate tighter clusters for each model type, including the models trained using the Xu et al. procedure and the models trained on the race and sex balanced VGGFace2 datasets.

\subsection{Nearest Neighbor Accuracy}\label{subsec:nearestNeighborExperiments}

In addition to $\mathrm{TPR}_{\placeholder}$, we also apply our experimental procedure to nearest neighbor classifiers. In particular, we utilize the procedure outlined in \cref{subsec:ExperimentalSetup}, but instead of examining $\mathrm{TPR}_{\placeholder}$ on pairs of embeddings, we measure accuracy of a nearest neighbors classifier on both natural and adversarial embeddings. The prediction on an embedding $\embeddingFunc(\sample)$ with ground truth identity $\labelElement$ is said to be correct if the nearest neighbor of $\embeddingFunc(\sample)$ is a centroid embedding with identity $\labelElement$. The nearest neighbors accuracy of unperturbed, natural LFW examples is depicted in  \cref{table:benign_lfw_nearest_neighbor_accuracy}. \cref{table:benign_lfw_nearest_neighbor_accuracy} computes nearest neighbors accuracy on all identities in LFW.

\Cref{table:NearestNeighbors_same,table:NearestNeighbors_diff} compute the nearest neighbor accuracy on the identities described with the ``Same demographic'' and ``Different demographic'' sections of \cref{subsec:ExperimentalSetup}. \Cref{table:NearestNeighbors_same} measures nearest neighbors accuracy for perturbed examples targeting identities within the same demographic. \Cref{table:NearestNeighbors_diff} measures nearest neighbors accuracy for perturbed examples targeting identities in the different demographics.

Some rows in both \cref{table:NearestNeighbors_same,table:NearestNeighbors_diff} depict nearest neighbor accuracy of embeddings generated on perturbed faces in both the black-box setting and white-box settings. The ``generated on'' metric embedding network is the network upon which targeted obfuscatory perturbations are generated. The ``targeted model'' is the emedding function applied to the perturbed faces.


\begin{table*}[phbt]
\footnotesize
\begin{center}

\begin{tabular}{llrrrrrr}
\toprule
                                        {\bf Input} &                            {\bf Male} &   {\bf Female} &    {\bf White} &    {\bf Asian} &    {\bf Black} &   {\bf Indian} \\
\midrule
                                  Pre-trained Facenet                   & 1.000000 & 1.000000 & 1.000000 & 1.000000 & 1.000000 & 1.000000 \\
                                  Reference Facenet                     & 0.983740 & 0.954955 & 0.986014 & 0.990909 & 1.000000 & 1.000000 \\
                        Xu et al. Procedure                     & 0.552846 & 0.672131 & 0.560241 & 0.609091 & 0.675676 & 0.886364 \\
          Race-Balanced &                    0.991870 & 0.981982 & 0.986014 & 0.990909 & 1.000000 & 1.000000 \\
            Sex-Balanced &                     1.000000 & 0.945946 & 0.972028 & 0.972727 & 0.972973 & 1.000000 \\
\bottomrule
\end{tabular}
\end{center}
\caption{Benign Accuracy, all identities in LFW} \label{table:benign_lfw_nearest_neighbor_accuracy}

\end{table*}





\begin{table*}[phbt]
\footnotesize
\begin{center}

\begin{tabular}{llrrrrrr}
\toprule
                                         {\bf Generated On} &                         {\bf Targeted Network} &     {\bf Male} &   {\bf Female} &    {\bf White} &    {\bf Asian} &    {\bf Black} &   {\bf Indian} \\
\midrule
                        Pre-trained Facenet &                        Pre-trained Facenet & 0.724352 & 0.667893 & 0.705762 & 0.670130 & 0.785059 & 0.895949 \\
              Xu et al. &                        Pre-trained Facenet & 0.995333 & 0.990667 & 0.974479 & 0.998077 & 1.000000 & 0.993137 \\
              Xu et al. &               Xu et al.  & 0.688015 & 0.678711 & 0.677117 & 0.590043 & 0.871300 & 0.939394 \\
 Race-Balanced &                        Pre-trained Facenet & 0.975610 & 0.949928 & 0.981559 & 0.990847 & 1.000000 & 1.000000  \\
 Race-Balanced & Race-Balanced & 0.909408 & 0.885413 & 0.897140 & 0.836116 & 0.999724 & 0.977273\\
 Sex-Balanced &                        Pre-trained Facenet & 0.996000 & 0.988000 & 0.990104 & 0.999359 & 1.000000 & 0.992157 \\
  Sex-Balanced &  Sex-Balanced & 0.869255 & 0.736701 & 0.799156 & 0.754298 & 0.997886 & 0.976345 \\
\bottomrule
\end{tabular}
\end{center}
\caption{Same Demographic Accuracy}\label{table:NearestNeighbors_same}

\end{table*}

\begin{table*}[phbt]
\footnotesize
\begin{center}

\begin{tabular}{llrrrrrr}
\toprule
                                         {\bf Generated On} &                         {\bf Targeted Network} &     {\bf Male} &   {\bf Female} &    {\bf White} &    {\bf Asian} &    {\bf Black} &   {\bf Indian} \\
\midrule
                    Pre-trained Facenet &                        Pre-trained Facenet & 0.725676 & 0.629891 & 0.666176 & 0.657881 & 0.789189 & 0.868013 \\
              Xu et al. &                        Pre-trained Facenet & 0.995333 & 0.970667 & 0.978125 & 0.996795 & 1.000000 & 1.000000 \\
              Xu et al. &               Xu et al.  & 0.702439 & 0.685410 & 0.674922 & 0.600875 & 0.856657 & 0.943266 \\
 Race-Balanced &                        Pre-trained Facenet & 0.975610 & 0.949781 & 0.981838 & 0.990909 & 1.000000 & 1.000000 \\
 Race-Balanced & Race-Balanced & 0.910244 & 0.885191 & 0.897412 & 0.836296 & 0.999900 & 0.977273  \\
  Sex-Balanced &                        Pre-trained Facenet & 0.998667 & 0.990000 & 0.981124 & 0.990236 & 1.000000 & 1.000000 \\
  Sex-Balanced &  Sex-Balanced & 0.869593 & 0.735956 & 0.799822 & 0.754007 & 0.998498 & 0.976768 \\
\bottomrule
\end{tabular}
\end{center}
\caption{Different Demographic Accuracy}\label{table:NearestNeighbors_diff}
\end{table*}

\end{document}